\documentclass[%
amsmath,amssymb,
]{revtex4-2}
\usepackage{graphicx}
\usepackage{dcolumn}
\usepackage{bm}
\usepackage[utf8]{inputenc} 
\usepackage{hyperref}
\hypersetup{
    colorlinks=true,
    linkcolor=red,
    filecolor=magenta,      
    urlcolor=blue,}
\usepackage{url}
\usepackage{xcolor}
\usepackage{titlesec}
\usepackage[labelformat=simple]{subcaption}

\usepackage{multirow}
\usepackage{amsmath}      
\usepackage{float}
\usepackage[normalem]{ulem}
\newtheorem{theorem}{Theorem}
\newtheorem{proposition}[theorem]{Proposition}

\newtheorem{corollary}{Corollary}

\newtheorem{proof}{Proof}

\DeclareUnicodeCharacter{03B2}{\ensuremath{\beta}}

\def\R{\mathbb{ R}}

\DeclareMathOperator{\softmax}{softmax}
\usepackage{subdepth}
\usepackage{color}

\newcommand{\bu}{{\bf u}}
\newcommand{\bx}{{\bf x}}
\DeclareMathOperator{\Tr}{Tr}
\DeclareMathOperator{\E}{\mathbb{E}}

\titlespacing*{\section}
{0pt}{4ex}{2ex}
\titlespacing*{\subsection}
{0pt}{4ex}{2ex}
\titlespacing*{\subsubsection}
{0pt}{4ex}{2ex}

\begin{document}

\title{{\color{black}Easy attention: A simple attention mechanism for temporal predictions with transformers}}
\author{Marcial Sanchis-Agudo$^{\text{1}}$, Yuning Wang$^{\text{1}}$, {\color{black}Roger Arnau$^{\text{2}}$},\\ Luca Guastoni$^{\text{3}}$, {\color{black}Jasmin Lim$^{\text{4}}$}, Karthik Duraisamy$^{\text{4,5}}$, Ricardo Vinuesa$^{\text{1}}$}
\affiliation{1: FLOW, Engineering Mechanics, KTH Royal Institute of Technology, SE-100 44 Stockholm, Sweden. \\
{\color{black}2: Instituto Universitario de Matem\'atica Pura y Aplicada, Universitat Polit\`ecnica de Val\`encia. Camino de Vera s/n, 46022 Val\`encia, Spain.}\\
3: School of Computation, Information and Technology, Technical University Munich, 85748 Garching, Munich, Germany. \\
4: Department of Aerospace Engineering, University of Michigan, Ann Arbor, USA. \\
5: Michigan Institute for Computational Discovery $\&$ Engineering (MICDE),
University of Michigan, Ann Arbor, USA.}

\begin{abstract}
{To} improve the robustness of transformer neural networks used for temporal-dynamics prediction of chaotic systems, we propose a novel attention mechanism called easy attention {which we demonstrate in time-series reconstruction and prediction.} 
{\color{black}While the standard} self attention only makes use of the inner product of queries and keys, it is demonstrated that the keys, queries and {$\rm softmax$} are not necessary for {obtaining} the attention score required to capture long-term dependencies in temporal sequences. Through the singular-value decomposition (SVD) on the $\rm softmax$ attention score, we further observe that self attention compresses the contributions from both queries and keys {\color{black}in the space spanned by} the attention score. Therefore, {our proposed} easy-attention {method} directly treats the attention scores as learnable parameters. This approach produces excellent results when reconstructing and predicting the temporal dynamics of chaotic systems exhibiting more robustness and less complexity than self attention or the widely-used long short-term memory (LSTM) network. {\color{black}We show the improved performance of the easy-attention method in the Lorenz system, a turbulence shear flow and a model of a nuclear reactor.}
\end{abstract}

\maketitle

\section{Introduction}

Dynamical systems are mathematical models used to describe the evolution of complex phenomena over time. These systems are prevalent in various fields, including physics and engineering. A dynamical system consists of variables that change over time, and their interactions are governed by differential equations. The behavior of such systems can range from simple and predictable to highly complex and chaotic. Furthermore, chaotic systems are characterized by their sensitivity to perturbations of the initial conditions, leading to unpredictability and complex dynamics~\cite{Orlu_2020}.  The prediction of chaotic dynamical systems stands as a relevant and challenging area of study. Accurate forecasting of such systems holds significant importance in diverse scientific and engineering disciplines, including weather forecasting, ecological modeling, financial analysis, and control of complex physical processes. 



\noindent Dynamical-system models leveraging the concept of time delays~\cite{Takens1980,mezic2004,arbabi2017,Karthik} have been applied for temporal-dynamics prediction. These models also leverage  decomposition to obtain a reduced-order representation of the system. For instance, proper-orthogonal decomposition~\cite{POD_org} provides a set of deterministic spatial functions with corresponding time coefficients. The decomposition can also be performed in the spectral domain~\cite{SPOD}.
Another widely used method in different scientific fields is the dynamic-mode decomposition (DMD)~\cite{schmid_2010,HDMD_15M1054924} which captures intricate temporal patterns and can reveal underlying modes of motion. DMD represents a numerical approach to the approximation of the Koopman operator~\cite{mezic2004,mezic2005spectral, koopman1932dynamical,koopman_Karthik}. Koopman-based frameworks have achieved promising results in the prediction of temporal dynamics of high-dimensional chaotic systems~\cite{DMD_org,koopman_steven,khodkar2019koopmanbased,bevanda2021koopman,eivazi_2021108816}.  In the context of time-series analysis, the well-known family of auto-regressive and moving-average (ARMA) models is used for forecasting dynamical systems~\cite{box2015time_series_analysis}. These methods focus on stochastic problems and they yield interpretable models~\cite{Karthik}.
Deep neural networks (DNNs), as a branch of machine-learning (ML) approaches, have demonstrated remarkable success in temporal-dynamics prediction, offering advantages of flexibility and adaptability~\cite{Varela_2024}. The family of recurrent neural networks (RNNs)~\cite{goodfellow2016deep} has been the dominant method for temporal-dynamics prediction for many years, enabling the model to capture temporal dependencies in the time series. Among RNNs, the long short-term memory (LSTM) networks~\cite{lstm_org} overcome the training problems related to vanishing gradients and this enables modelling long-term dependencies. Herefore, LSTM networks are effective ML models for temporal predictions of complex dynamical systems~\cite{fukami_lstm_2020, srinivasan2019predictions, eivazi_2021108816}. Besides RNNs, the temporal convolutional networks (TCNs)~\cite{tcn_org}, which employ one-dimensional convolutions for multi-variable time-series prediction~\cite{bai2018empirical_tcn,tcn_app_windspeed}, have shown promising performance on temporal-dynamics prediction~\cite{non_instruct_pod_transformer}.

\noindent In recent years, transformer neural networks~\cite{vaswani2017attention} have revolutionized the field of ML, demonstrating remarkable capabilities in natural-language-processing (NLP)~\cite{transformer_nlp} and computer vision~\cite{transformer_image}, but also fluid mechanics~\cite{yousif_zhang_yu_Vinuesa_lim_2023}. Thanks to their matrix-multiplication-based attention mechanism~\cite{vaswani2017attention}, {transformers have} shown the capability to identify and predict the temporal dynamics of chaotic systems by capturing long-term {dependencies in the data}~\cite{non_instruct_pod_transformer,geneva2022transformers,solera2023beta}. This promising performance {has motivated several studies focused on} improving the attention mechanism of transformer models used for physical modeling~\cite{cao2021choose, katharopoulos2020transformers_rnn}. Furthermore, in the context of reducing computational complexity in natural language processing tasks such as machine translation and language modeling, more studies have emerged on improving the attention mechanism, in particular self-attention, have emerged~\cite{kitaev2020reformer,wang2020linformer,dao2023flashattention2}, and they have reported promising results while reducing computational cost. Note that all the mentioned improvements on self-attention follow the idea of the retrieval process which comprises queries, keys and values. However, in the present study, the approach is inspired by recent work on Koopman operators, which suggests treating attention as a finite-rank operator~\cite{ koopman_steven,khodkar2019koopmanbased,bevanda2021koopman,eivazi_2021108816}, we {propose} a new attention mechanism {denoted by} easy attention, which uses neither the {$\rm softmax$}~\cite{cao2021choose,wang2020linformer} nor {the} linear projections of queries and keys; this has the goal of reducing the complexity of the {model} and improving {its} interpretability. Our objective is to predict the temporal evolution of dynamical systems. Typically, self-attention is computed for every input; however, this choice can be less effective in chaotic systems as the input can vary arbitrarily, while keeping the same underlying system dynamics. The Kolmogorov–Arnold–Moser (KAM) Theory shows that conservative attractors remain invariant under small perturbations~\cite{Chierchia2009}. Based on this intuition, we propose an easy-attention formulation that is optimized during training and it is not input-dependent at inference time. Thus, we learn an importance score for the input variables that encodes the system dynamics rather than  the instantaneous attention of the input values. This approach provides significant performance improvements in the prediction of the trajectories of chaotic systems, as it fully exploits the spatial periodicity that is naturally present in most chaotic systems, assuming an underlying constant pattern over time, the invariant set. 

This paper is structured as follows: Section~\ref{sec:interpret} exploits the self-attention mechanism performance eigendecomposing the $\alpha$ to elaborate the rationality of easy-attention. 
Section~\ref{sec:experiments} discusses the application of easy-attention in a transformer architecture for temporal-dynamic prediction, and shows the results. The conclusions are presented in Section~\ref{sec:conclusion} and finally, Section~\ref{sec:easy-attn} offers an overview of self-attention mechanisms and introduces the easy-attention mechanism in detail.

\section{Interpretation of attention mechanism}
\label{sec:interpret}
In this section, we provide a motivating example for simplifying the {self-attention mechanism} to the proposed easy-attention technique described in the Methods section. We start with a reconstruction task of a simple wave function, using the self-attention module. Based on the data, we investigate the eigenspace given by the self-attention mechanism, and analyze the main limitation of this mechanism. Subsequently, we discuss how to address such limitation using the easy attention method and we compare the reconstruction results. 

\noindent Harmonic functions are often used in the study of temporal signals, therefore, in our example, we consider three sine waves with the same frequency but different phase shifts, expressed as:
{\color{black}
\begin{equation}
    y_i(t_j)={\rm sin}\left(t_j\frac{\pi}{2} + {i} - 1 \right); \quad t_{j}\in \mathbb{N}, \quad j,i \in \mathbb{Z},
\label{input}
\end{equation}

\noindent
where $t$ is time spanning
the range $[0,3N]$ and each $y_i$ a trajectory.
A satisfactory performance of the different attention mechanisms for this case is a prerequisite to attempt the prediction of more challenging problems. Here
our focus is on identification and reconstruction of the wave motion and we assess the results obtained just with the attention module, without any extra hidden layers and activation functions. For this case, each sample consists of the three sine waves with different phase shifts at 3 consecutive time instants, $\mathbf{y_i} = \{y_i(t_{3p}-k \Delta t)\}_{k=0}^{2}$, where $\Delta t = 1$. Interpreting $\mathbf{y}_i$ as a column vector, we can write the input matrix for the attention mechanism as $\mathbf{Y} = \text{concat} \{ {\bf y}_i \}_{i=1}^3 $, where \text{concat} refers to concatenation, as represented in Eq.~(\ref{matrix}). Here we refer to $p$ as the current sample in the batch and $k\Delta t$ as the time delay.
\begin{equation}
     \mathbf{Y} = \begin{bmatrix}
        y_1(t_{3p}-2) & y_2(t_{3p}-2) & y_3(t_{3p}-2) \\
        y_1(t_{3p}-1) & y_2(t_{3p}-1) & y_3(t_{3p}-1) \\
        y_1(t_{3p})   & y_2(t_{3p})   & y_3(t_{3p})
    \end{bmatrix}.
\label{matrix}
\end{equation}

\noindent We train the model to predict the value of the sine waves at the next three time steps, \textit{i.e.} $\{\mathbf{y}(t_{3p}+k\Delta t)\}_{k=1}^{3}$. In order to perform the training,
we adopt stochastic gradient descent (SGD) with momentum of 0.98 as optimiser, and the learning rate is set to $10^{-3}$. The mean-squared error (MSE) is employed as loss function: 

\begin{equation}
   \mathrm{MSE} = \left\langle \displaystyle \sum_{k=1}^{3} \displaystyle \sum_{i=1}^{3} \big(y_i(t_{3p}+k\Delta t) - \hat{y}_i(t_{3p} +k\Delta t) \big)^2 \right\rangle_{p \in \rm batch}, \quad \rm{batch} \subseteq [0,N]
\label{loss}
\end{equation}

\noindent 
where $y_i$ and $\hat{y}_i$ indicate the ground truth and the corresponding prediction, respectively. The model is trained for 1,000 epochs, $N=1 \times 10^3 $ and \rm{batch} size of 8. In order to compute the self attention, we need to construct the query, key and value matrices as described in Ref.~\cite{vaswani2017attention}, specifically $\mathbf{W}_Q, \mathbf{W}_K,\mathbf{W}_V \in \R ^{3 \times 3}$:
\begin{equation}
    \mathbf{Q}(\mathbf{y}_i) = \mathbf{W}_Q^T \mathbf{y}_i, \quad
    \mathbf{K}(\mathbf{y}_i) = \mathbf{W}_K^T \mathbf{y}_i, \quad
    \mathbf{V}(\mathbf{y}_i) = \mathbf{W}_V^T\mathbf{y}_i.
\end{equation}
Note that the query and key matrices are used to compute the dot product $\langle\mathbf{Q}(\mathbf{y}_i) ,\mathbf{K}(\mathbf{y}_j)\rangle$ and the attention scores $\alpha_{ij} = {\rm softmax}_j( \langle \mathbf{Q}(\mathbf{y}_i) ,\mathbf{K}(\mathbf{y}_j) \rangle / \sqrt{k} )$, where $k$ is the well known $\rm{d_{model}}$ size for transformers or latent dimension. Since the trace of the operator is invariant to cyclic permutations of its argument, we can write:
\begin{equation}
     \langle\mathbf{Q}(\mathbf{y}_i) ,\mathbf{K} (\mathbf{y}_j)\rangle = \langle\mathbf{K}(\mathbf{y}_j) ,\mathbf{Q}(\mathbf{y}_i)\rangle = 
     {\rm{Tr}(\mathbf{W}_Q \mathbf{W}_K^T \mathbf{y}_j \mathbf{y}_i^T)}.
\end{equation}

 where $\Tr$ denotes the sum of diagonal entries of its matrix argument (i.e. the usual matrix trace). Intuitively, $\Tr(\mathbf{M})$ measures the total covariance captured along each coordinate direction of the vectors feeding into $\mathbf{M}$, where $\mathbf{M} = \mathbf{W}_Q \mathbf{W}_K^T \mathbf{y}_j \mathbf{y}_i^T)$.This enables identifying two matrices: $\mathbf{W}_Q\mathbf{W}_K^T$ and $\mathbf{y_j}\mathbf{y_i^T}$. We will refer to $\alpha_{ij}$ as the transformed covariance factor given that $\boldsymbol{\alpha}= \mathbf{C} \iff \mathbf{W_Q}\mathbf{W_K}^T = \mathbb{I}$, where $\mathbb{I}$ is the identity matrix, $\mathbf{C} = \mathbf{Y}^T\mathbf{Y}$ is the covariance matrix of the input matrix $\mathbf{Y}$ and $C_{i,j} = \rm{Tr}(\mathbf{y_j}\mathbf{y_i}^T)$ the covariance factor; note that $\mathbf{y}_j\mathbf{y}_i^T$ is the covariance factor matrix. In order to gain insight into its structure, we apply singular-value decomposition (SVD)~\cite{DMD} to the matrix product $\mathbf{W_Q}\mathbf{W_K}^T\mathbf{y}_j\mathbf{y}_i^T$.
\begin{figure}[h!]
    \centering
    \begin{subfigure}{0.45\linewidth}
        
        \includegraphics[width=\textwidth]{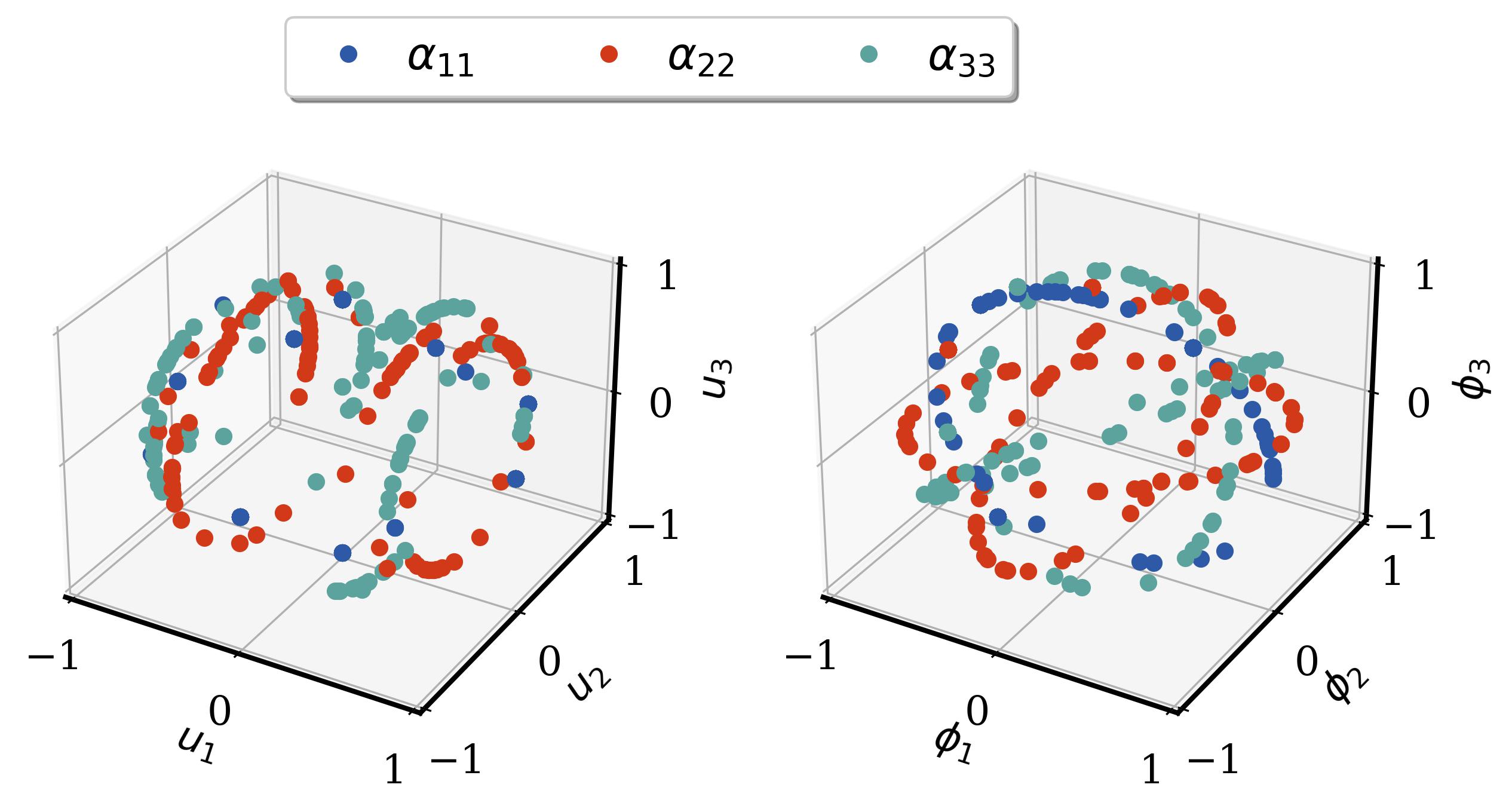}
        \caption{Decomposition of the covariance factor matrix: ${\rm SVD}(\mathbf{y}_j\mathbf{y_i}^T)$}
        \label{fig:self-combin}
    \end{subfigure}
    \quad
    \begin{subfigure}{0.45\linewidth}
        
        \includegraphics[width=\textwidth]{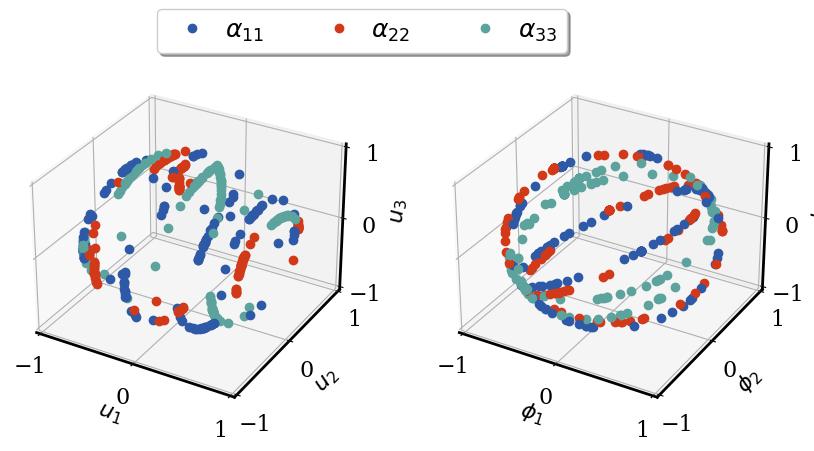}
        \caption{Decomposition after including contributions of the weights: ${\rm SVD}(\mathbf{W}_Q \mathbf{W}_K^T \mathbf{y}_j \mathbf{y}_i^T)$}
        \label{fig:qkt}
    \end{subfigure}
    \quad

    \begin{subfigure}{0.6\linewidth}
        
        \includegraphics[width=0.8\textwidth]{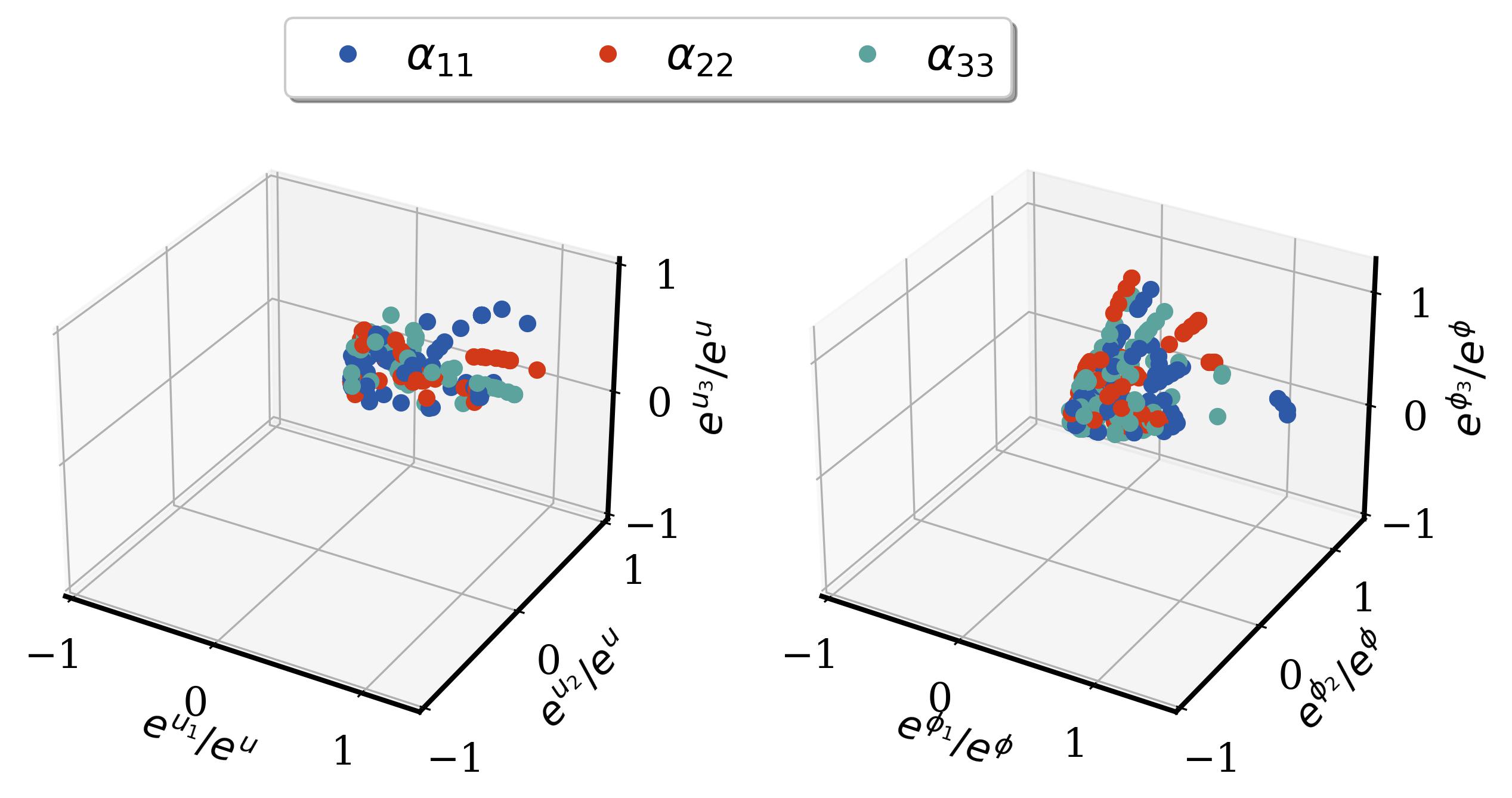}
        \caption{Decomposition of attention score: ${\rm SVD}\left(\rm{softmax} \big( \frac{\rm{Tr}(\mathbf{W}_Q \mathbf{W}_K^T \mathbf{y}_j \mathbf{y}_i^T)}{\sqrt{k}} \big) \right)$}
        \label{fig:Evo}
    \end{subfigure}

    \caption{\textbf{Three-dimensional scatter plots of left and right singular vectors ($u_i$ and $\phi_i$) of diagonal elements in the {\color{black}transformed covariance matrix} $\boldsymbol{\alpha}$ of the inputs.} {\color{black}Singular Value Decomposition of the (a) covariance factor matrix, (b) transformed covariance factor matrix} and (c) attention score. The diagonal elements are denoted by black ($\alpha_{11}$), red ($\alpha_{22}$) and cyan ($\alpha_{33}$), respectively.}
    \label{fig:Prediction Results}
\end{figure}

{\color{black}\noindent The SVD of the transformed covariance factor matrix given by $\mathbf{y}_j\mathbf{y}_i^T$ as well as the product of the query and key matrices used to later compute the attention scores, together with the corresponding scores in $\boldsymbol{\alpha}$ are shown in Fig.~\ref{fig:Prediction Results}. Note that we only show the main diagonal elements of the $\boldsymbol{\alpha}$ matrix as these are the main contributors when computing the attention score.
From the transformed covariance factor matrix, one can observe that {the spanned space by the} left singular {vectors}, $\mathbf{u}$, exhibits an approximately cylindrical shape. However, the right singular vectors ${\boldsymbol{\phi}}$ exhibit a shape closer to that of a sphere.
These shapes are connected within the unit circle, which is used to study unique sinusoidal functions. When analyzing multiple waves with different phase shifts, the unit circle is expected to grow in dimensions as the transformed covariance factor matrix between input vectors increases. It can be claimed that as $r \rightarrow \infty $ for $\mathbf{Y} \in \mathbb{R}^{3 \times r}  $; $r$ being the size of the time-delays, the sphere will become denser, creating a fully compact surface if and only if the new waves in Eq.~(\ref{matrix}) are linearly independent with the rest of input columns. The eigenspaces for queries and keys, shown in Fig.~\ref{fig:qkt}, are observed to be fixed by rotations and translations of $\mathbf{W}_Q\mathbf{W}_K ^T$. These orbits constitute the subspace of the real domain, where the inner product exists. However, after the space compression forced by the trace and {$\rm softmax$}, this subdomain is reduced to a scalar, which is shown in Fig.~\ref{fig:Evo} as a non-trivial space compression. One can interpret $\boldsymbol{\alpha}$  as the transformed covariance matrix between inputs of the transformer, such that {$\alpha_{i,j}\in[0,1], \forall i,j$},{\color{blue}} where $\alpha_{i,j}$ is defined as the scalar entry in row i, column j of the matrix $\mathbf{\alpha}$}. In fact, the $\rm{softmax}$ is a probability distribution over three different possible trajectories, each trajectory being an input vector. For instance, the second element in the first row of $\boldsymbol{\alpha}$ is expressed as:

\begin{equation}
    \alpha_{1,2} = \frac{ \exp \big(\frac{\rm {Tr}(\mathbf{y}_2 \mathbf{y}_1^T \mathbf{W}_Q\mathbf{W}_K^T)}{\sqrt{k}} \big) }{\sum_{j=1}^{3} \exp \big(\frac{\rm {Tr}(\mathbf{y}_j \mathbf{y}_1^T \mathbf{W}_Q\mathbf{W}_K^T)}{\sqrt{k}} \big) }.
    \label{eq:alpha_trace}
\end{equation}
}
\noindent {\color{black}  In Eq.~(\ref{eq:alpha_trace}), it can be observed that the numerator fulfills $\exp \big({\rm Tr}(\mathbf{W}_Q\mathbf{W}_K^T\mathbf{y}_2\mathbf{y}_1^T) / \sqrt{k} \big) = \exp \big(\sum_{i=1}^{3}\lambda_i / \sqrt{k} \big)$, where $\lambda_i$ represent the eigenvalues of the matrix product, and this is, after normalization, the {\color{black}attention} factor between $\mathbf{y}_1$ and $\mathbf{y}_2$: $\alpha_{1,2}$. The latter is a probability distribution: when we combine all possible trajectories spanned by all possible $x_j = \mathbf{V(y_j)}$ with their respective attention factors, we obtain the expected trajectory as a reconstruction for the next time {instances} of the time series $\mathbf{y}_1$, expressed as $\mathbf{\hat{y}}_1$. Therefore, for a fixed $i$ and $X_i = \{ x_1, x_2, x_3 \}$ with probabilities $p_j = \alpha_{i,j}$:
\begin{equation}
    \mathbb{E}(X_i) = \sum_{j=1}^3 p_j x_j = \sum_{j=1}^3 \alpha_{i,j} \mathbf{W}_V\mathbf{y}_j = \mathbf{\hat{y}}_i.
\end{equation}
}

\noindent When computing the trace we compress all spatio-temporal dependecies.
In a continuous {\color{black}sense}, this can be interpreted as a weighted integration over the past time.
In conclusion, the $\boldsymbol{\alpha}$ matrix highlights the temporal dependencies between time instances, while $\mathbf{W}_V$ includes the dependence of spatial inputs.
In this example, due to the independence between wave phases, $\mathbf{W}_V$ should become a near-diagonal matrix. It is not possible to make any strong claim regarding $\boldsymbol{\alpha}$, as the key of our approach is to relax any assumption on temporal dependencies and learn a hidden optimal temporal basis, {\it i.e.} local minimum.

\noindent It is important to {\color{black}recognize} the difference between language (for which the self-attention framework was mainly developed)  and physical problems. When analyzing text, the attention matrix is mostly diagonal, nearly a bijective mapping, as for each word that we input there exists a (near) one-to-one relationship with another word in the ground truth. However, for physical problems this behavior cannot be assumed and this becomes obvious during the training phase.


\noindent Based on the discussion above, it is reasonable to directly learn the exact $\boldsymbol{\alpha}$ value instead of learning the query and key matrices. Here, we propose removing the $\mathbf{W}_K$, $\mathbf{W}_Q$ and $\rm softmax$ from the self attention, which (assuming square matrices $\mathbf{W}_K$ and $\mathbf{W}_Q$ after embedding into the latent dimension $k$) reduces the computational complexity of the problem from $4n_i^2$ to $2n_i^2$, where $n_i$ is the size of the input vectors, $\mathbf{y}_i(t)$.

\noindent {\color{black}Note that other kinds of attention methods exist besides the dot-product-based one shown in Eq.~(\ref{eq:alpha_trace}).
This is the case of the neural machine translation version presented in Ref.~\cite{bahdanau2014neural}, in which the attention scores are computed directly as the output of a recurrent neural network the inputs of which are the past observations.
However, in easy attention, those scores are learned and then fixed, so they are independent on the past observations, allowing for a constant-operator interpretation of the model as in Koopman theory and Hamiltonian mechanics.
}

\noindent At this stage, we employ the easy-attention approach using an input of the same size as that of the self attention, and the same training setup for sinusoidal-waves reconstruction.
We evaluate the reconstruction accuracy by the relative $l_2$-norm error defined as: 
\begin{equation}
    \varepsilon = \frac{|| S - \Tilde{S} ||_2}{||S||_2}, 
    \label{eq:l2-error}
\end{equation}
\noindent where $S$ is the {ground-truth} sequence and $\Tilde{S}$ is the prediction obtained {by the} attention module. In the Supplementary Material we show details regarding $l_2$-norm errors, computation time for training ($t_c$) and number of floating-point operations ($N_f$) of the employed easy-attention and self-attention modules. The results indicate that easy attention reduces complexity by 50\%, as compared to self-attention, under identical embedding dimensions, saves computation time by 26\% and outperforms the self-attention, with an error of $0.0018\%$ compared with the $10\%$ given by the latter. To conclude the intepretation of the attention mechanism,  



\section{Temporal-dynamics predictions of chaotic systems using a transformer}
\label{sec:experiments}
In the following section we will use a modified transformer encoder with easy attention in order to conduct a systematic study regarding the performance of the latter against other state-of-the-art deep-learning models, LSTM and self-attention transformers. In particular, we first study the Lorenz system (a three-dimensional canonical chaotic system), secondly the Moehlis model~\cite{moehlis2004low} for wall-bounded turbulence (which is a system comprising nine ordinary differential equations) and finally a model~\cite{osti_1868762} of a Fluoride-salt-cooled high-temperature (FHR) reactor.


\subsection*{Lorenz system}

\noindent The Lorenz system~\cite{lorenz1963deterministic} is governed by:
\begin{equation}
    \frac{{\rm d}x}{{\rm d}t} = \sigma (y - x), \quad \frac{{\rm d}y}{{\rm d}t} = x (\rho - z) - y, \quad \frac{{\rm d}z}{{\rm d}t} = xy - \beta z,
    \label{eq:lorentz_equations}
\end{equation}
\noindent where $x$, $y$, and $z$ are the state variables, whereas $\sigma$, $\rho$, and $\beta$ are the system parameters. We use the classical parameters of $\rho$ = 28, $\sigma$ = 10, $\beta$ = 8/3, which lead to chaotic behaviour. For this numerical example, we provide random initial conditions with uniform distributions $x_0 \sim \mathcal{U}(-5, 5)$, $y_0 \sim \mathcal{U}(-5, 5)$, $z_0 \sim \mathcal{U}(-5, 5)$, where $\mathcal{U}$ is the discrete uniform distribution in the interval $[-5,5]$, to generate 100 time series for training and validation with a split ratio of 80:20. Each time series contains 10,000 time steps with a time-step size of $\Delta t$ = 0.01 solved using a Runge--Kutta numerical solver. The test data is generated by integrating the system with new initial states of $x_0 =  y_0 =  z_0 = 6$ with independent  perturbations $\epsilon_x$, $\epsilon_y$ and $\epsilon_z$ where $\epsilon_{i} \sim \mathcal{N}(0,1)$ for each variable. {Note that $\mathcal{N}(0, 1)$ denotes a normal distribution with 0 average and a standard deviation of 1.}
We adopt the same time-step size and number of time steps of 10,000 to generate one time series as test data. We employ the relative $l_2$-norm error $\varepsilon$ to assess the prediction accuracy, which is expressed as: 
\begin{equation}
    \varepsilon = \frac{||\mathbf{x} - \hat{\mathbf{x}}||_2}{||\mathbf{x}||_2} \times 100\%,
    \label{eq:error_eps}
\end{equation}
\noindent {\color{black}where $\hat{\mathbf{x}} = (\hat{x}, \hat{y}, \hat{z})$ and $\mathbf{x} = (x,y,z)$} are the model prediction and the ground truth from numerical solver, respectively. Following the assessment in Ref.~\cite{geneva2022transformers}, {\color{black}we focus on the short-time accuracy and compute the error on the first 512 time steps of the test sequence. The time window in which the error is computed is then $T_\varepsilon=5.12$ convective time units.
We also assess the computation time of training ($t_c$) and number of floating-point operations ($N_f$) for one forward propagation with batch size of 1. {\color{black}We summarize the results in the supplementary material in Tab.~\ref{tab:metrics_lorentz}}, where the proposed model using dense easy attention yields the lowest error, 1.99\%, demonstrating the superior performance of the easy-attention method applied on temporal-dynamics predictions. The LSTM model has the highest number of learnable parameters, but the lowest training time and number of floating-point operations. However, it yields the highest $l_2$-norm error of 37.68\%, which indicates that the LSTM model is not able to reproduce the correct dynamical behaviour over a long time given the same set up as the transfomer. The dense-easy-attention model reduces the time for training by 17\% and the computational complexity by 25\% with respect to the self-attention-based model while surpassing the number of learnable parameters of the self-attention-based model. This is due to the fact that the dense-easy-attention module comprises $\boldsymbol{\alpha} \in \mathbb{R}^{h \times p \times p}$, where $h$ is the number of heads, $p$ the size of the input after embedding, and $\mathbf{W}_V \in \mathbb{R}^{p \times d_{\rm model}}$ while the self-attention module comprises $\mathbf{W}_Q, \mathbf{W}_K, \mathbf{W}_V \in \mathbb{R}^{p \times d_{\rm model}}$ which has fewer parameters in the present study. However the computational cost for the easy-attention models is lower than that of the self attention since the softmax and the inner product inside the latter are eliminated. Note that the model employing sparse easy attention exhibits an error of 2.79\% while only having 52\% of the learnable parameters of the self-attention-based transformer, which implies that the sparsification induced by an offset of 0 benefits the robustness of the easy-attention method applied to temporal-dynamics prediction. The sparsity is defined as follows: the $\boldsymbol{\alpha}$ matrix has a number of non-zero diagonals equal to $2\cdot \rm{offset}+1$. \textcolor{black}{We also implement the averaged relative Euclidean norm error $\psi(t)$ between the ground truth and the prediction, which is proposed in Ref.~\cite{eivazi_2021108816}: 
\textcolor{black}{\begin{equation}
    \psi(t)  = \left \langle  \frac{\lVert \mathbf{x}(t) - \hat{\mathbf{x}}(t) \rVert_2}{\left \langle\lVert \mathbf{x}( t) \rVert_2 \right \rangle_{ t}}     \right \rangle_{\rm ens},
\end{equation}}\noindent where $\langle \cdot \rangle_{\rm ens}$ and $\langle \cdot \rangle_t$ indicate ensemble averaging over 100 sets of time series and over 10,000 time steps, respectively. Note that the set of time series used for evaluation are again generated by integrating the system with new initial states of $x_0 =  y_0 =  z_0 = 6$ with independent  perturbations $\epsilon_x$, $\epsilon_y$ and $\epsilon_z$ where $\epsilon_{i} \sim \mathcal{N}(0,1)$ for each variable. Fig.~\ref{fig:lorentz_eps_t} depicts the evolution $\psi(t)$ for the different models. One can observe that the easy-attention-based transformer deviates from the ground truth later than the LSTM and the self-attention-based transformer in terms of short-term prediction. We adopt a threshold of $\psi$ = 0.4 as an adequate agreement with respect to the reference trajectory, it can be observed that the transformers using easy attention and sparse easy attention produce accurate instantaneous predictions for 7.04 and 5.97 time units, respectively. However, the self-attention-based transformer and the LSTM yields 4.90 and 1.10 respectively. The results indicate that the easy-attention-based transformer outperforms the self-attention-based transformer and the LSTM for short-term prediction.}

\begin{figure}[ht]
    \centering
    \includegraphics[width=0.8\textwidth]{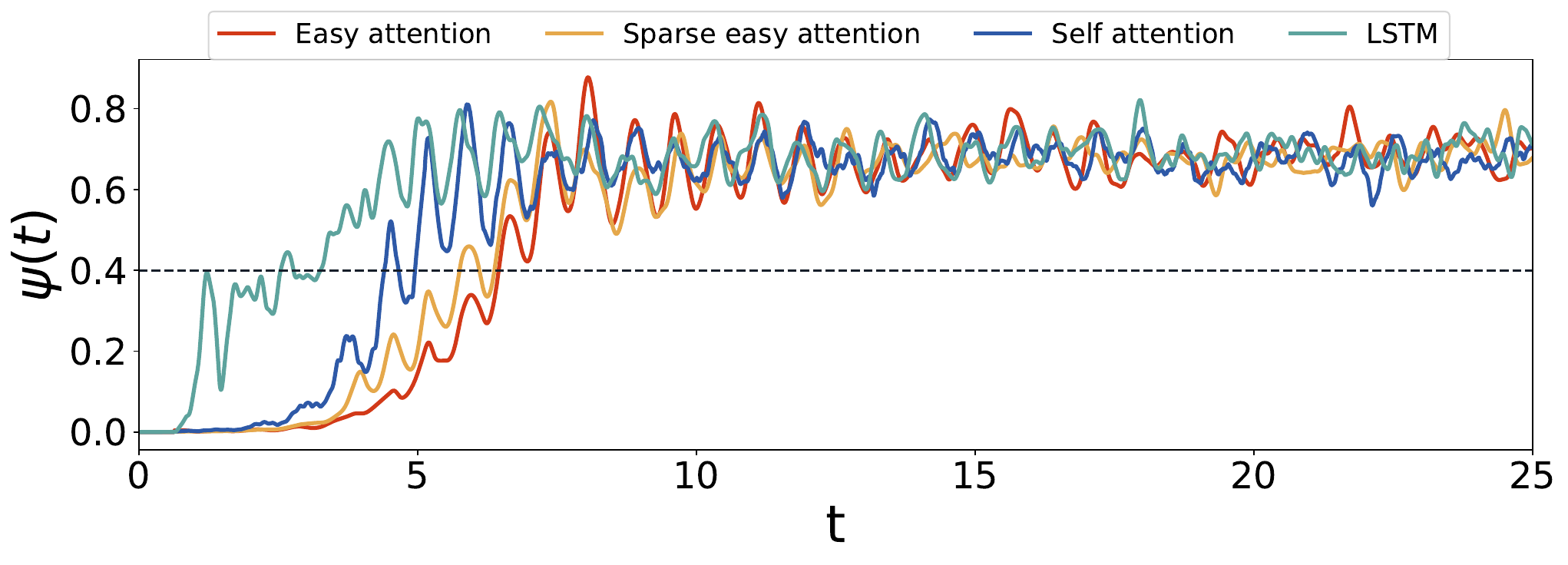}
    \caption{\textcolor{black}{{\bf Assessment of short-term predictions for the various models.} Relative Euclidean norm of error $\psi(t)$ averaged over 100 sets of time series comprising 10,000 time steps, generated by integrating the system with new initial states of $x_0 =  y_0 =  z_0 = 6$ with independent  perturbations $\epsilon_x$, $\epsilon_y$ and $\epsilon_z$ where $\epsilon_{i} \sim \mathcal{N}(0,1)$ for each variable. The dashed horizontal line shows the threshold value considered for accurate predictions, namely $\psi(t)$} = 0.4.}
    \label{fig:lorentz_eps_t}
\end{figure}

\noindent {\color{black}Moreover, it can be observed that all the models produce similar levels of accuracy for long-term predictions. Due to the chaotic behavior of the Lorenz system and the fact that the errors accumulate over time, we do not only aim to keep $\psi(t)$ low for large $t$, but to preserve the general trajectory dynamics.
To this end, we report the so-called Lorenz map (see Ref.~\cite{lorenz1963deterministic}) in Fig.~\ref{fig:lorentz_analysis}. The Lorenz map shows the evolution of two consecutive maximum values of $z$ for the various prediction methods. This figure shows that, while the two easy-attention-based transformers accurately reproduce the trend of the ground truth, the self-attention-based transformer shows some deviations and the trend from the LSTM significantly deviates with respect to the reference. 

\noindent Furthermore, we compute the dominant Lyapunov exponent using the method described in Ref.~\cite{srinivasan2019predictions} to further examine to what extent the various methods can reproduce the sensitivity in the chaotic behavior of the reference chaotic system. For two time series, 1 and 2,  we define the deviation of their respective trajectories in terms of the Euclidean norm: 
\begin{equation}
    |\delta \mathbf{A}(t)|  = \left [ \sum^3_{i=1}(x_{i,1}(t) - x_{i,2}(t))^2 \right]^{\frac{1}{2}},
    \label{eq:lyap}
\end{equation}
\noindent and we denote the separation at $t = t_0$ as $|\delta \mathbf{A}_0|$. The initial divergence of both trajectories can be assumed to behave as $|\delta \mathbf{A}(t')| = \exp(\lambda t')|\delta \mathbf{A}_0|$, where $\lambda$ is the so-called Lyapunov exponent and $t' = t - t_0$. In the present study we introduced a perturbation, with norm $|\delta \mathbf{A}_0|=10^{-6}$ at $t_0 = 400$, where all the variables are perturbed, and subsequently we analyzed its divergence with respect to the unperturbed trajectory. Fig.~\ref{fig:lorentz_analysis} (bottom) depicts the Lyapunov exponents obtained for the Lorenz system and models via ensemble averaging 10 perturbed test datasets with initial states of $x_0 =  y_0 =  z_0 = 6$, and predictions done after $t = 500$ with the studied models.

\noindent The easy-attention based transformer yields a comparable Lyapunov exponent of $\lambda = 0.919$, which is in very good agreement with the ground truth of $\lambda = 0.929$ and other numerical approximations~\cite{pathak2017using}. This indicates that the model learns the chaotic nature of the Lorenz system. However, the model using sparse easy attention yields $\lambda = 0.823$, deviating from the ground truth. One possible reason is the reduction of parameters in the attention module, which leads to the model becoming not sensitive enough to detect the separation of the trajectories via perturbation. It can be stated, the easy-attention performance relies solely on the quality of the available data. Given richer training data it could be possible to improve the prediction of the Lyapunov exponent, replicating more realistic behaviors.

\begin{figure}[ht]
    \centering
    \begin{subfigure}{0.7\linewidth}
        \centering
        \includegraphics[width=0.7\textwidth]{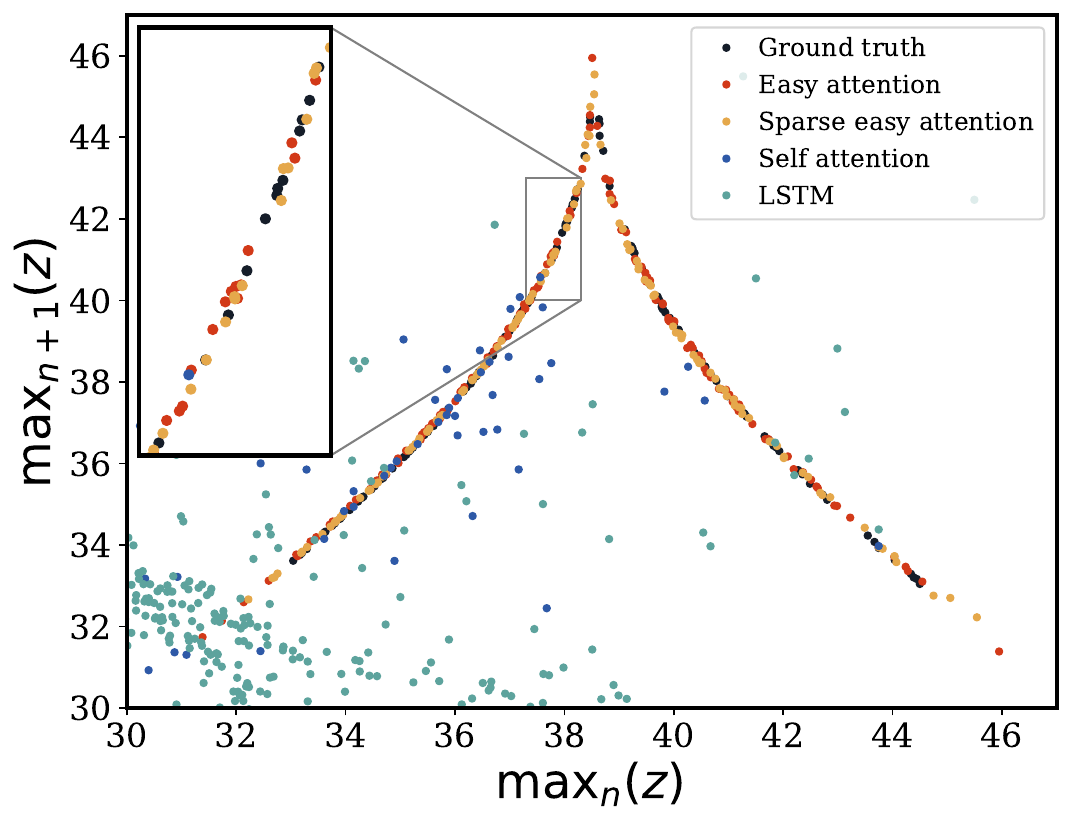}
        \label{fig:Lorentz_map}    
    \end{subfigure}
    \quad
    \begin{subfigure}{0.7\linewidth}
        \centering
        \includegraphics[width=\textwidth]{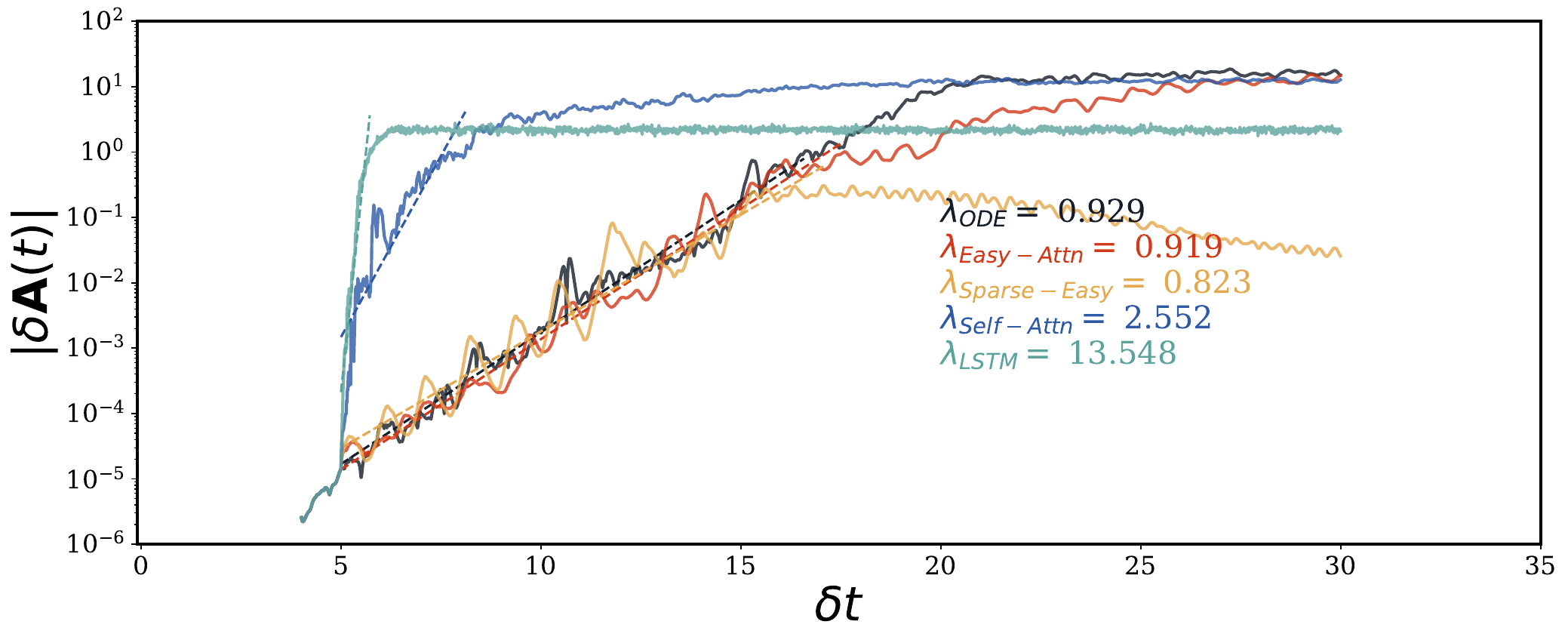}
        \label{fig:Lorentz_lyap}    
    \end{subfigure}
    \caption{\textcolor{black}{{\bf Characterization of the dynamics of the Lorenz system}.(Top) The Lorenz map representing the two consecutive local maxima of $z$ produced by each model. Note that we amplify the region where $\max_n (z) \in (37.3, 38.3)$ and $\max_{n+1} (z) \in (40, 43)$. (Bottom) Leading Lyapunov exponents obtained by adding a perturbation at time step of 400 with a magnitude of $ \delta=1\times10^{-6}$. Note that we report the obtained Lyapunov exponents by ensemble averaging 10 perturbed test dataset with initial states of $x_0 =  y_0 =  z_0 = 6$ whereas ODE denotes the results obtained by Lorenz system. The obtained Lyapunov exponents were scaled by the adopted time step $\Delta t = 0.01s$ in the present study, and $\Delta t$ denotes the time after the introduction of the perturbation. The dashed lines denote the fitted exponential evolution used to obtain the corresponding Lyapunov exponents.}}
    \label{fig:lorentz_analysis}  
\end{figure}



\noindent {\color{black} Furthermore, Fig.~\ref{fig:lorentz_pdf} depicts the probability density functions (PDF) of the $x$ and $z$ variables. To highlight the entire path and avoid focusing near $(0,0,0)$, an arc-length parametrization is performed to ${\bf x}(t) = \big( x(t), y(t), z(t) \big)$ to obtain a natural curve ${\bf \hat x}(s) = {\bf x}[t(s)]$. This is done by considering $t(s)$, the inverse function of $s(t) = \int_{0}^t \| {\bf x}'(r) \|_2 {\rm dr}$.
As before, the models with the most similar PDF to the one of the test data are the two easy-attention-based transformers.
Both exhibit a better performance when considering the symmetry on the $x$ axis, as shown below. It is necessary to mention the importance of replicating symmetries when learning chaotic systems as the main exoticism of complex systems relies on their unique structures or invariant-sets, spatio-temporally; attractors and probability density functions. Being able to reproduce such structures it is not a trivial consequence of the training procedure: while the model has as single objective temporal predictions, inspecting the results in Fig.~\ref{fig:lorentz_pdf} one can conclude that not only the space is well reconstructed but the velocity, the time derivative of the state variable too, as the frequency per volume unit is also learned. In Fig.~\ref{fig:lorentz_pdf} we depict the embedded manifold given by the joint PDF (JPDF). Note the excellent agreement with the ground truth observed after the symmetry transformation.} The possibility of accurately reproducing spatio-temporal structures will become critical in more challenging chaotic problems such as turbulence. 

\begin{figure}[ht]
    \centering
    \begin{subfigure}{0.3\linewidth}
        \centering
        \caption{Easy attention}
        \includegraphics[width=0.9\textwidth]{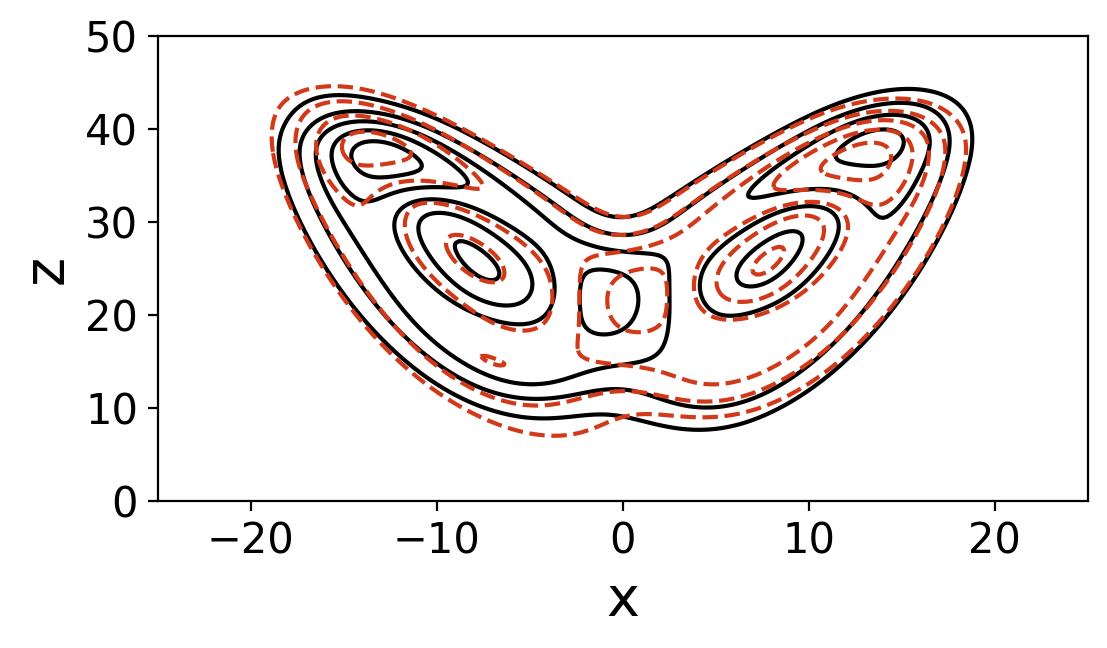}
    \end{subfigure}
    \quad
    \begin{subfigure}{0.3\linewidth}
        \centering
        \caption{Sparse easy attention}
        \includegraphics[width=0.9\textwidth]{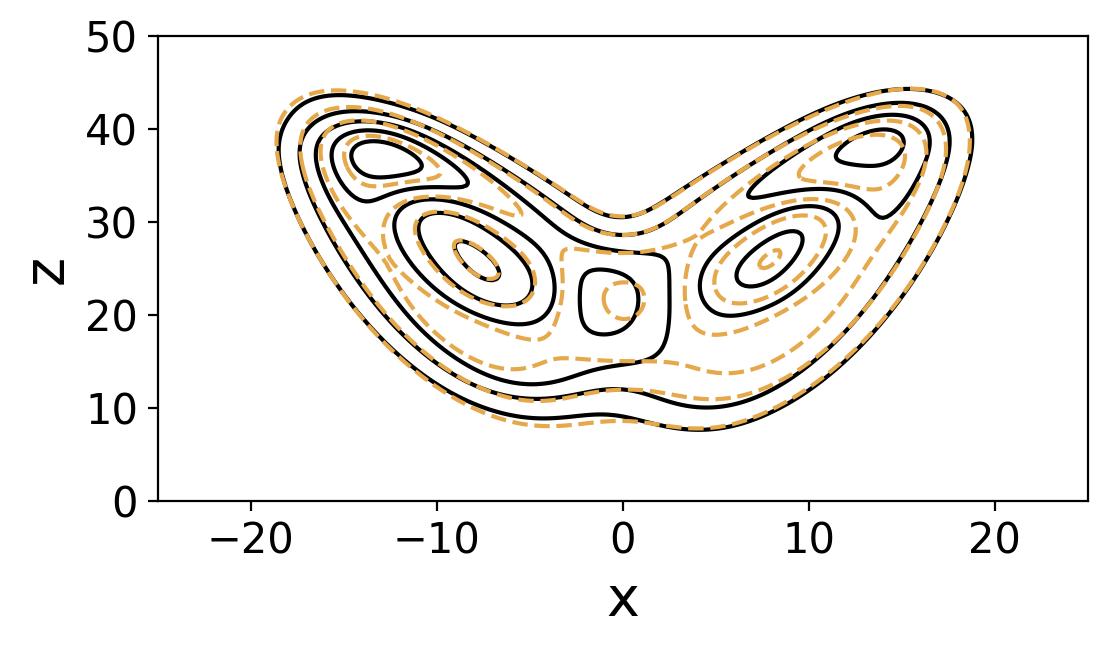}
    \end{subfigure}
    \quad
    \begin{subfigure}{0.3\linewidth}
        \centering
        \caption{Self attention}
        \includegraphics[width=0.9\textwidth]{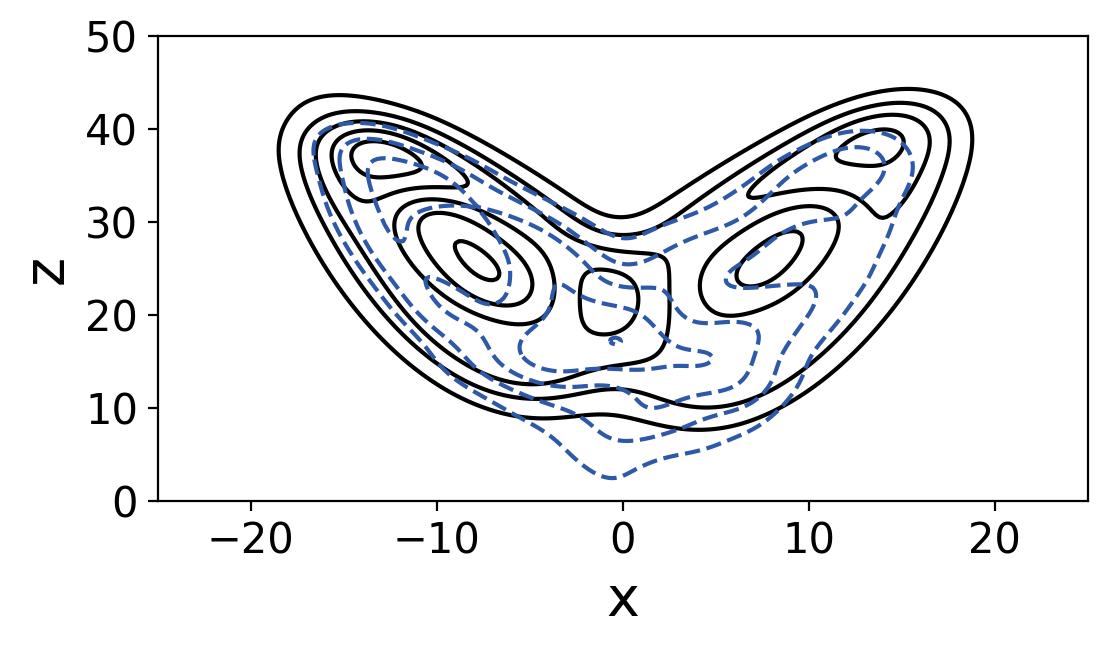}
    \end{subfigure}

    \begin{subfigure}{0.3\linewidth}
        \centering
        \caption{Easy attention (Flipped)}
        \includegraphics[width=0.9\textwidth]{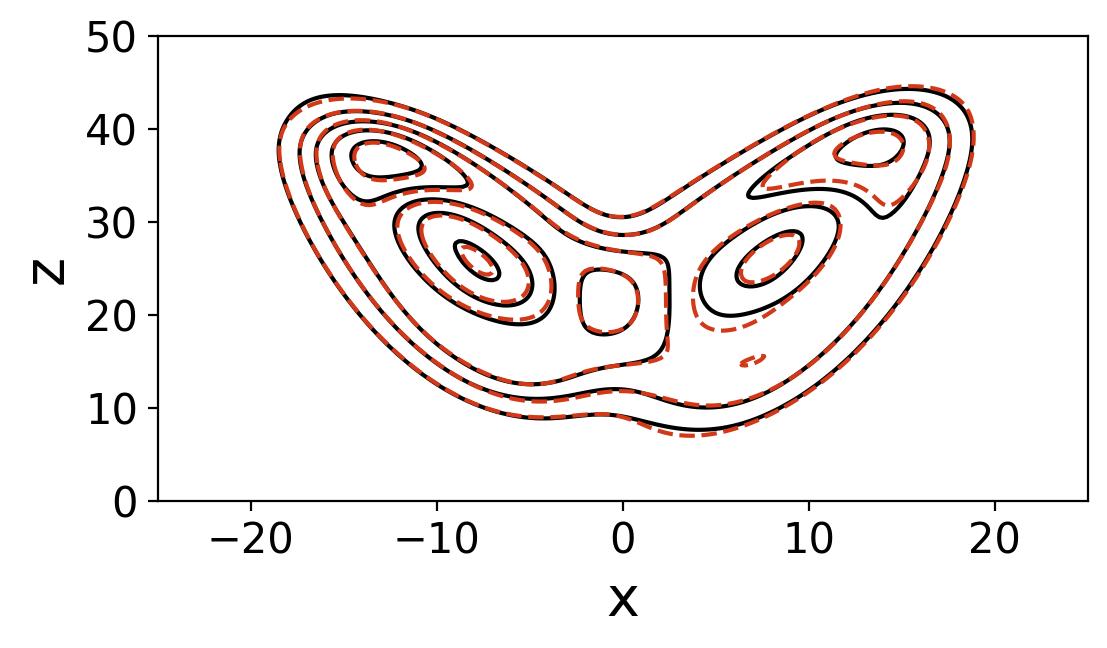}
    \end{subfigure}
    \quad
    \begin{subfigure}{0.3\linewidth}
        \centering
        \caption{Sparse easy attention (Flipped)}
        \includegraphics[width=0.9\textwidth]{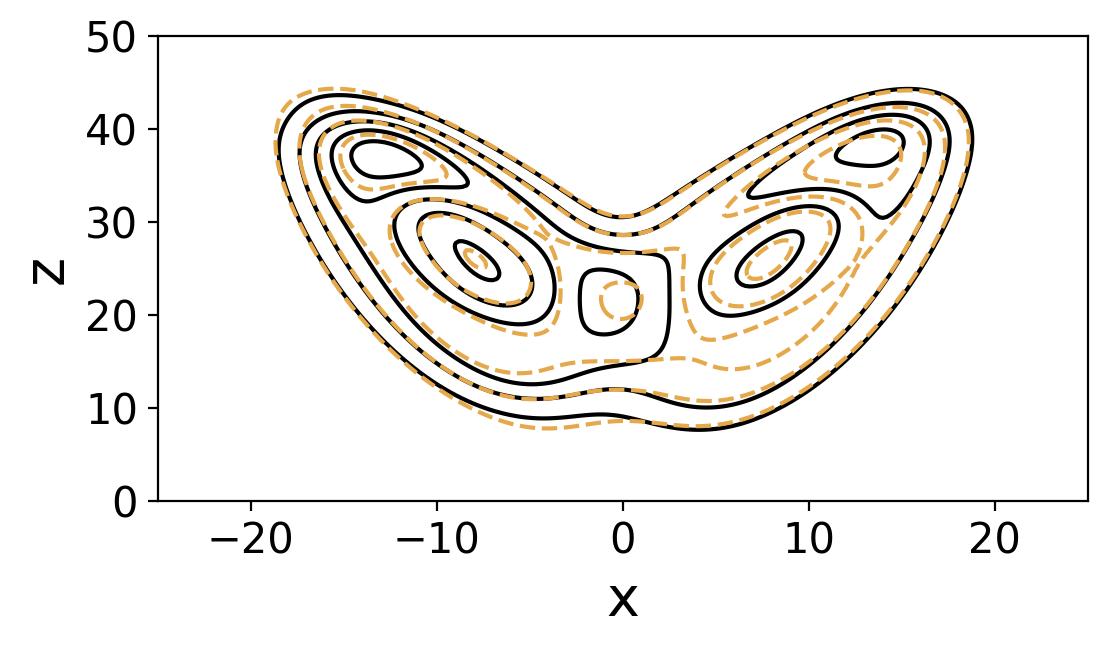}
    \end{subfigure}
    \quad
    \begin{subfigure}{0.3\linewidth}
        \centering
        \caption{LSTM}
        \includegraphics[width=0.9\textwidth]{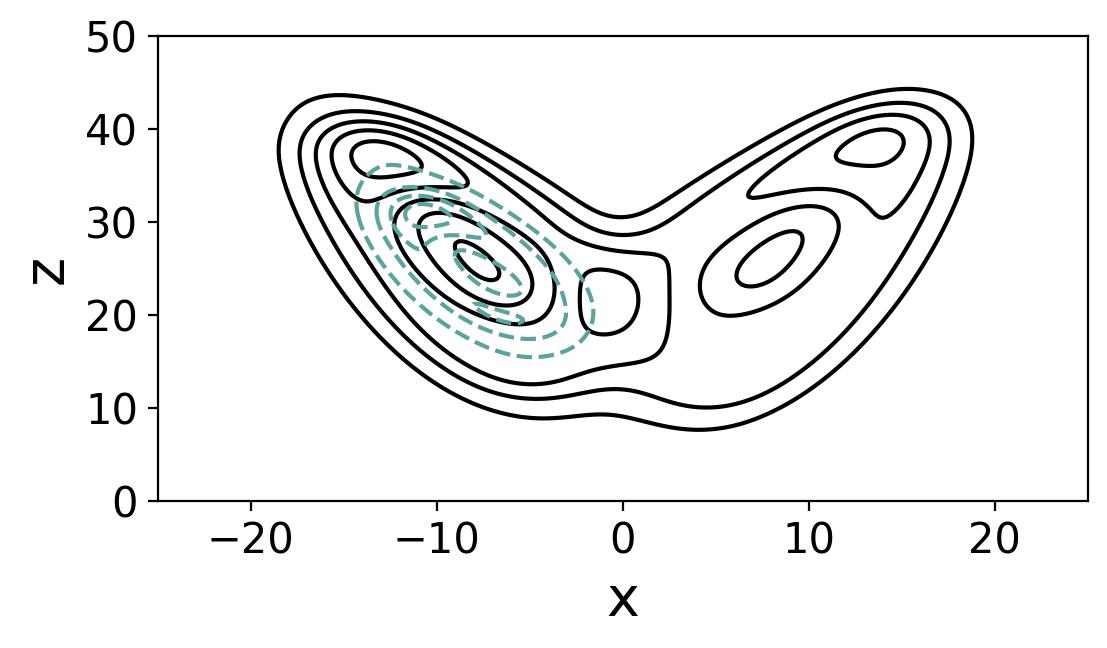}
    \end{subfigure}
    
    \caption{{\color{black} {\bf Joint contour plots of the probability density functions of variables $x$ and $z$ of the Lorenz system and predictions using different models}. Note that the trajectory has been parametrized by the arc length and the prediction is flipped on the $x$ axis in the bottom panels showing easy-attention models.}}
    \label{fig:lorentz_pdf}
\end{figure}

\noindent{\color{black}Compared with the self-attention-based transformer, the transformers using easy attention have lower computational cost for training and exhibit higher accuracy on reproduction of long-term dynamics, which indicates that the transformer models using easy attention are more robust for temporal-dynamics prediction than the model using self attention.}

\subsection*{Turbulence shear flows}
\label{sec:experiments1}
\noindent We now use the easy-attention-based seq2one transformer model to predict a more complex chaotic system with a higher dimension of 9, namely the low-order representation of near-wall turbulence described by the model proposed by Moehlis {\it et al.}~\cite{moehlis2004low}. The system comprises nine spatial modes $\bu_m(\mathbf{x})$ which represent the velocity mean profile, streamwise vortices, the streaks and their instabilities as well as their coupling~\cite{eivazi_2021108816}. 

\noindent In this model, the Reyonlds number $Re$ is defined in terms of the laminar velocity $U_0$ at a distance of $h/2$ from the top wall where the full height of channel is $h$. In the present study, following the previous studies~\cite{srinivasan2019predictions, eivazi_2021108816}, we consider $Re = 400$ and employ $U_0$ and $h$ as velocity and length scales, respectively. The ODE model was used to produce over 10,000 time series of the nine amplitudes, each with a time span of 4,000 time units, for training and validation with a ratio of 80:20. The domain size is $L_x = 4\pi$, $L_y = 2$ and $L_z = 2\pi$, where $x$, $y$ and $z$ are the streamwise, wall-normal and spanwise coordinates respectively, and we consider only time series that are chaotic over the whole time span, which means that we do not consider relaminarized cases. The instantaneous velocity field is given by $\bu(\bx ,t) = \sum_{m=1}^9 a_m(t) \bu_m(\bx)$, where $\bx = (x, y, z) \in D \subseteq \R^3$ denotes the spatial coordinates, $t \in [0, T_{\rm max}]$ the time and $D$ is the spatial domain. From the incompressible Navier--Stokes equations, the free-slip boundary conditions, and computing the Galerkin projections, the following system of ordinary differential equations for the amplitudes $a_m(t)$ is obtained:
\begin{align}
\begin{split}
    \frac{{\rm d}a_1}{{\rm d}t} & = - \eta_1 + \eta_1 a_1 + \phi_1( a_1, a_2, \ldots, a_9 ), \\
    \frac{{\rm d}a_m}{{\rm d}t} & = \eta_m a_m + \phi_m( a_1, a_2, \ldots, a_9 ) \qquad m = 2, 3, \ldots, 9.
\end{split}
\label{eq:EDO_9}
\end{align}

\noindent As introduced above, we will investigate the transformer's ability to reconstruct and capture both temporal and spatial dynamics for this turbulent shear flow. We will use essentially the same architecture as that used for the Lorenz system, changing the embedding module from time2vec to the time-space embedding module proposed in Refs.~\cite{solera2023beta,WANG2024109254}, see Tab.~\ref{tab:attn_arch_9eq} in the Supplementary Material for additional details. We will also compare these results with the one-layer LSTM architecture reported in Ref.~\cite{eivazi_2021108816}.
}

\noindent In Fig.~\ref{9eq_poincare} (top) we show the PDF of the Poincar\'e section with $a_2=0$ (and ${\rm d}a_2 / {\rm d} t <0$) for variables $a_1$ and $a_3$. It can be observed that both the easy-attention-based transformers and the LSTM reconstruct accurately the spatial behavior depicted by the reference Poincar\'e section. On the other hand, the self attention does not exhibit the level of accuracy produced by the other models. Furthermore, Fig.~\ref{9eq_poincare} (bottom) shows reference and predicted turbulence statistical quantities, {\it i.e.} the streamwise mean velocity profile $\overline{u}$, the streamwise velocity fluctuations $\overline{u'^2}$ and the Reynolds shear stress $\overline{u'v'}$. These quantities are obtained by averaging 500 time series spanning 4,000 time units each with $\Delta t = 0.01$ and over the periodic directions $x$ and $z$. Our results show that the LSTM yields the best reconstruction for the mean velocity with a relative error of $0.45 \%$ while the easy attention remains the most accurate when studying the streamwise velocity fluctuations: $0.42 \%$. Taking this into account and considering the superior robustness of the easy attention exemplified in the Lorenz system, it can be concluded that easy-attention-based transformer outperforms the self attention when predicting this turbulent-flow model. The sparse easy attention shows really promising results when considering how the accuracy of the results is not affected drastically by the parameter reduction induced by the sparsity imposed on the attention matrix. Based on the results reported here, easy attention remains the preferable model as it slightly underperforms LSTM networks when reconstructing the streamwise mean velocity by $0.37$ percentage points (\rm{pp}) but outperforms the latter by $2.07$ \rm{pp} considering the streamwise velocity fluctuations.
{\color{black}When studying complex systems through deep learning it is important to carefully select the loss function. In the present work all the models are trained wit the purpose of obtaining, an accurate reconstruction and prediction of the next step in the temporal series, a goal which is represented by Eq.~(\ref{loss}). Improved performance in other metrics may be possible by properly designing the training stage~\cite{wayland2024mapping}, leveraging topological studies on the different realizations explored during the training stage.




\begin{figure}[ht]
            \centering
            \begin{subfigure}{0.23\linewidth}
                \centering
                \includegraphics[width=\textwidth]{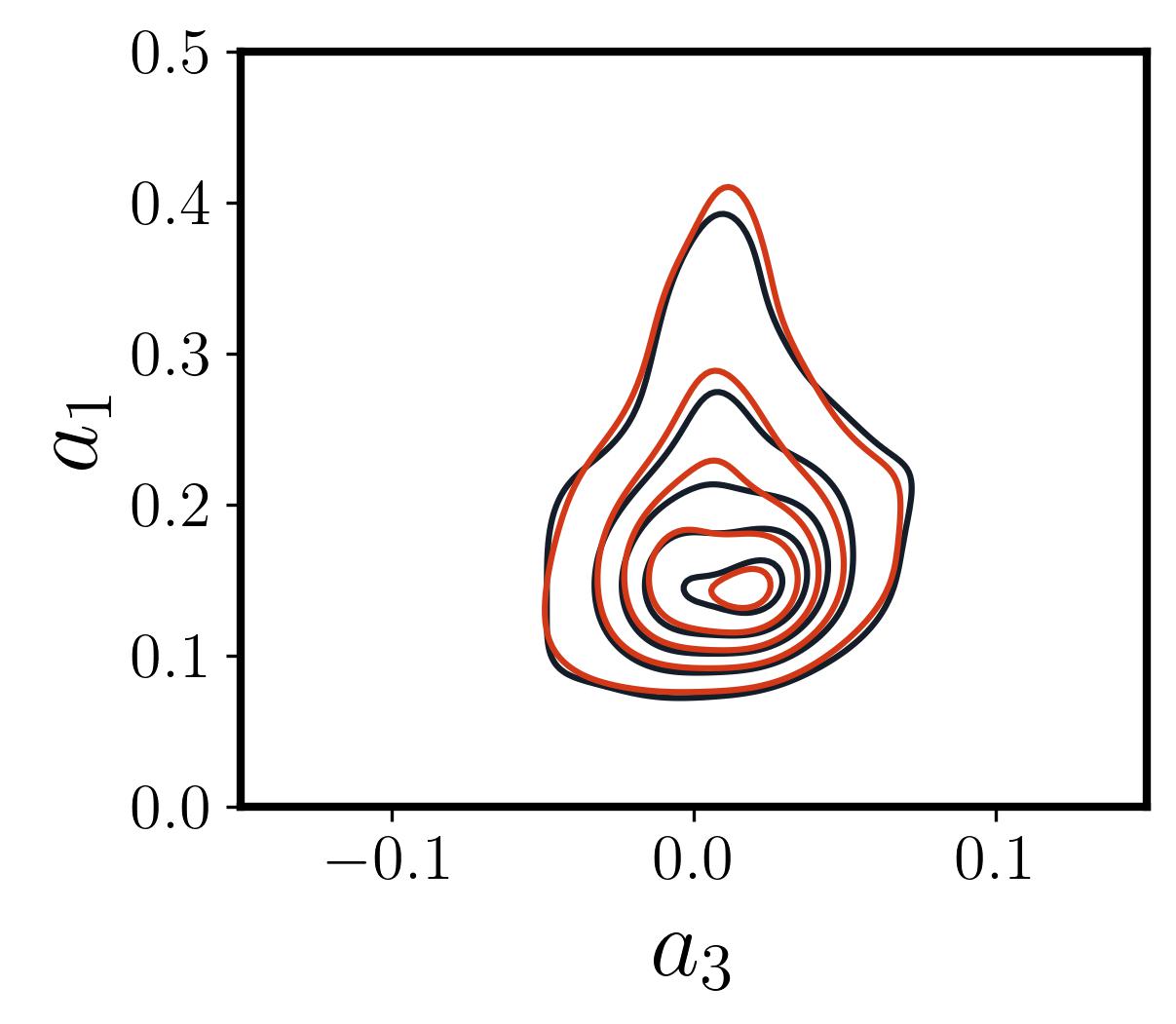}
                
            \end{subfigure}
            \quad
            \begin{subfigure}{0.23\linewidth}
                \centering
                \includegraphics[width=\textwidth]{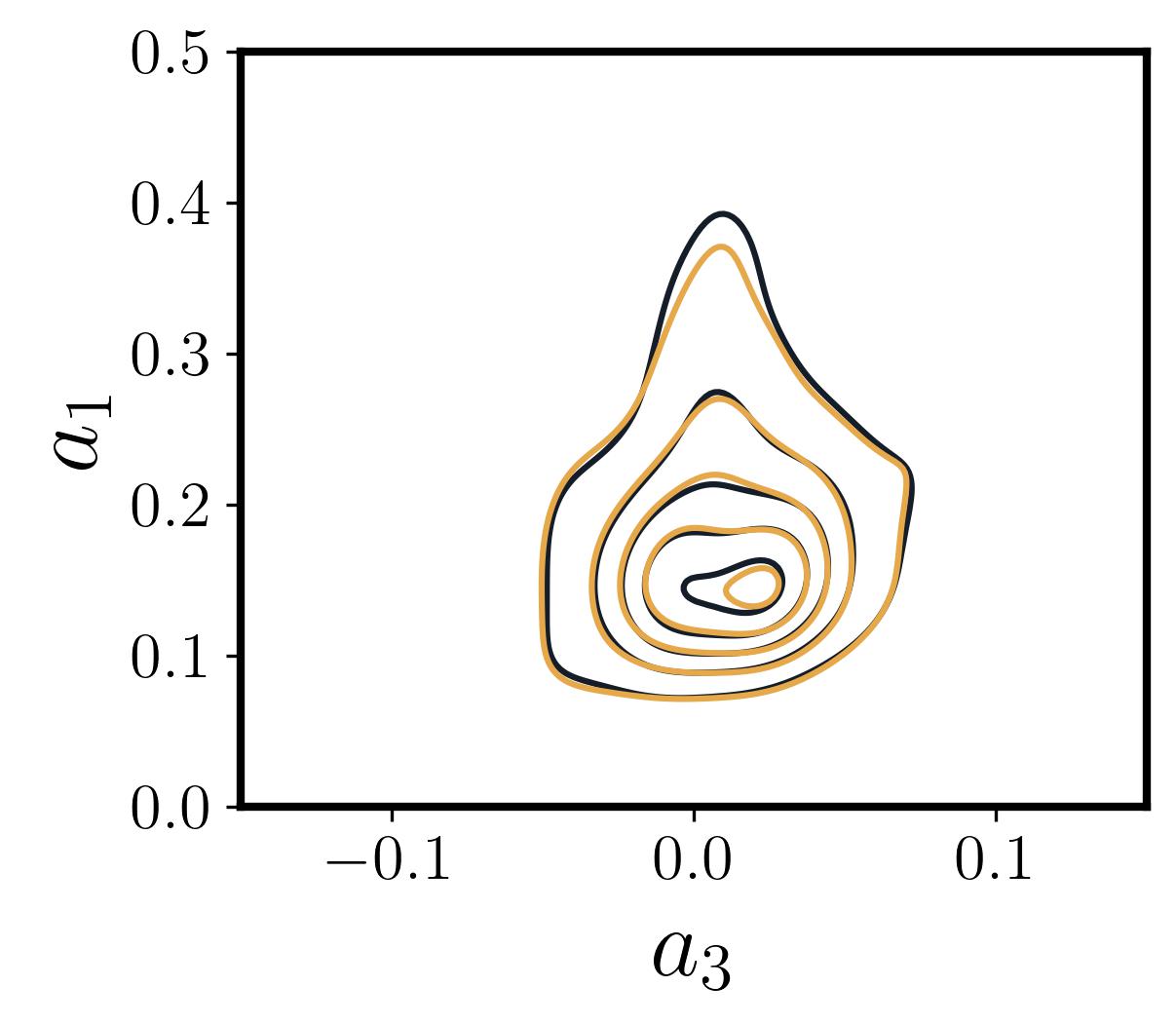}

            \end{subfigure}
            \quad
            \begin{subfigure}{0.23\linewidth}
                \centering
                \includegraphics[width=\textwidth]{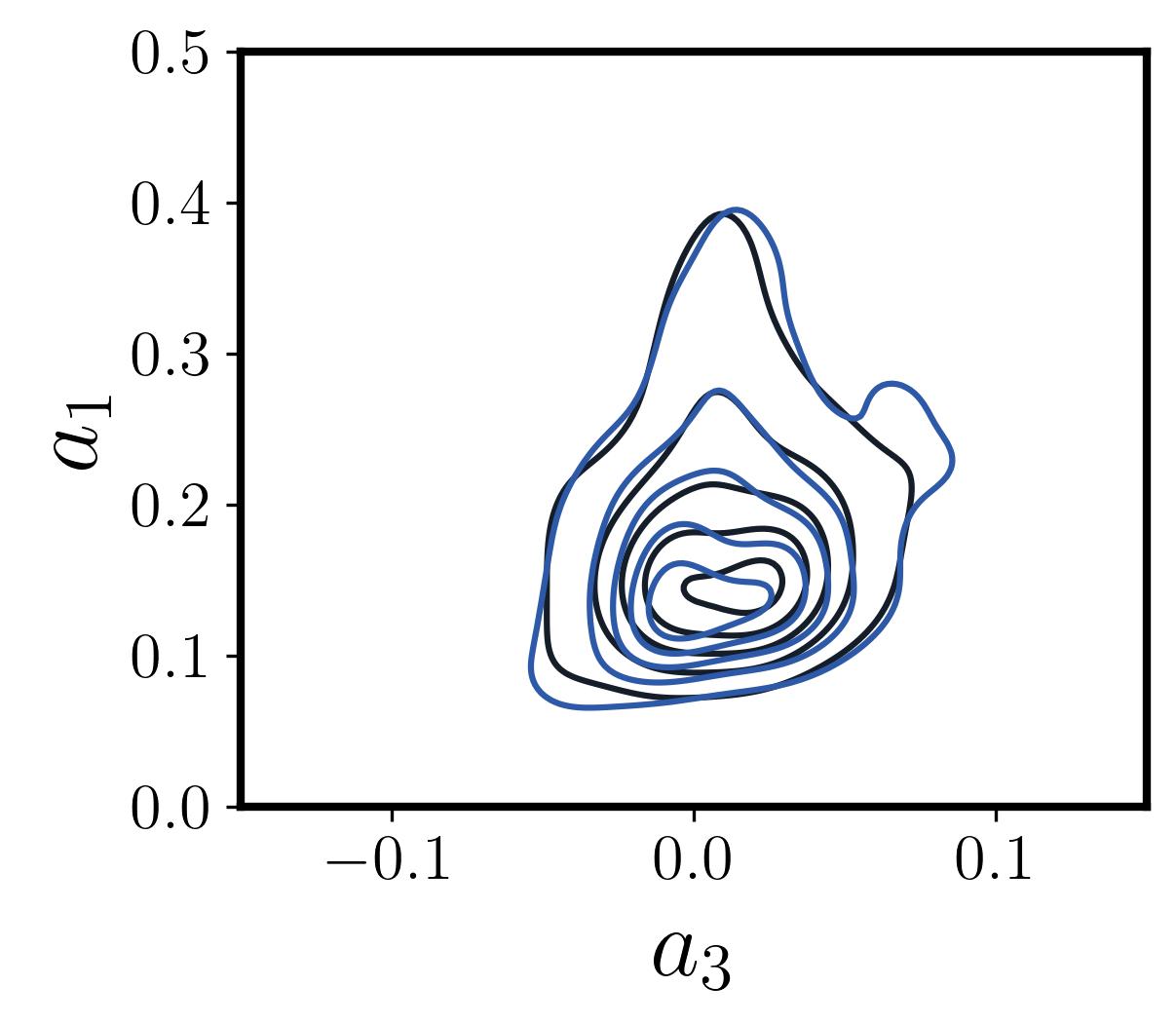}

            \end{subfigure}
            \quad
            \begin{subfigure}{0.23\linewidth}
                \centering
                \includegraphics[width=\textwidth]{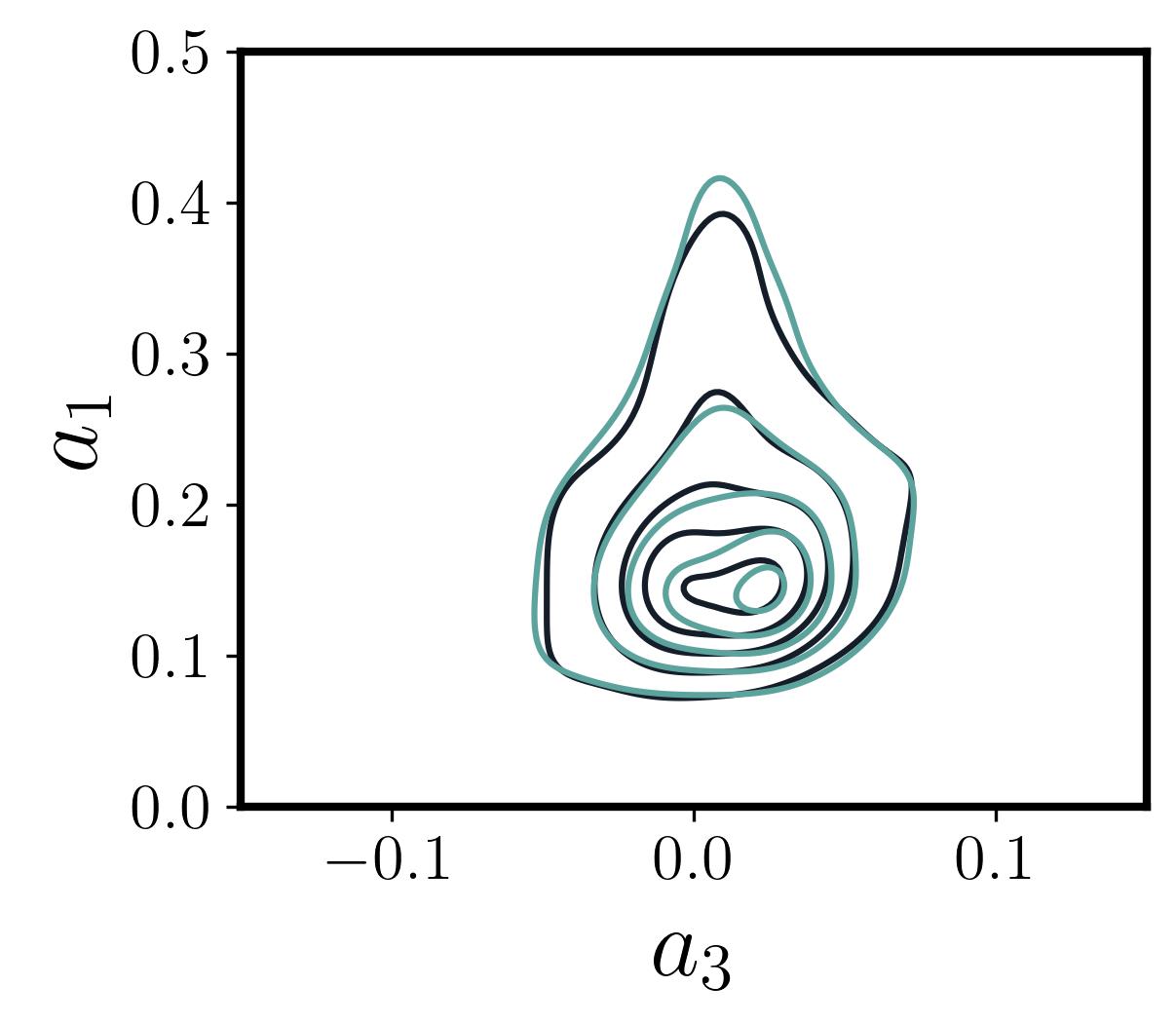}
            \end{subfigure}
            \begin{subfigure}{0.8\linewidth}
                \centering
                \includegraphics[width=\textwidth]{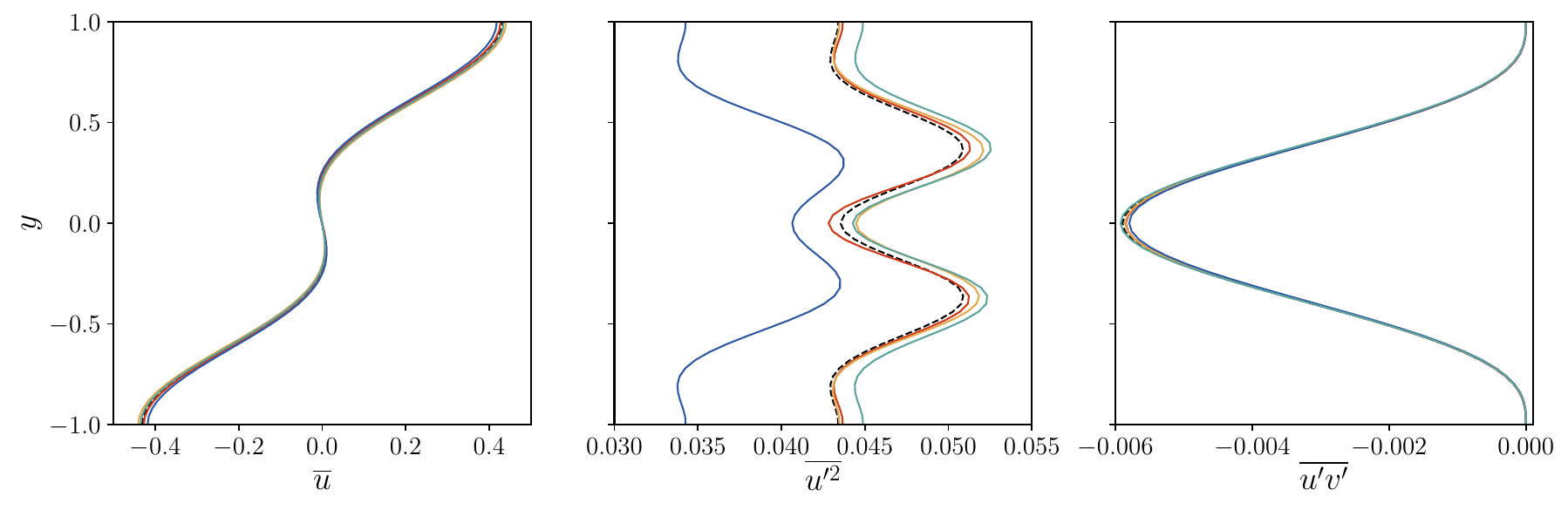}
            \end{subfigure}
            \caption{{\color{black}{\bf Turbulent-flow predictions by the various methods.} (Top) Probability density function of the Poincar\'e maps showing the intersection with $a_2=0$ plane (with ${\rm d} a_2 / {\rm d}t <0)$. In all the cases black denotes the ground truth while the colors correspond to predictions based on (from left to right): easy attention, sparse easy attention, self attention and LSTM. (Bottom) Turbulence statistics corresponding to (left) streamwise mean profile, (middle) streamwise velocity fluctuations and (right) Reynolds shear stress. Colors correspond to the cases on the top panels.}}

\label{9eq_poincare}
\end{figure}

\subsection*{Model of a Nuclear Reactor}
\label{sec:experiments2}
To conclude the assessment of our easy-attention-based model, we will consider a more complex problem of industrial relevance: a model of a nuclear reactor. In particular, we will evaluate the feasibility of using easy attention as a surrogate model to predict the temporal evolution of the state variables in the reactor in order to increase safety and decrease operational costs. The pebble-bed fluoride-salt-cooled high-temperature (PB-FHR) reactor under study here is a Generation-IV nuclear reactor~\cite{locatelli:13} that utilizes a liquid salt coolant, solid fuel pebbles and a graphite moderator. The fluoride-salt coolant has attractive properties that motivate the FHR design both economically and safety-wise, including high boiling point, low operational pressure and high heat-transfer efficiency~\cite{Qualls:17}. We will consider the generic FHR (gFHR), a publicly accessible benchmark model created by Kairos Power~\cite{osti_1868762} which retains the main neutronics and thermal-hydraulics information of their proprietary Karios Power FHR (KP-FHR) model. Data is generated with the System Analysis Module (SAM), a tool developed by Argonne National Laboratory for whole-plant transient analysis of advanced reactor concepts~\cite{hu:21}. The SAM gFHR model considered here was developed using a hybrid method~\cite{Li:22} that combines high-fidelity neutronics simulations and intermediate-fidelity thermal-hydraulics simulations for an entire-plant representation of the reactor configuration. In this study we will focus on predicting the temporal evolution of 13 state variables, and more details regarding the employed datasets can be found in the Supplementary Material in Tabs.~\ref{tab:attn_arch_nuclear} \& \ref{tab:attn_arch_nuclear_lstm}


\noindent All variables are computed by the benchmark model introduced above, considering the latter as the reference.
\begin{figure}[ht]
            \centering
            \begin{subfigure}{0.23\linewidth}
                \centering
                \includegraphics[width=\textwidth]{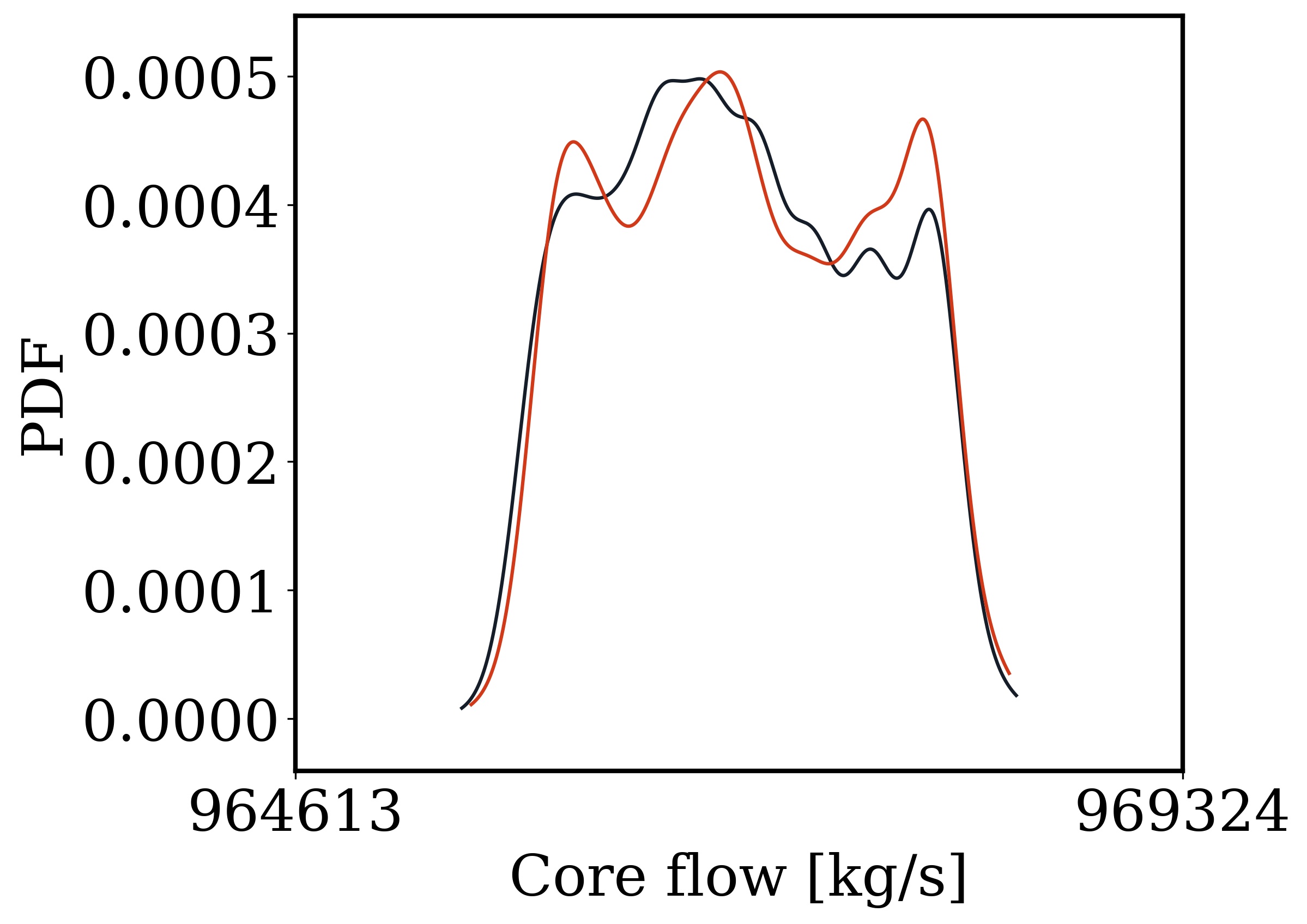}
                
            \end{subfigure}
            \quad
            \begin{subfigure}{0.23\linewidth}
                \centering
                \includegraphics[width=\textwidth]{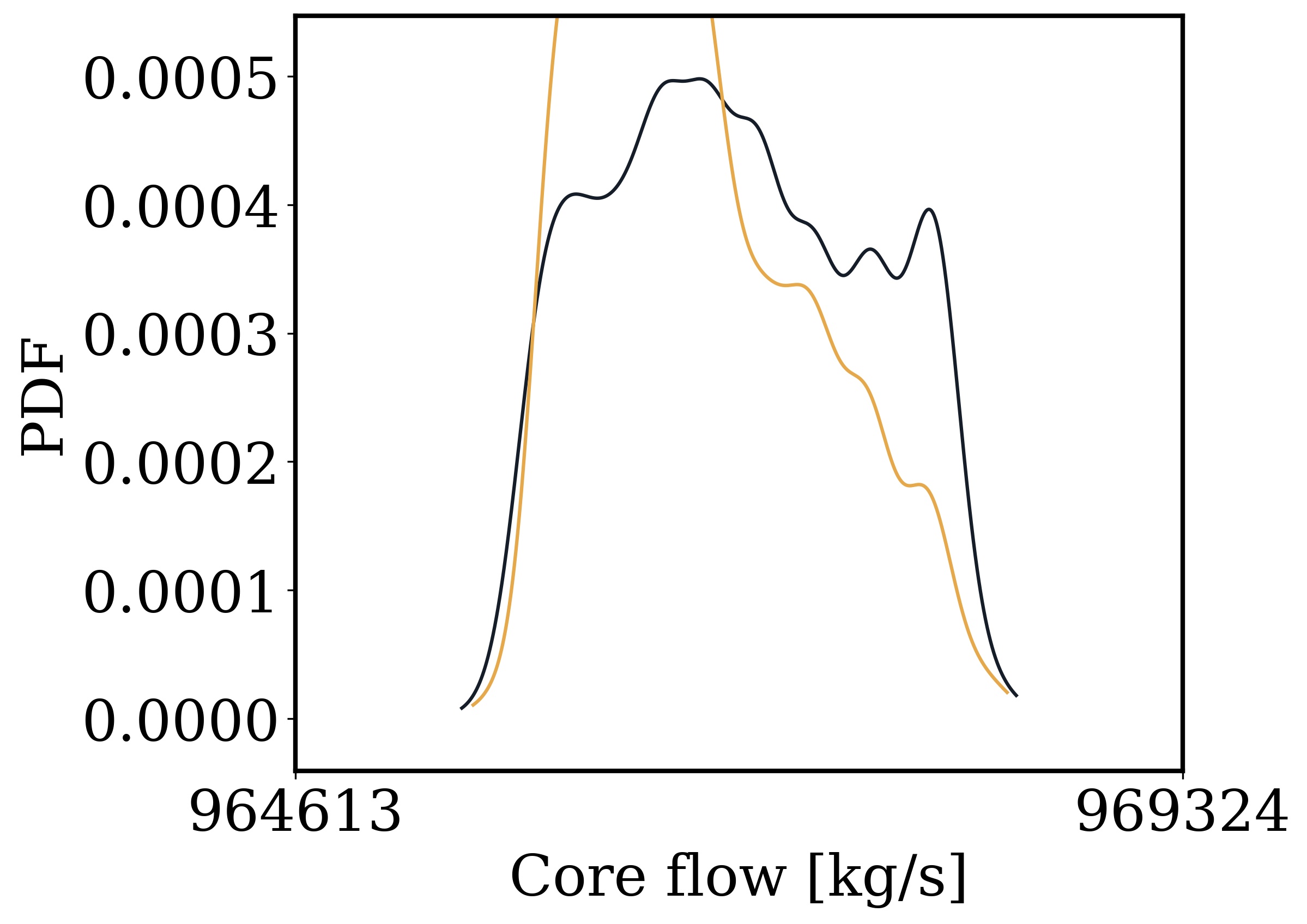}

            \end{subfigure}
            \quad
            \begin{subfigure}{0.23\linewidth}
                \centering
                \includegraphics[width=\textwidth]{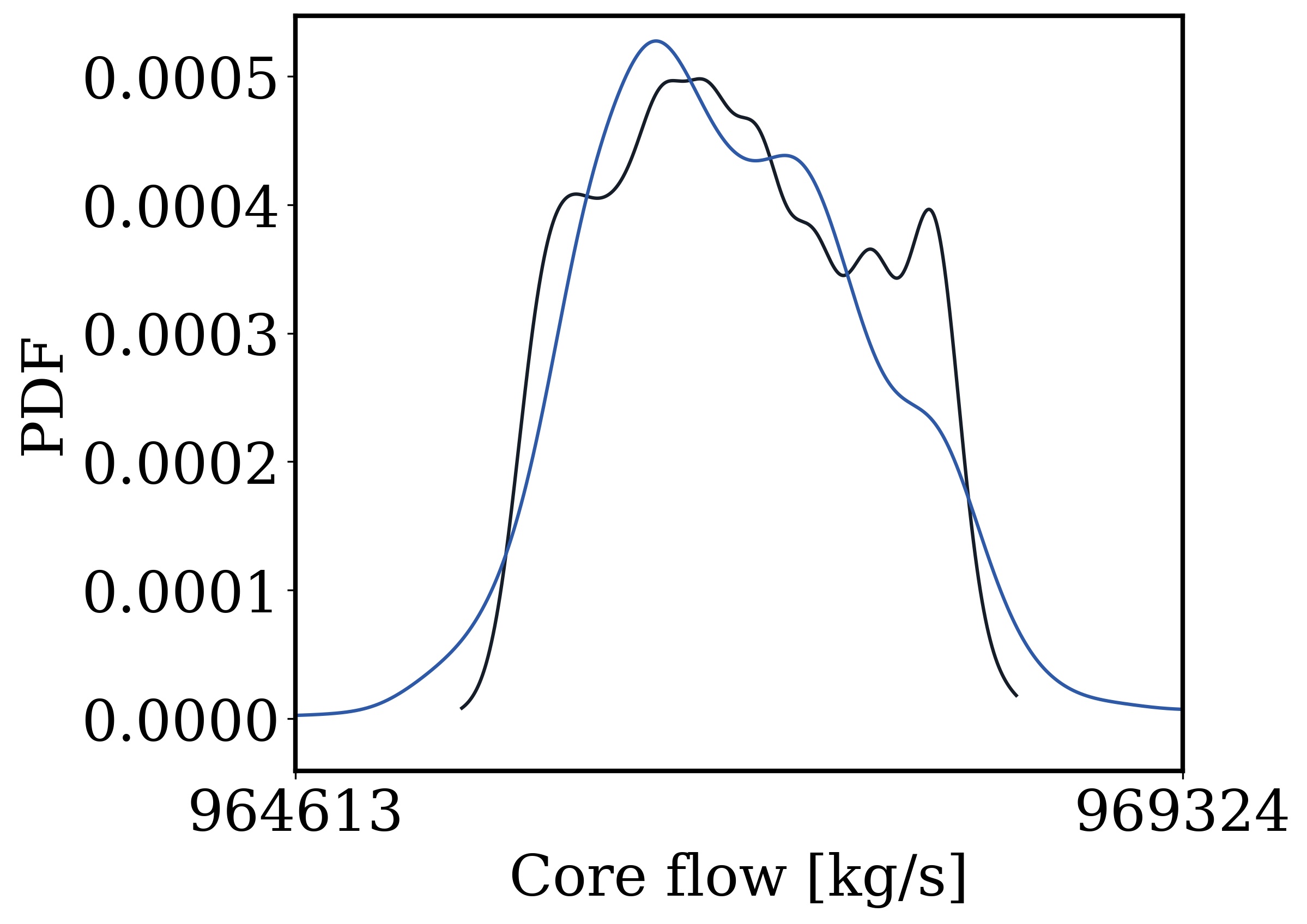}

            \end{subfigure}
            \quad
            \begin{subfigure}{0.23\linewidth}
                \centering
                \includegraphics[width=\textwidth]{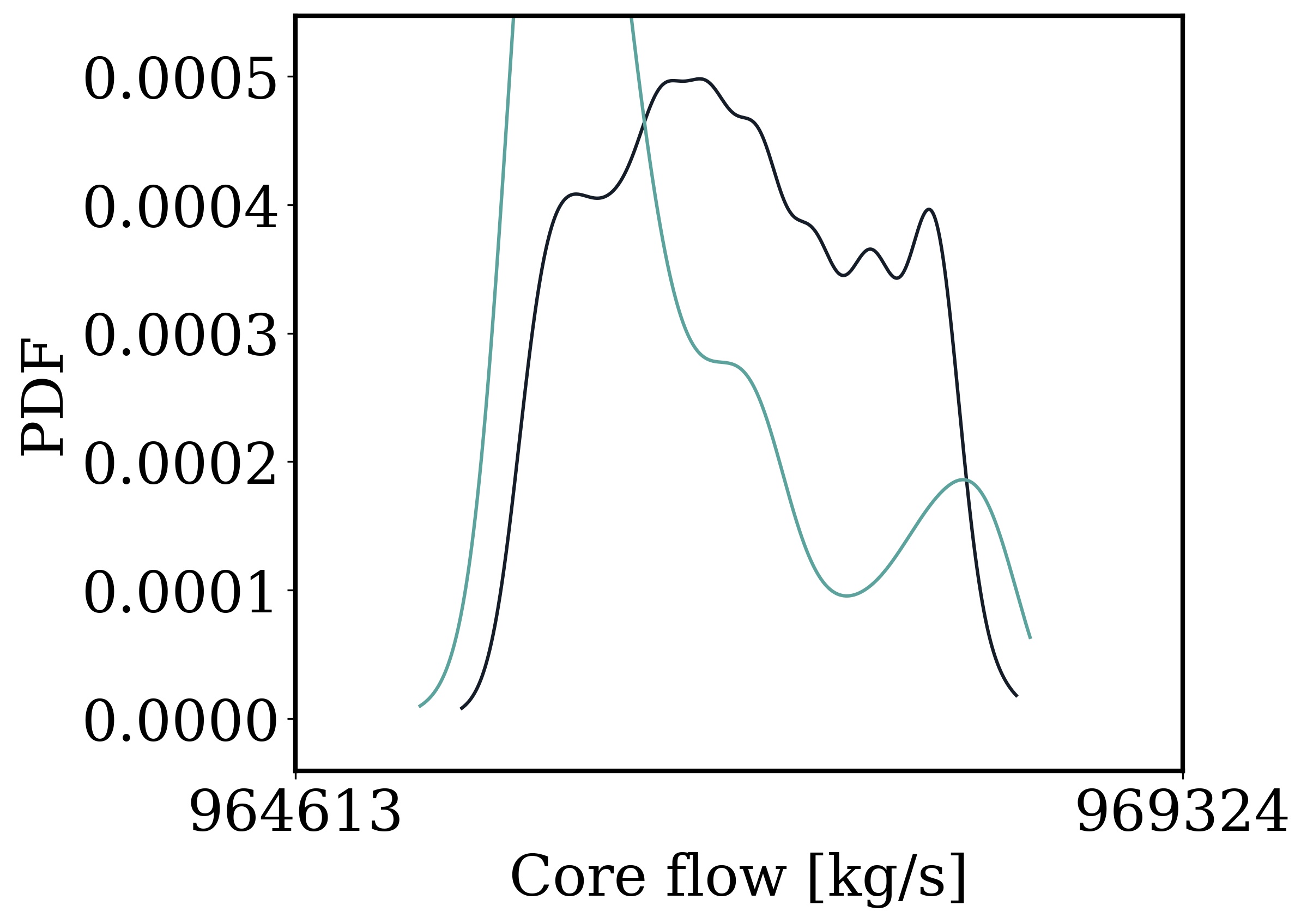}
            \end{subfigure}
            \begin{subfigure}{0.3\linewidth}
                \centering
                \includegraphics[width=\textwidth]{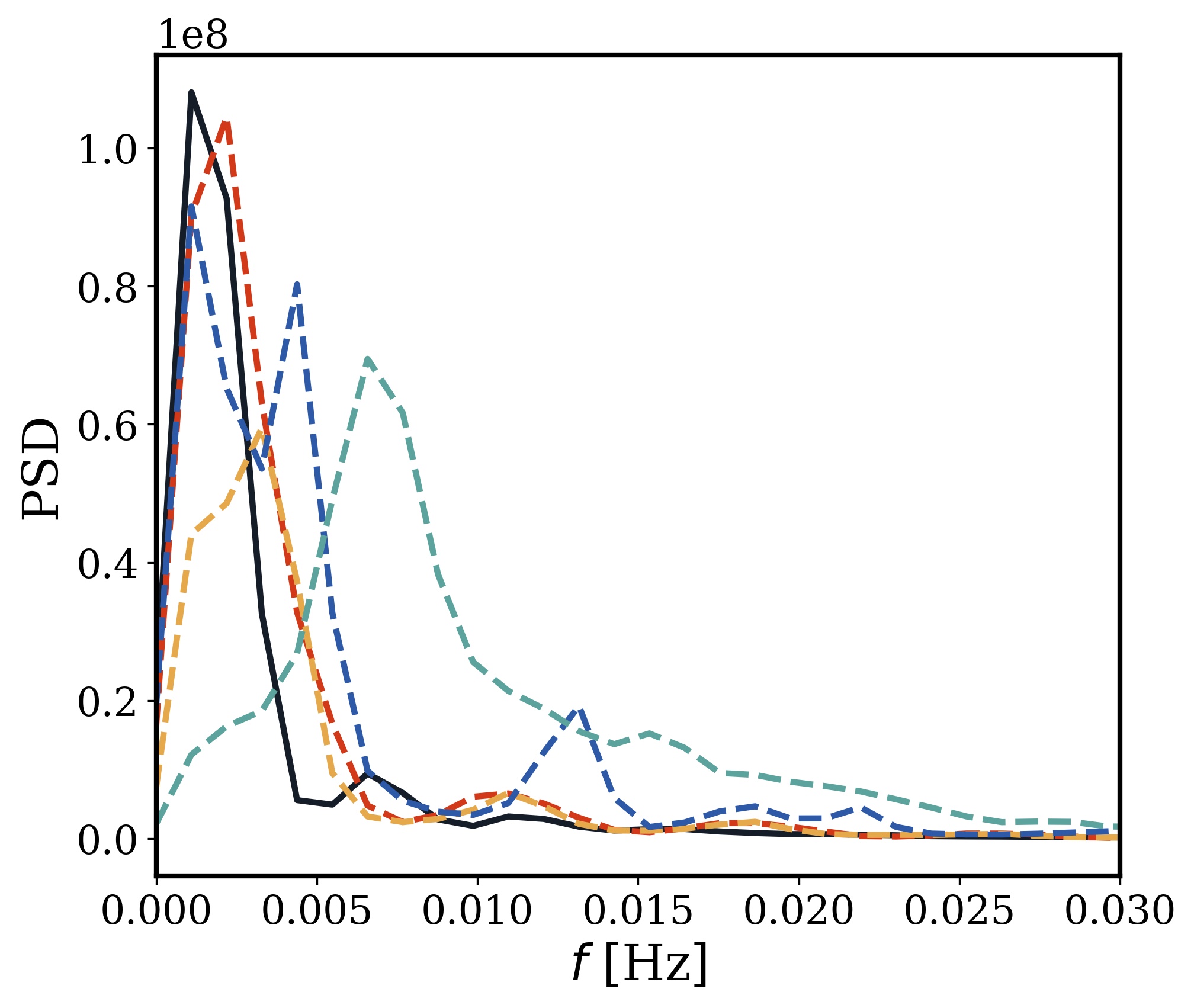}
            \end{subfigure}
            \quad
            \begin{subfigure}{0.3\linewidth}
                \centering
                \includegraphics[width=\textwidth]{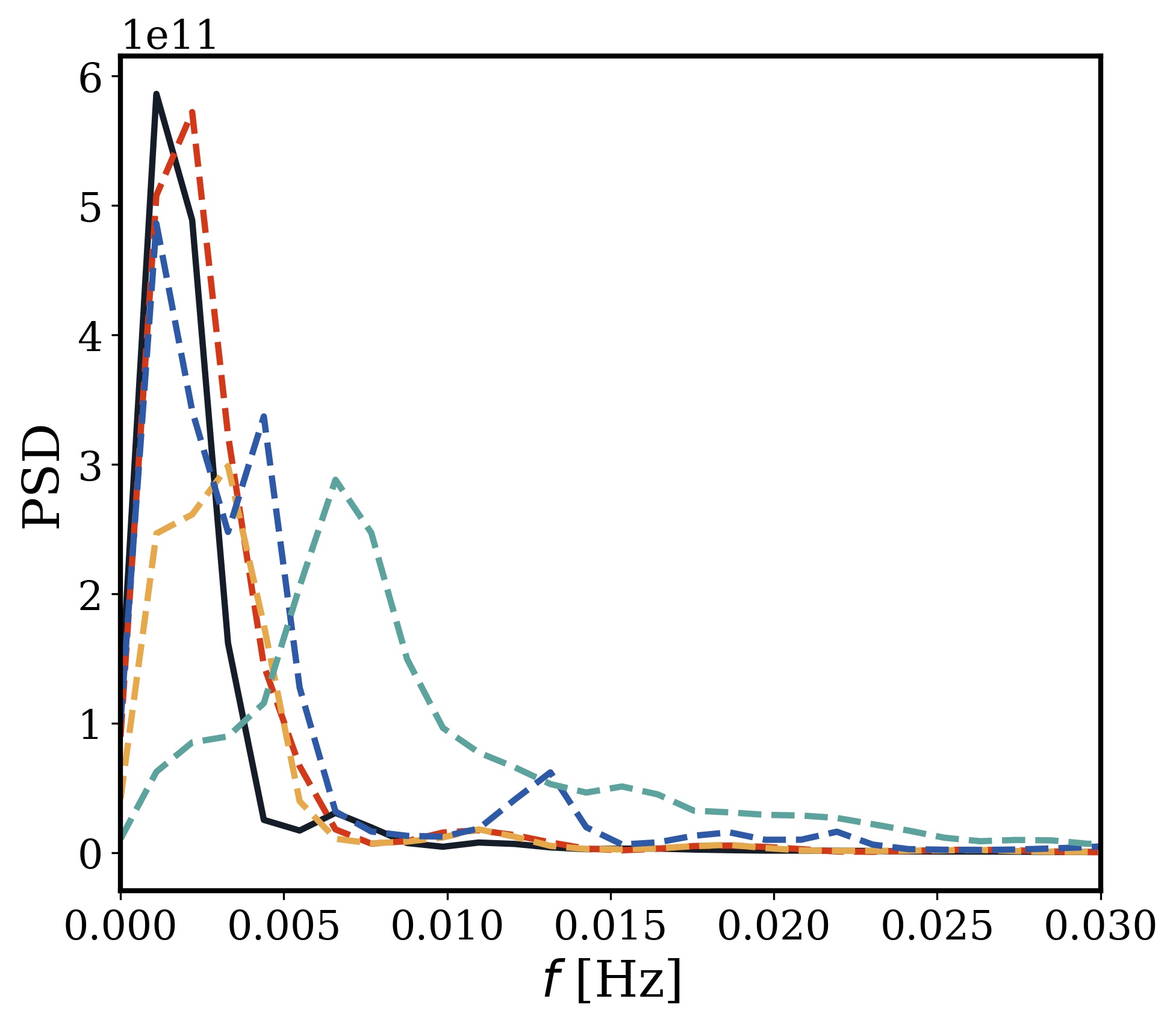}
            \end{subfigure}
            \caption{\color{black}{\bf Characterization of variables in a nuclear reactor reconstructed by the various prediction methods.} (Top) Probability density function of the Core Flow. In all the cases black denotes the ground truth while the colors correspond to predictions based on (from left to right): easy attention, sparse easy attention, self attention and LSTM. (Bottom) Power-spectral density of (left) Core Flow and (right) Core Energy, where $f$ represents frequency. Colors as in the top panels.}
\label{Reactor}
\end{figure}

\noindent In the Supplementary Material we show the temporal evolution for several reactor variables, together with predictions from the various models. In Fig.~\ref{Reactor} (top) we depict the PDF of the Core Flow of the reactor, which is one of the most relevant operation variables in determining reactor performance. The easy-attention model exhibits the best reconstructive capability when comparing with the ground-truth data, while the sparse easy attention significantly over predicts the peaks towards the left and under predicts the ones on the right. On the other hand, the self-attention model can reproduce the peaks reasonably well, but the tails of the distribution surpass the limits of the steady-state distribution. In Fig.~\ref{Reactor} (bottom), we show the power-spectral density for both the Core Flow and the Core Energy. Once again, the easy-attention model exhibits the best predictions considering both the position and magnitude of the predicted spectral peak. The self-attention model yields reasonably accurate results for low frequencies, but it introduces a second peak. Both the LSTM and sparse-easy-attention models lack the level of accuracy exhibited by the other two models, with the sparse-easy-attention yielding a slightly better qualitative agreement with the reference than the LSTM. This behavior highlights the importance of the $\boldsymbol{\alpha}$ matrix in the transformer encoder, as it is responsible for learning temporal dependencies between time instances. 



{\color{black}
\section{Summary and conclusions}
\label{sec:conclusion}
We proposed a novel attention mechanism for transformer neural networks, namely the easy attention, which does not rely on queries and keys to generate attention scores and eliminates the nonlinear $\rm softmax$ function. The easy attention arises from the nonelastic nature of the trace, as this makes impossible any backwards traceability of the information, decreasing the relevance of the exact value of each matrix element. Taking this into account, we propose to learn the attention scores $\boldsymbol{\alpha}$ directly, instead of learning the query and key matrices, removing the $\mathbf{W}_K$, $\mathbf{W}_Q$ and $\rm softmax$ from the attention mechanism, yielding a considerable reduction in the numbers of parameters. Furthermore, we propose the sparse-easy-attention method, which further reduces the number of learnable parameters of the tensor $\boldsymbol{\alpha}$ by only learning the (near-) diagonal elements. We also propose a method to perform multi-head easy attention which employs multiple $\boldsymbol{\alpha}$ to improve the long-term predictions. {Considering chaotic systems as our predictive problem, the invariant nature of attractors and the sensitivity to small perturbations on the initial conditions, suggest that it may be more convenient to use an input-independent model. In future works we will expand the interpretability of the model with operator theory~\cite{Igor}, leveraging ideas from both Koopman Operator Theory and quantum mechanics, specifically the Schr\"odinger equation, where the operator evolving the observable is input independent.}

\noindent We first study sinusoidal functions and we investigate the core idea of self-attention by implementing SVD on the $\mathbf{Q}^T\mathbf{K}$ product and the $\rm softmax$ attention score. The results reveal that the self attention in fact compresses the input sequence transformed covariance martrix. Following this observation, we propose our easy-attention mechanism. For input transform-wave reconstruction, the easy-attention model achieves 0.0018\% $l_2$-norm error with less computation time for training and complexity than self attention, which exhibits an $l_2$-norm error of 10\%.  Motivated by the promising performance on sinusoidal-wave reconstruction, we propose the multi-easy-attention method, which combines the discrete Fourier transform with attention mechanism for signal reconstruction. The unique capability of easy attention to learn long-term dynamics for periodic signals suggests that this method may have great potential for additional applications, such as signal de-noising.

\noindent Subsequently, we apply easy attention in the proposed transformer model for temporal-dynamics prediction of chaotic systems and we investigate the performance of our models on the Lorenz system. The transformer model with dense easy attention yields the lowest relative $l_2$-norm error of 2.0\%, outperforming the same transformer architecture with the self attention (7.4\%) and the LSTM model (37.7\%). Furthermore, when inspecting the nine-equation model of wall-bounded turbulence by Moehlis {\it et al.}~\cite{moehlis2004low} the long-term statistics discussed in Fig.~\ref{9eq_poincare} show how LSTM yields very similar performance in the mean flow compared with the easy attention. Regarding the PDF of the Poincar\'e section shown in Fig.~\ref{9eq_poincare}, the easy attention also exhibits better predictions than the other models, especially when considering the symmetries observed in Fig.~\ref{fig:lorentz_pdf} (top), which highlight the model’s ability to replicate invariant structures. Therefore, the easy-attention method emerges as the most accurate model when studying chaotic systems. Regarding the studied model of a nuclear reactor, the easy attention shows the best reconstructive capability when considering PSD and PDF analysis. The results also show that, compared with the self-attention-based transformer, the transformers using easy attention have a much lower computational cost, exhibit less complexity and have less parameters. Thus the transformer model using easy attention is more suitable for temporal-dynamics predictions.

\noindent Furthermore, a {\color{black}significant advantage} of the easy-attention-based transformer is its accurate reconstruction of the invariant sets, as shown in Figs.~\ref{fig:lorentz_pdf}, \ref{Reactor}. These invariant sets arise when inspecting the systems independently of time, as the latter converge to a stationary probability density function and a fixed attractor in state space. The PDF is adequately reconstructed and the information density per neighborhood is accurately represented by the easy-attention model. In uniformly hyperbolic dynamical systems, the shadowing lemma guarantees that any learned or numerical \emph{pseudo-orbit} given by a sequence $\{x_k\}$ satisfying:
\[
\|x_{k+1} - \Phi_{\Delta t}(x_k)\|\le\varepsilon
\quad\text{for }k=0,\dots,N-1,
\]
with small $\varepsilon>0$, remains uniformly close to a true trajectory: there exists an exact orbit $\{\tilde x_k\}$ such that 
\[
\|\tilde x_k - x_k\|\le C\,\varepsilon
\quad\text{for all }k,
\]
where $C>0$ depends on the system’s hyperbolicity constants~\cite{Pilyugin:1999,Katok:1997}. 
Hence, even when our transformer’s predictions diverge from the reference due to sensitive dependence on initial conditions, they form a valid pseudo-orbit that is shadowed by an actual orbit.  This result justifies using the transformer’s output to recover statistically and geometrically meaningful features—such as invariant measures and attractor geometry—of the underlying chaotic system.  

\noindent In conclusion, the easy-attention mechanism, due to its robustness, exhibits promising performance in the tasks of signal/trajectory reconstruction as well as temporal-dynamics prediction in chaotic systems. Our results also indicate potential for application in more complex large-scale dynamical systems, since transformer networks are naturally designed for high-dimensional sequence data~\cite{non_instruct_pod_transformer,geneva2022transformers,solera2023beta,cao2021choose, katharopoulos2020transformers_rnn}.

\section{{Methods}}
\label{sec:easy-attn}

In this section, we will focus on the mathematical development of the attention part of the model and its implications on the final output, since the other blocks follow the standard transformer architecture~\cite{vaswani2017attention}.
To this end we first examine the self-attention model, then explain the easy-attention mechanism in detail.
The mathematical implementation {of} a transformer encoder block is expressed as follows~\cite{John}:
a transformer block {is a} parameterised function $T:\mathbb{R}^{n\times d} \rightarrow \mathbb{R}^{n\times d}$, usually consisting of an attention part and a feed forward-network, both with residual connections and normalization layer, see the transformer block used for each case in the Supplementary Material.

\subsection*{Notation}
Adhering to the customary notation, given any matrix ${\bf A} \in \mathbb R^{n \times d}$, $[{\bf A}]_{ij}$ will denote the element on the $i$-th row and $j$-th column of ${\bf A}$ and given a vector ${\bf a} \in \mathbb R^d$, $[{\bf a}]_j$ will denote its $j$-th component.
We represent by lowercase ${\bf a}_i$ the $i$-th row of any matrix ${\bf A}$.
As it is usual, we will interpret all vectors, such as ${\bf a}_i \in \mathbb R^d$, as column vectors.
Thus, assuming a concatenation by columns and wrote formally, ${\bf A} = \left( \text{concat} \{ {\bf a}_i \}_{i=1}^n \right)^T$. In the following we will assume that ${\bf X} \in \mathbb R^{n \times d}$ is the input of the model, composed by the past $n$ observations of $d$ features each. For example, in the Lorenz case $n = d = 64$ (after the time embedding).

\subsection*{Attention}
\label{sec:easy_attn}
Let us focus on the attention sub-block, which is in turn another transformation $\mathbb{R}^{n\times d} \rightarrow \mathbb{R}^{n\times d}$, $\mathbf{X} \mapsto \hat{\mathbf{X}}$.
That mapping, applied to each row of $\mathbf{X}$, takes the following form:
\begin{equation}
	\mathbf{\hat{x}}_i = \sum^{n}_{j=1}\alpha_{i,j}{\mathbf{W}_V}^T\mathbf{x}_j.
	\label{eq:attn}
\end{equation}
The matrix ${\mathbf{W}_V}$ applies the same linear transformation into the input $\mathbf{x}_i$.
Denoting as $\boldsymbol{\alpha} \in \mathbb R^{n \times n}$ the matrix formed by the $\alpha_{ij}$ values, {\it i.e.} $[\boldsymbol{\alpha}]_{ij} = \alpha_{ij}$ and $\boldsymbol{W_V} \in \mathbb R^{d \times k}$ where $k=d$ for our study:
\begin{align*}
	[{\bf \hat X}]_{ik}
	& = [{\bf \hat x}_i]_k 
	= \sum_{j=1}^n \alpha_{ij} [{\mathbf{W}_V}^T {\bf x}_j]_k
	= \sum_{j=1}^n \sum_{l=1}^n \alpha_{ij} [{\mathbf{W}_V}^T]_{kl} [{\bf x}_j]_l \\
	& = \sum_{l=1}^n [{\mathbf{W}_V}^T]_{kl} [\boldsymbol{\alpha} \cdot {\bf X}]_{il}
	= \sum_{l=1}^n [\boldsymbol{\alpha} \cdot {\bf X}]_{il} [{\mathbf{W}_V}]_{lk}
	= [\boldsymbol{\alpha} \cdot {\bf X} \cdot {\mathbf{W}_V}]_{ik}.
\end{align*}
Therefore the model can be written in a matrix form as:
\begin{equation}
	{\bf \hat X} = \boldsymbol{\alpha} {\bf X}  {\mathbf{W}_V}.
	\label{eq:attention_develop}
\end{equation}
In Eq.~(\ref{eq:attention_develop}) it can be clearly seen that${\mathbf{W}_V}$ and $\boldsymbol{\alpha}$ are a spatial and a temporal transformation, respectively.

\subsection*{Self-attention}
In the standard implementation of the transformer~\cite{vaswani2017attention} the attention is computed using a scaled dot-product attention as follows.
Given an input $\mathbf{X}$, the query, key and value of each observation are computed respectively as
\begin{equation}
	\mathbf{Q}(\mathbf{x}_i) = \mathbf{W}_Q^T \mathbf{x}_i, \quad
    \mathbf{K}(\mathbf{x}_i) = \mathbf{W}_K^T \mathbf{x}_i, \quad
    \mathbf{V}(\mathbf{x}_i) = \mathbf{W}_V^T \mathbf{x}_i.
	\label{eq:QKV_xi}
\end{equation}
These quantities are constructed based on three different weight matrices: $\mathbf{W}_Q, \mathbf{W}_K, \mathbf{W}_V \in \mathbb{R}^{k \times d}$.
These matrices project each observation into a vector in a latent space of dimension $k$ (at the moment assume $k = d$) and are all learned during training. The attention score $\alpha_{i,j}$, that correlates the observation $\mathbf{x}_i$ with $\mathbf{x}_j$, is computed as:
\begin{equation}
	\alpha_{i,j} = {{\rm softmax}}\left(\frac{\langle \mathbf{Q}(\mathbf{x}_i), \mathbf{K}(\mathbf{x}_j) \rangle}{{\sqrt{k}}}\right);
	\quad {\rm softmax}(\mathbf{z}_j) = \frac{{\rm exp}(\mathbf{z}_j)}{\sum_{k=1}^n {\rm exp}(\mathbf{z}_k)},
 \end{equation}
in which ${\rm softmax}$ is applied to each row of $\boldsymbol{\alpha}$.
That well-known activation function is a generalization of the logistic function used for compressing {$d$-dimensional} vectors of arbitrary real entries into real vectors of the same dimension in the range $[0,1]$, with normalized $l_1$-norm.
Once the attention scores are computed, using Eq. \eqref{eq:attn} the output of the module is obtained.

\noindent It is possible represent the product ${\mathbf{W}_V}^T \mathbf{x}_i$ (assuming $\mathbf{W}_V$ to be square, as discussed below) as a linear combination of the eigenvectors and eigenvalues {($\mathbf{v}_k, \lambda_k$)} of $\mathbf{W}_V$ for unique coefficients $c_{k,j}$ depending only on $\mathbf{x}_j$, resulting on:
\begin{equation}
	\mathbf{\hat{x}}_i = \sum_{j=1}^{n}\alpha_{i,j}\mathbf{V}(\mathbf{x}_j)
	= \sum_{j=1}^{n} \alpha_{i,j} \left( \sum_{k=1}^n \lambda_k \mathbf{v}_k c_{k,j} \right).
\end{equation}

\noindent For analytical purposes in Fig.~\ref{fig:Evo} we will express the inner product inside the {$\rm softmax$} operation in terms of the trace so it is possible to take advantage of its cyclic property:
\begin{equation}
	\alpha_{i,j} = {\rm softmax} \left(\frac{\rm{Tr}(\mathbf{W}_Q \mathbf{W}_K^T \mathbf{x}_j \mathbf{x}_i ^T)}{ \sqrt{k}}\right).
\end{equation}
\noindent In this study we asses the importance of the weight matrices for the accuracy of the predictions and note that the result depends on the eigenvalues of $\mathbf{W}_Q \mathbf{W}_K^T \mathbf{x}_j \mathbf{x}_i ^T$. Given the nature of the trace, introduced in the interpretation of the attention mechanism, the transformer only uses the final result of the sum. For example: let $\mathbf{Z} \in \R^{n \times n} \Rightarrow {\rm Tr}(\mathbf{Z})= a+b+c=d$, where $a,b,c,d\in \mathbb{R}$, and in principle there are infinite possible values of $a$, $b$, $c$ such that the sum is equal to $d$.

Note that in the general attention mechanism the query, key and value can be computed using different input observations, $\mathbf{Q}(\mathbf{x}_i)$, $\mathbf{K}(\mathbf{y}_i)$, and $\mathbf{V}(\mathbf{z}_i)$.
This is the case of, for example, translation models in which the original words are compared with the translated ones, leading to an attention matrix that correlates the original and translated words (observations), see Ref.~\cite{vaswani2017attention}.
In our temporal prediction case, the only relevant information is the past observations, so the values in Eq.~\eqref{eq:QKV_xi} are all computed for the same observations $\mathbf{x}_i$.

\subsection*{Easy attention}
In the easy-attention mechanism, the key and query of self attention are eliminated, and~\eqref{eq:attn} is computed directly.
We directly learn $\boldsymbol{\alpha} \in \mathbb{R}^{n \times n}$ as a learnable attention-score tensor of the spanned eigenspace of the {multiplication of the time-delay} input matrix by its transpose.
As consequence, only $\boldsymbol{\alpha}$ and $\mathbf{W}_{V}$ are required to be learned.
In particular, an input $\mathbf{x}$ is projected as the value tensor $\mathbf{V}$ by $\mathbf{W}_V$. Subsequently, we compute the product of $\mathbf{V}$ and the attention scores $\alpha_{i,j}$ as the output $\mathbf{\hat{x}}_i$. 
The matrix formula version in Eq.~\eqref{eq:attention_develop}, together with the fact that the attention scores are constant, suggest that $\hat {\mathbf{X}}$ is a rewriting of $\mathbf{X}$ choosing another basis (ideally, the best) for time ($\boldsymbol{\alpha}$) and features (${\bf W_V}$). 

As the attention scores are fixed once trained, they remain constant during the prediction process, as if the values $\langle \mathbf{Q}(\mathbf{x}_i), \mathbf{K}(\mathbf{x}_j) \rangle$ only depend on $i$ an $j$ but not on the inputs $\mathbf{x}_i$ and $\mathbf{x}_j$.
This non-trivial assumption, which may not be applicable to scenarios where the inputs could be presented in different orders, is powerful in time-series prediction, since the information is always provided in the same way: the (ordered) past observations. However, considering time series sampled with a non-uniform time step, seems reasonable not to consider the inner product as measure for quantifying correlation, since the distance in the real vector space between vector elements does not consider the different sampling rate of the given elements in the state space neither the rate of information propagation. Moreover, as the order of the inputs $\mathbf{x}_i$ is always the same, an encoding of the position of each $\mathbf{x}_i$ is not needed, enabling the use of embedding instead.

\noindent Furthermore, $\boldsymbol{\alpha}$ can be sparsified by only learning the diagonal elements, or adding elements in off-diagonal positions both above and below the main diagonal if necessary, which further reduces the number of learnable parameters of the easy-attention mechanism. At the stage of forward propagation or inference, those parameters are assigned to the corresponding indices in a matrix of dimension $\mathbb{R}^{n \times n}$. At the Supplementary material one can find illustrations to further differentiate between the dense and sparse easy-attention methods. In the present study, we only consider learning main-diagonal elements as sparsification.

\subsection*{Multi-head strategies}
To conclude the study, we will introduce the implementation of the easy attention with multi-head strategies.
Assume now that the attention mechanism is computed $h$ times and the concatenated, as follows:
Then,
\begin{equation}
	{\bf \hat X} = \text{concat} \left\{ \boldsymbol{\alpha}^l {\bf X} {\bf W}_{V}^l \right\}_{l=1}^h.
	\label{eq:multi-head_formula}
\end{equation}
\noindent Note that it is common to consider a smaller dimension on the value space: ${\bf W}_{V}^l \in \mathbb R^{k \times d}$, while $\boldsymbol{\alpha}^l$ remains to be in $\mathbb R^{n \times n}$.
The matrix with all the values, ${\bf V}^l({\bf X}) = {\bf X} \cdot {\bf W}_{V}^l$, which needs to be determined for each head, can be computed for all of them together as ${\bf V}({\bf X}) = \text{concat} \{ {\bf V}^l({\bf X}) \}_{l=1}^h = {\bf X} {\bf W}_V$, where ${{\bf W}_V} = \text{concat} \{ {{\bf W}_{\bf V}^l} \}_{l=1}^h$, and then, ${\bf \hat X} = \text{concat} \{ \boldsymbol{\alpha}^l \cdot {\bf V}^l({\bf X}) \}_{l=1}^h$.
See the Supplementary Material for additional details.
This allows to consider only one ${\bf W}_V$ square matrix as in the single-head case.

\noindent In the most general case of multi-head self-attention, another matrix ${\bf W}_O \in \mathbb{R}^{k \cdot h \times d}$ is added to~\eqref{eq:multi-head_formula}, as ${\bf \hat X} = \text{concat} \{ \boldsymbol{\alpha}^l {\bf X}  {\bf W}_{V}^l \}_{l=1}^h  {\bf W}_{0}.$
This process is also omitted in the easy attention case.

\noindent Compared with the previous one-head formula, it can be observed that any modification on the features (right-matrix multiplication by ${\bf W}_{\bf V}^T$) remains the same, but the time transformation (left-matrix multiplication by ${\boldsymbol{\alpha}}$) now depends also on the (group of) feature variables.
The multi-head easy attention is beneficial to preserve long-term dependencies when $d$ becomes large. The mechanisms of single- and multi-head easy attention are illustrated in the Supplementary Material. }

\subsection*{Operator‐Theoretic Connections: From Self‐Attention and POD/DMD to Easy‐Attention}

We now provide a self‐contained, rigorous account of how (i) self‐attention, POD/DMD, and EDMD each approximate the Koopman operator; (ii) they suffer finite‐data or nonlinear limitations; and (iii) Easy‐Attention provides the optimal finite‐data approximation.

\medskip

\paragraph*{Notation.}
We use the following symbols throughout this section:
\begin{itemize}
  \item $d$: dimension of each state vector $y(t)\in\R^d$.
  \item $p$: number of delay coordinates (history length).
  \item $N$: number of training samples (pairs of delay‐embedded vectors and their one‐step–ahead targets).
  \item $k$: latent projection dimension in self‐attention ($\bf W_Q,W_K,W_V\in\R^{d\times k}$).
  \item $\Delta t$: time lag between successive observations in the embedding.
  \item $\kappa$: sub‐Gaussian norm bound of the random vectors $y_j$.
  \item $\delta\in(0,1)$: failure‐probability parameter in concentration bounds.
  \item $r$: half‐width of the sparsity band in sparse Easy‐Attention.
  \item $\varphi_i(x(t)) = x\bigl(t-(i-1)\Delta t\bigr)$: the $i$‑th delay coordinate, and $\varphi = (\varphi_1,\dots,\varphi_p)^T$.
  \item $\dagger$: Moore–Penrose pseudoinverse of a matrix.
  \item $\bf C = \E[y\,y^T]$: true covariance of the embedding vectors.
  \item ${\bf \widehat C} = 1/p\sum_{j=1}^p y_jy_j^T$: empirical covariance.
\end{itemize}

\medskip

\paragraph{1. Self‐Attention as a Dynamic Covariance Surrogate.}
Let \(p\) be the history length (number of past time steps) and \(d\) the dimension of each state vector. Given a history of delay‐embedded vectors: 
\[
{\bf Y} = 
\begin{bmatrix}
y(t_1) \\ y(t_2) \\ \vdots \\ y(t_p)
\end{bmatrix}
\in\R^{p\times d},
\]
and learnable projections \(\bf W_Q,\bf W_K,\bf W_V\in\R^{d\times k}\), the single‐head self‐attention computes:
\[
\bf Q = \bf Y\bf W_Q,\quad K = Y\bf W_K,\quad V = Y\bf W_V,
\quad
A = \text{softmax}\!\bigl(QK^T/\sqrt{k}\bigr),
\quad
\widehat Y = AV.
\]
Since \(\bf QK^T = Y\,(\bf W_Q\bf W_K^T)\,Y^T\) is a rotated empirical covariance, \(\bf A\) is its softmax‐normalized form.  This yields contextual adaptivity and an implicit nonlinear projection, but introduces sampling noise (see Theorem~\ref{thm:cov-conc} below), softmax distortion, and \(\mathcal O(p^2d)\) computational cost.

\medskip

\begin{theorem}[Covariance Concentration \& Attention Deviation]
\label{thm:cov-conc}
Let \(\{y_j\}_{j=1}^p\subset\R^d\) be i.i.d.\ sub‐Gaussian with zero mean, covariance \(C\), and sub‐Gaussian norm bounded by \(\kappa\).  Define the empirical covariance:
\[
{\bf \widehat C} = \frac1p\sum_{j=1}^p y_jy_j^T,
\]
and let 
\[
{\bf A} =  \softmax_{\mathrm{row}}\!\bigl(({\bf Y}\,{\bf W_Q})({\bf Y}\,{\bf W_K})^T/\sqrt{k}\bigr),
\quad
{\bf A_\infty} = \softmax_{\mathrm{row}}\!\bigl(({\bf Y}\,{\bf W_Q})({\bf Y}\,{\bf W_K})^T/\sqrt{k}\bigr)\Big|_{{\bf Y^T Y = p\,C}}.
\]
Then for any \(\delta\in(0,1)\), with probability at least \(1-\delta\),
\[
\|\widehat {\bf C} - {\bf C}\|_2 
\;\le\;
2\,\kappa^2\Bigl(\sqrt{\tfrac{d+\ln(2/\delta)}{p}}
                  +\tfrac{d+\ln(2/\delta)}{p}\Bigr),
\]
and since row‐wise softmax is 1‐Lipschitz,
\[
\|{\bf A - A_\infty}\|_2
\;\le\;\frac{\|\bf W_Q\bf W_K^T\|_2}{\sqrt{k}}\;\|\widehat {\bf C} - {\bf C}\|_2.
\]
\end{theorem}

\begin{proof}[Sketch]
Apply Tropp’s noncommutative Bernstein inequality~\cite{tropp2015introduction} to the zero‐mean deviations \(Z_j=y_jy_j^T - C\) to bound \(\|\widehat {\bf C} - {\bf C}\|_2\).  The Lipschitz property of row‐wise softmax follows since each exponential is 1‐Lipschitz and normalization preserves this constant.
\end{proof}

\medskip

\paragraph{2. POD and DMD as Linear Surrogates.}
Proper-Orthogonal Decomposition computes the eigenmodes of:
\[
{\bf C_{\rm emp}} = \frac1N {\bf X X^T},\qquad {\bf X}=[y(t_1),\dots,y(t_N)]\in\R^{d\times N},
\]
via \(C_{\rm emp} u_i = \lambda_i u_i\).  Dynamic-Mode Decomposition then fits:
\[
{\bf Y' \approx A_{\rm DMD}\,X,\qquad A_{\rm DMD} = Y'X^\dagger},
\quad
{\bf Y'} = [\,y(t_1+\Delta t),\dots,y(t_N+\Delta t)\,].
\]
Both assume the dynamics lie in a linear subspace, incurring large projection error for strongly nonlinear or chaotic systems.

\medskip

\paragraph{3. Extended DMD (EDMD) via Delay Embeddings.}
To approximate the infinite‐dimensional Koopman operator:
\[
(\mathcal K_{\Delta t}g)(x)=g\bigl(\Phi_{\Delta t}(x)\bigr),
\]
for states \(x(t)\in\R^d\), we form the \emph{delay embedding}
\[
\varphi\bigl(x(t)\bigr)
=\begin{bmatrix}
x(t) \\ x(t-\Delta t) \\ \vdots \\ x(t-(p-1)\Delta t)
\end{bmatrix}
\;\in\;\R^{p\,d},
\]
and collect:
\[
{\bf X} = [\,\varphi(x(t_1)),\dots,\varphi(x(t_N))\,]\in\R^{p\,d\times N},
\quad
{\bf Y} = [\,\varphi(x(t_1+\Delta t)),\dots,\varphi(x(t_N+\Delta t))\,].
\]

\begin{theorem}[EDMD Equivalence~\cite{williams2015data,klus2016koopman}]
The minimizer
\[
 \boldsymbol{ \alpha^*} 
=\arg\min_{\alpha\in\R^{p\times p}}\|\bf Y -  \boldsymbol{\alpha\,}X\|_F^2
\;=\;\bf Y\,X^\dagger
\]
is exactly the EDMD approximation of \(\mathcal K_{\Delta t}\) on the subspace:
\[
\mathrm{span}\{\varphi\}
=\Bigl\{\,f:\R^n\to\R\;\Bigm|\;f(x)=\sum_{i=1}^p c_i\,\varphi_i(x),\;c_i\in\R\Bigr\}.
\]
\end{theorem}

\begin{corollary}[Convergence under Mild Conditions~\cite{takens1981detecting,mezic2005spectral}]
If (i) the trajectory is ergodic, (ii) \(\varphi\) satisfies Takens’ embedding (\(p>2n\)), and (iii) \(X\) has full row rank as \(N\to\infty\), then:
\[
 \boldsymbol{ \alpha^*\;}\longrightarrow\;\mathcal {\bf K}_{\Delta t}\big|_{\mathrm{span}(\varphi)}
\quad\text{in operator norm as }N\to\infty.
\]
\end{corollary}

\medskip

\begin{proposition}[Optimal Finite‐Rank Operator via Least Squares]\label{prop:ls-op}
Let \(V(y)=y\bf W_V\) with \(\bf W_V\in\R^{d\times k}\), and define the one‐step prediction loss:
\[
\mathcal{L}( \boldsymbol{\alpha})
=\E\bigl\| \boldsymbol{\alpha\,}\bf V(y)\;-\;V\bigl(\Phi_{\Delta t}(x)\bigr)\bigr\|_2^2.
\]
If \(\E[\bf V(y)V(y)^T]\) is invertible, then the unique minimizer is
\[
 \boldsymbol{\alpha^*}
=\E\bigl[\bf V\bigl(\Phi_{\Delta t}(x)\bigr)\,V(y)^T\bigr]\,
 \bigl(\E[V(y)V(y)^T]\bigr)^{-1}.
\]
\end{proposition}

\begin{proof}
Setting \(\nabla_\alpha\mathcal{L}=0\) yields the normal equation 
\ \(\alpha\,\E[\bf V(y)V(y)^T]=\E[V(\Phi_{\Delta t}(x))V(y)^T]\), whose unique solution is the stated \(\alpha^*\).
\end{proof}

\medskip

\paragraph{4. Easy‐Attention: Optimal Finite‐Data Approximation.}
Easy‐Attention discards \(\bf W_Q,\bf W_K\) and the softmax, and directly learns:
\[
(\boldsymbol{\alpha^*},\bf W_V^*)
=\arg\min_{\alpha\in\R^{p\times p},\,\bf W_V\in\R^{p\times d}}
\bigl\|\,Y - \boldsymbol{\alpha\,}X\,\bf W_V\bigr\|_F^2.
\]
By Proposition~\ref{prop:ls-op}, the resulting $\boldsymbol{(\alpha^*)}$ is the unique least‐squares finite‐rank operator that minimizes one‐step prediction error for the chosen dictionary, with sampling variance \(O(1/N)\) rather than \(O(1/\sqrt{N})\).

\medskip

\paragraph{5. Sparse Easy‐Attention.}
To reduce both parameter count and inference cost, we can impose a banded sparsity pattern on \(\alpha\):
\[
\alpha_{ij}\neq0\quad\Longleftrightarrow\quad |i-j|\le r,\quad r\ll p.
\]
Inference then costs \(O(p\,r\,d)\) rather than \(O(p^2d)\), while retaining the same finite‐data optimality guarantees of Proposition~\ref{prop:ls-op} and greatly lowering memory footprint.

\medskip

\paragraph{Computational Efficiency.}
\begin{itemize}
  \item {\bf Self‐Attention:} 
    \(O(p\,d\,k)+O(p^2k)+O(p^2)+O(p^2d)\approx O(p^2d)\).
  \item {\bf Easy‐Attention:}
    \(O(p^2d)+O(p\,d\,k)\approx O(p^2d)\) with smaller constants and less operations
  \item {\bf Sparse Easy‐Attention:}
    \(O(p\,r\,d)+O(p\,d\,k)\approx O(p\,r\,d)\).
\end{itemize}

Thus Easy‐Attention unites the adaptivity of self‐attention with the operator‐theoretic guarantees of EDMD, achieving the best finite‐data, nonlinear, and irregular‐sampling approximation of the Koopman operator. 

\medskip

\paragraph{Error Tradeoff.}
Let:
\[
E_{\rm SA}(p)\;=\;\frac{\|{\bf W_QW_K^T}\|_2}{\sqrt{k}}\;2\kappa^2\Bigl(\sqrt{\tfrac{d+\ln(2/\delta)}{p}}
                                   +\tfrac{d+\ln(2/\delta)}{p}\Bigr)
\]
be the operator‐norm deviation of self‐attention from its infinite‐sample limit (Theorem~\ref{thm:cov-conc}), yielding a one‐step error of order \(O(E_{\rm SA}(p))\).  Let:
\[
E_{\rm EA}(N)\;=\;O\bigl(1/N\bigr)
\]
be the sampling variance of the least‐squares estimator \(\alpha^*\).  Equating \(E_{\rm SA}(p)\approx E_{\rm EA}(N)\) shows a crossover at
\[
N\;\approx\;\frac{2\kappa^2\|{\bf W_QW_K^T}\|_2}{\sqrt{k}}\;\sqrt{p},
\]
so that for \(N\gg\sqrt{p}\), Easy‐Attention’s \(O(1/N)\) decay dominates, while for \(N\ll\sqrt{p}\), self‐attention’s \(O(1/\sqrt{p})\) floor may be smaller.

\medskip

\noindent\textbf{Why Chaotic Forecasting Differs from Language Modeling.}
These two regimes explain why Easy‐Attention excels in chaotic dynamical systems.  Chaotic systems are \emph{stationary} and \emph{ergodic}: long trajectories sample the same invariant attractor, so a single static operator \(\alpha^*\) applies universally.  As \(N\gg\sqrt{p}\), the Law of Large Numbers drives the finite‐data error \(O(1/N)\) well below the covariance floor \(O(1/\sqrt{p})\).

In contrast, natural language is inherently \emph{non‐stationary} and context‐dependent.  Each new sentence or document has distinct topical and stylistic distributions, effectively giving \(N\approx1\) per “dataset.”  The \(O(1/N)\) advantage vanishes, and only per‐sample, input‐dependent self‐attention can adaptively reweigh token affinities to track shifting semantics. To conclude, the adaptive nature of the self-attention could be coupled with the easy-attention when studying more complex systems where multiple attractors co-exist. Given so, an individual easy-attention module can learn each attractor independently for a later correlation through the self-attention.

\section{Future Work and Potential Directions}

While Easy‐Attention already demonstrates clear advantages in finite‐data, nonlinear, and irregular‐sampling regimes, several promising avenues remain for further investigation:

\begin{itemize}
    \item \textbf{Parameter and Memory Efficiency.}  
  Explore structured or low‐rank factorizations of \(\alpha\) (e.g.\ sparse, block‐diagonal, Toeplitz) to further reduce storage and computation, especially for very long histories; this echoes recent structured modeling work in Etometry~\cite{AghabalyanGR23}.

  \item \textbf{Regularization and Constraints.}  
    Investigate convex penalties (e.g.\ $\ell_1$ sparsity, spectral radius bounds, nonnegativity) on \(\alpha\) to enforce stability, interpretability, or desired dynamical properties.

  \item \textbf{Spectral Analysis and Stability.}  
    Analyze the eigenvalues and pseudospectra of the learned \(\alpha\) to characterize system stability, timescales, and identify dominant coherent structures, akin to Koopman spectral theory.

  \item \textbf{Online and Adaptive Learning.}  
    Extend Easy‐Attention to streaming data by developing incremental or adaptive solvers for \(\alpha\), enabling real‐time model updating under nonstationary dynamics.

  \item \textbf{Hybrid Physics–ML Models.}  
    Combine Easy‐Attention with classical reduced‐order or physics‐based operators (e.g.\ DMD, Galerkin models, Kalman filters) in a \emph{hybrid} framework, leveraging known governing equations and learned corrections.

  \item \textbf{Uncertainty Quantification.}  
    Place Bayesian or Gaussian‐process priors directly on \(\alpha\) to quantify predictive uncertainty and guide experimental design in sparse‐data regimes.

  \item \textbf{Transfer and Multi‐Task Learning.}  
    Study how a learned \(\alpha\) for one chaotic system can be fine‐tuned or transferred to related systems, reducing data requirements for new applications~\cite{eiximeno2024deeplearningbasedclosuresalgebraicsurrogate}.

  \item \textbf{Multi‐Scale and Multi‐Modal Embeddings.}  
    Incorporate hierarchical or multi‐resolution embeddings (e.g.\ wavelets, graph‐based delays) into \(\bf W_V\) to capture interactions across scales or modalities (e.g.\ coupling of fluid and structural dynamics).

  \item \textbf{Interpretability and Visualization.}  
    Develop tools to visualize the learned \(\alpha\) as a transition graph or directed network, providing physical insight into dominant pathways of information flow.

  \item \textbf{Integration with Control and Reinforcement Learning.}  
    Take advantage of the Easy‐Attention in model‐based control or policy learning, where \(\alpha\) can serve as a predictive surrogate in optimal control or reinforcement‐learning loops.
\end{itemize}

These directions promise to deepen the theoretical foundations of Easy‐Attention, broaden its applicability to new domains, and enhance its integration within scientific and engineering workflows.

	

\subsubsection*{Computation}
In the present study, all the model training and testing are implemented on a single central-processing unit (CPU) from a machine with the following characteristics: AMD Ryzen 9 7950X with 16 cores, 32 threads, 4.5 Ghz and 62 GB RAM.
All machine-learning models and experiments were conducted using the PyTorch framework~\cite{paszke2017automatic} (version 2.0.0). PyTorch was chosen due to its flexibility, ease of use, and extensive support for deep learning research. The custom models and neural-network architectures used in this work were implemented using the modular design from PyTorch, allowing seamless composition of layers and custom components. All the data and codes will be made available open access upon publication of the manuscript in the following repository: \url{https://github.com/KTH-FlowAI}

\section*{Acknowledgments}
R.V. acknowledges financial support from ERC grant no.‘2021-CoG-101043998, DEEPCONTROL’ and the EU Doctoral Network MODELAIR. Part of the ML-model testing was carried out using computational resources provided by the National Academic Infrastructure for Supercomputing in Sweden (NAISS). Jasmin Lim and Karthik Duraisamy are supported by APRA-E under the project \textit{SAFARI: Secure Automation for Advanced Reactor Innovation} at the University of Michigan.
Roger Arnau was supported by a contract of PAID-01-21 from the Universitat Polit\`ecnica de Val\`encia.
{The authors would like to thank to Prof. Igor Mezi\'{c} for helpful feedback on the manuscript Francisco-Javier Granados-Ortiz for fruitful discussions.
.} This article is based on previous versions found in Refs.~\cite{sanchisagudo2023easy,vinuesa2023easy}.

\section*{References}
\bibliography{ref}

\section*{Supplementary Material}
\noindent In this Supplementary Material we expand the description and validation of the architectures discussed in the main article, facilitating their interpretation through additional figures and results. \\


\section*{Sinusoidal-wave reconstruction with attention mechanisms}
\label{sec:sin}
Firstly, Fig.~\ref{fig:Sin} depicts the reconstruction capability for both attention mechanisms when inspecting a periodic sinusoidal wave, in red and blue, the easy and self attention respectively.
\begin{figure}[H]
    \centering
    \includegraphics[width=0.6\textwidth]{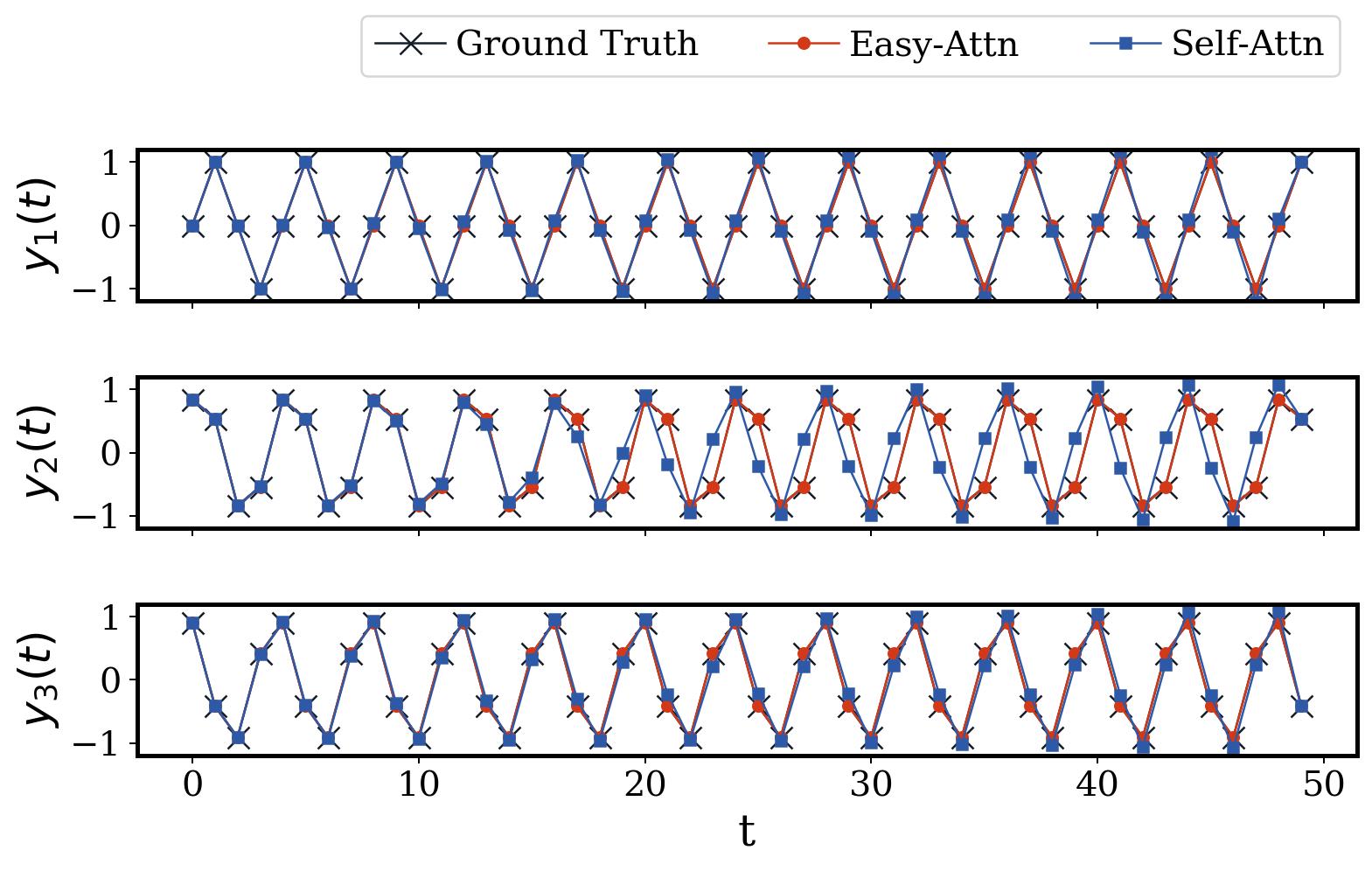}
    \caption{Reconstruction results of sine functions with phase shifts. Here, we compare the ground truth with reconstructions through the self- and easy-attention methods.}
    \label{fig:Sin}
\end{figure}

\noindent Table~\ref{tab:sine_wave_results} summarizes the obtained $l_2$-norm error, as well as the computation time for training ($t_c$) and number of floating-point operations ($N_f$) of the employed easy-attention and self-attention modules. Note that $t_c$ is calculated using the resources described in the Computation section from the main article, and that we compute $N_f$ for one forward propagation using a batch size of 1. 

\begin{table}[h!]
    \begin{tabular}{c|cccc}
    \hline
                  & Parameters & $\varepsilon$ (\%) & $t_c$ (s) & $N_f$ \\ \hline
    {Self-Attn} & 36         & 10\%               & 26.88     & 81 \\ \hline
    {Easy-Attn} & 18         & \textcolor{red}{0.0018\%}           & 19.20     & 45 \\ \hline
    \end{tabular}
    \caption{Summary of a single self-attention and an easy-attention module used for sinusoidal-wave reconstruction. In red we show the best reconstruction.}
    \label{tab:sine_wave_results}
\end{table}

\section*{Chaotic Van der Pol oscillator with discrete Fourier transform}
\label{sec:vdp_fft_rec}

In this section, motivated by the promising reconstructing capability of the easy-attention module, the objective is to analyze the specific  mechanism for both attention models searching for explainability between input, output and learning parameters.
Given the results discussed above, it can be observed that the easy attention exhibits promising performance on sinusoidal-wave reconstruction, which implies that the easy-attention method has the potential to take advantage of periodicity.
This unique potential motivates us to study a more complex system through discrete Fourier transform, which produces the sinusoidal wave by implementing the Euler formula. To this end, we propose a method called multi-easy attention, which employs multiple easy-attention modules to learn the frequencies of the dominant $K$ amplitudes which are used to reconstruct the trajectory.  In the present study, we determine the value of $K$ by requiring that the reconstructed trajectory must exhibit a reconstruction accuracy $E_{\rm rec}$ of 90\%. We define the reconstruction accuracy as: 
\begin{equation}
    E_{\rm rec} = \left |1 - \frac{|| S - \widehat{S} ||_2}{||S||_2} \right| \times 100\%,
\end{equation}
\noindent where $S$ and $\widehat{S}$ denote original trajectory and the one reconstructed from the truncated representation, respectively.

\noindent Fig.~\ref{fig:Multi} illustrates the implementation of the multi-easy-attention method combined with the discrete Fourier transform. The trajectory is transformed into the frequency domain via the discrete Fourier transform, and we sort out the dominant $K$ amplitudes from all $M$ amplitudes. Subsequently, we employ $K$ attention modules to identify and reconstruct the frequencies in the form of the Euler formula. We utilize reconstructed frequencies from modules and their corresponding amplitudes to reconstruct the trajectory. Note that for each frequency we employ a single attention module with one head, {and the the input size ($d_{\rm in}$) is equal to the smallest integer of the real part of its period ($P$), such that $d_{\rm in} \geq P$. In this way, we are again adopting the time delay for the prediction of frequency for each amplitude.} If the period is infinity, we set $d_{\rm in} = 2$. We employ SGD with learning rate of $1\times 10^{-3}$ as the optimizer and MSE as the loss function. Each module is trained for 1,000 epochs with a batch size of 8. To investigate the difference between easy attention and self attention in terms of performance, we substitute the easy-attention module by self attention to build a multi-self attention method for reconstruction.
\begin{figure}[ht]
    \centering
    \includegraphics[width=0.6\textwidth]{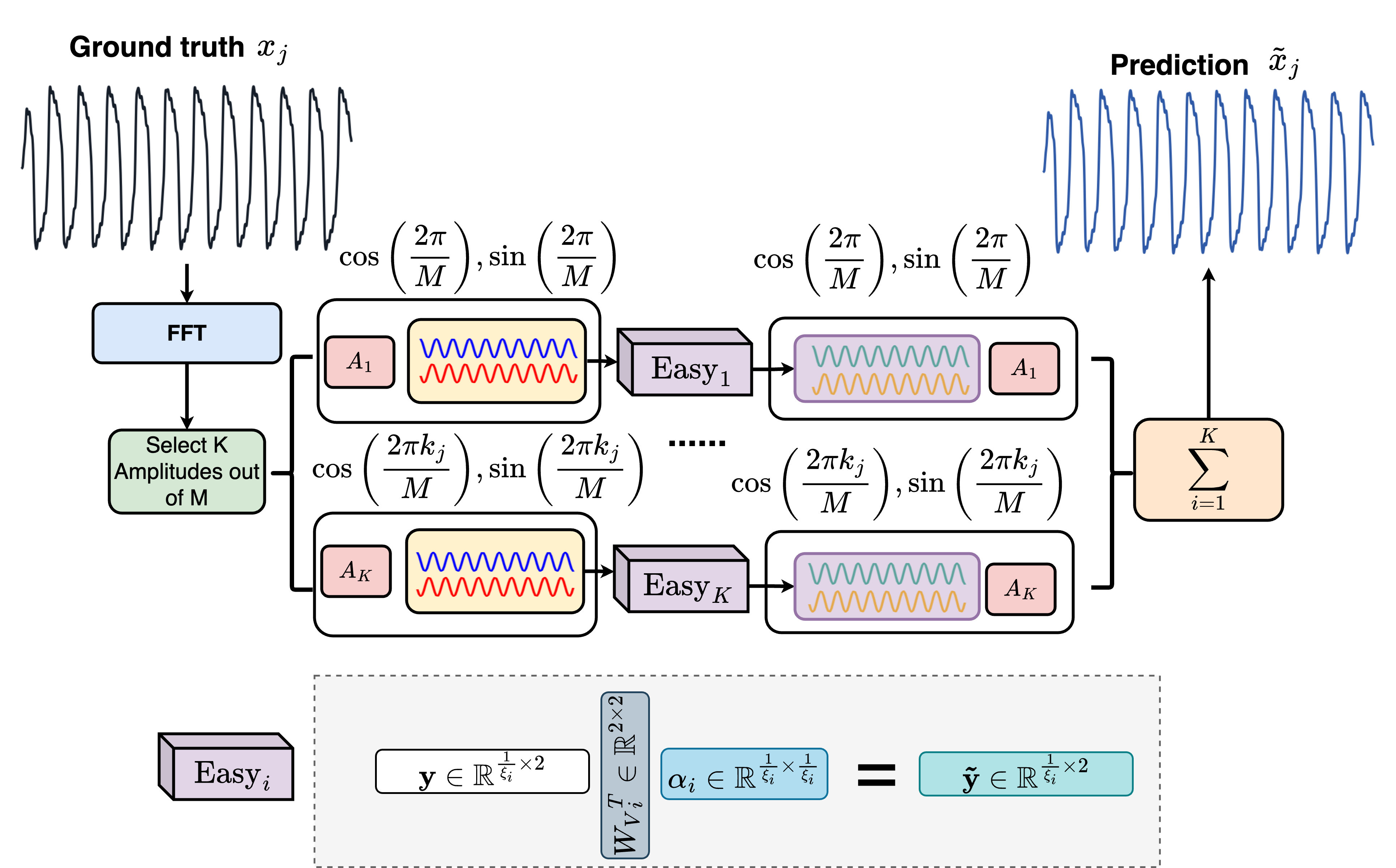}
    \caption{Schematic illustration of the multi-easy-attention method. We show the reconstruction of sinusoidal wave decomposed from discrete Fourier transform using the $K$ dominant amplitudes.}    \label{fig:Multi}
\end{figure}

\noindent In the present study, we apply our method to study the van der Pol chaotic system with a source term~\cite{Marios}:
$$\frac{{\rm d}x}{{\rm d}t} = y, \quad \frac{{\rm d}y}{{\rm d}t} = -x + a(1-x^2)y + b\cos(\omega t),$$
\noindent where $a$, $b$ and $\omega$ are the parameters of the system. We investigate three types of solutions, namely periodic $(a=5,b=40,\omega=7)$, quasi-periodic $(a=5,b=15,\omega=7)$ and chaotic $(a=5,b=3,\omega=1.788)$, as discussed in  Ref.~\cite{Marios}. Note that in the present study, we only investigate the temporal evolution of $x$ in the van der Pol system. The objective of this study is to further inspect the capability of the each attention mechanism to capture and learn different frequencies.

 
\noindent Table~\ref{tab:Fourier} lists the results of trajectory reconstruction for multi-easy-attention and multi-self-attention, which are denoted as $\rm easy$ and $\rm self$, respectively. We compute the $l_2$-norm error as in Eq.(\ref{eq:l2-error}) for each frequency and then average over the retained $K$ frequencies. The results demonstrate that easy attention significantly outperforms self attention, as it is not only able to reconstruct all frequencies with a higher accuracy but also identifies differences in phase between the waves based on sine and cosine.

\begin{table}[h!]
    \centering
    \begin{tabular}{c|ccc}
              \hline
                            \textbf{Case}       & \textit{K} & $\overline{{\varepsilon}}_{\rm easy}$ (\%) & $\overline{{\varepsilon}}_{\rm self}$ (\%)\\ \hline
                            Periodic       & 100  & \textcolor{red}{0.149} &   92.517       \\ \hline
                            Quasi-periodic & 60   & \textcolor{red}{0.119} &   114.79            \\ \hline 
                            Chaotic        & 130  & \textcolor{red}{0.141} &   110.60 \\ \hline  
        
    \end{tabular}
    \caption{Summary of $l_2$-norm error of prediction obtained by self attention and easy attention on the sinusoidal function of first $K$ amplitudes, where the best performance is colored in red. Note that $\overline{\varepsilon}$ indicates the average error calculated on each amplitude individually and that the errors are computed based on the Fourier-truncated signal.}
    \label{tab:Fourier}
\end{table}

\noindent Furthermore, Table  \ref{tab:Fourier} demonstrates the capability of the multi-easy-attention to recognise and reconstruct dynamics. In Fig.~\ref{fig:fft_reconstructions} it can be observed that the easy attention replicates the dynamics of the original signal for all the studied sets of parameters, not only in the short term but as long as the period remains constant. Next, we will assess the performance of the exact dynamic-mode decomposition (DMD)~\cite{tu_dynamic} when predicting this dynamical system

\begin{figure}[h!]
    \centering
    \includegraphics[width=0.6\textwidth]{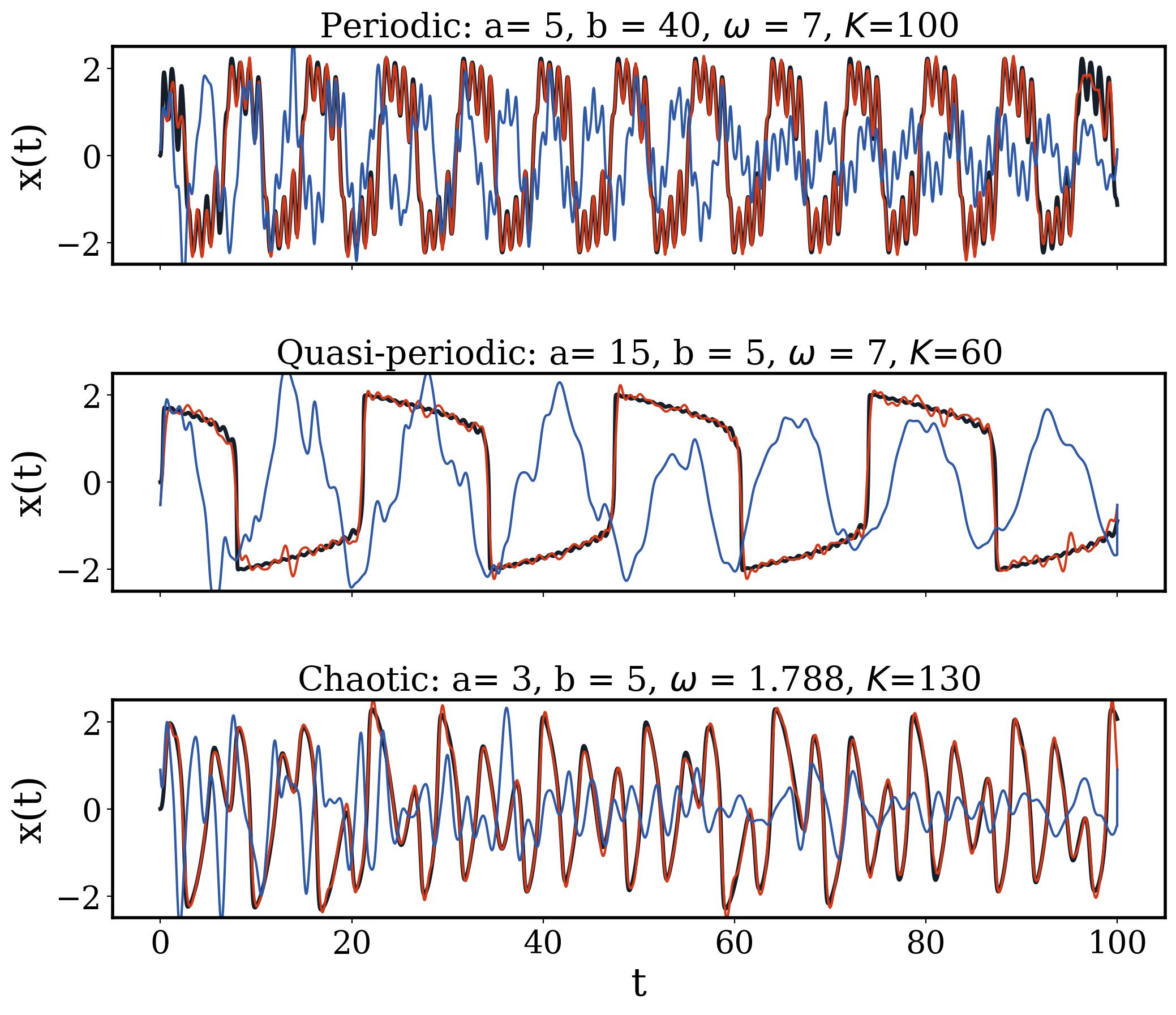}
    \caption{Reconstruction of temporal evolution of variable $x$ via sinusoidal wave decomposed from discrete Fourier transform using dominant $K$ amplitudes using varied models for three cases. In each subplot, the black line denotes the reference trajectory, whereas the red line and blue line denote the reconstruction of multi-easy-attention and multi-self-attention, respectively.}
    \label{fig:fft_reconstructions}
\end{figure}

\subsubsection{Exact dynamic-mode decomposition (DMD) for reconstruction of van der Pol system}
We employ the exact-DMD algorithm~\cite{tu_dynamic} for reconstructing the variable $x$ in the van der Pol system by implementing the discrete Fourier transform on the trajectory, and subsequently apply the DMD to reconstruct the frequencies in Euler form. Note that we use the same amount of leading DMD modes as the number of dominant amplitudes used in the discrete Fourier transform for the periodic, quasi-periodic and chaotic cases, respectively. Fig.~\ref{fig:plain_dmd} depicts the results reconstructed by DMD, where it can be observed that the DMD reconstructions meet in very good agreement with the reference trajectories for all three cases. We compute the $l_2$-norm error as in (\ref{eq:l2-error}) for each frequency and the average over the retained $K$ frequencies. The DMD led to ${\varepsilon} = 9.3 \times 10^{-10} \%$, $1.6 \%$ and $1.25 \times 10^{-9}\%$ for the periodic, quasi-periodic and chaotic cases, respectively ({\it i.e.} 100, 60 and 130).
\begin{figure}[h!]
    \centering
    \includegraphics[width=0.6\textwidth]{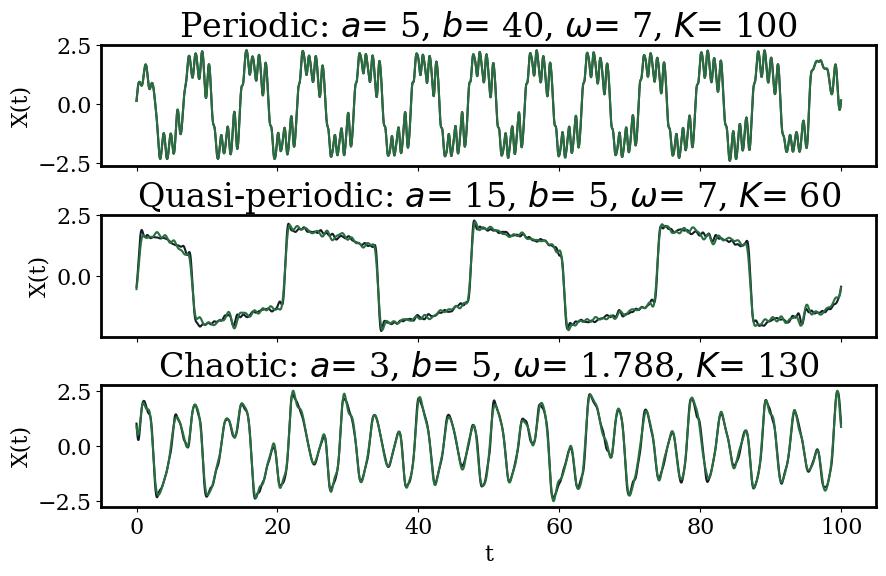}
    \caption{Reconstruction of temporal evolution of variable $x$ via exact-DMD algorithm using the $K$ leading DMD modes the three cases under study. In each subplot, the black line denotes the reference trajectory, whereas the green line denotes the reconstruction of the exact DMD.}
    \label{fig:plain_dmd}
\end{figure}

\begin{equation}
    \varepsilon = \frac{|| S - \Tilde{S} ||_2}{||S||_2}, 
    \label{eq:l2-error}
\end{equation}

\noindent  Note that in equation~(\ref{eq:l2-error}) $S$ is the {ground-truth} sequence and $\Tilde{S}$ is the prediction obtained {by the} attention module. To conclude this comparison between DMD and the multi easy-attention method, it is important to acknowledge the similarities between both models. The easy-attention idea is based on finding the hidden optimal spatio-temporal basis of the system, similarly to DMD. Both approaches follow an optimization procedure where all information needs to be presented. The globality of the system is revealed to the transformer through the training period, so the exposure to the invariancies of the system along this stage allows for the ability to predict or reconstruct invariant sets. On the other hand, the optimization procedure of the operator matrix achieved through DMD also requires also global information for reconstructing accurately. Our model is capable of predicting up to the Lyapunov time and after this point it is also capable of reconstructing accurate dynamics by reproducing the desired attractor or probability density function, while the DMD lacks this ability, as shown in Fig.~\ref{fig:hdmd_attractor}.

\section*{Chaos analysis}
\label{sec:chaos}
In this section, we will introduce and develop technical aspects regarding the architectures for all study cases.

\begin{table}[ht]
    \centering
    \begin{tabular}{c|c|c|c}
        \hline
        \textbf{Cases} & \textbf{Lorenz system}  & \textbf{Turbulent flow} & \textbf{Nuclear reactor}\\ \hline
        \rule{0pt}{2.5ex} No.Train Samples         & $1\times10^{6}$           & $4\times10^{7}$ & 89,038 \\
        \rule{0pt}{2.5ex} No.Test Samples         & $1\times10^{4}$           & $2\times10^{6}$  & 43,845 \\
        \rule{0pt}{2.5ex} $\Delta t$         & 0.01 s        & 0.01 s          & 5 s     \\
        \rule{0pt}{2.5ex} No.Variables       &  3         & 9               & 13\\
        \hline
    \end{tabular}
    \caption{Summary of all training data used by LSTM, self-attention and easy-attention transformer networks.}
    \label{tab:data_describe}
\end{table}

\noindent As can be observed in Table~\ref{tab:data_describe}, we studied the easy-attention mechanism in three different scenarios, from more canonical mathematical problems to more challenging and applied scenarios. Note that in the table we document the size of the dataset, the time steps and the number of variables in each of the cases. To further examine the capabilities of the easy-attention method, we apply it on a transformer encoder for temporal-dynamics prediction. We adopt the idea of time delay~\cite{Karthik} for the task, which uses data sampled from the previous $p$ time steps to predict the next time one.
A schematic illustration of the proposed transformer architecture is presented in Fig.~\ref{fig:arch_transformer}. It comprises an embedding layer, a transformer encoder block and two decoding layers for time delay. Note that the multi-head attention in the encoder block is either easy attention or self attention in the present study, see Methods section in the main article for more information regarding the attention mechanisms.
Instead of the positional encoding used in the original transformer, we adopt the sine time2vec embedding~\cite{kazemi2019time2vec} to enhance identification of the temporal information present in the input sequence.
Observe that, since the attention weights $\alpha_{i,j}$ of easy attention only depend on the position and not on the input, an encoding that adds information about the relative position of the input is not necessary.
The rectified linear unit (ReLU) is employed as activation function for the feed-forward neural network. For the output encoding, a one-dimensional convolution neural network (Conv-1D) and a multilayer perceptron (MLP) are employed to project the information as the prediction of the next time step. The difference between the dense and sparse easy attention is that, for each head $i$, the dense easy attention learns all elements $\boldsymbol{\alpha}^i \in \mathbb{R}^{p \times p}$ while in the sparse one we assume that the attention is a tridiagonal matrix.

\subsection{Lorenz system architecture and results}
In the following section we introduce the architecture employed in Lorenz system section in the main article. The employed transformer architectures are summarized in Table~\ref{tab:attn_arch_lorentz}. Note that we apply dense and sparse multi-head easy attention as the easy-attention respectively. 
\begin{figure}[H]
    \centering
    \includegraphics[angle=270,width=0.55\textwidth]{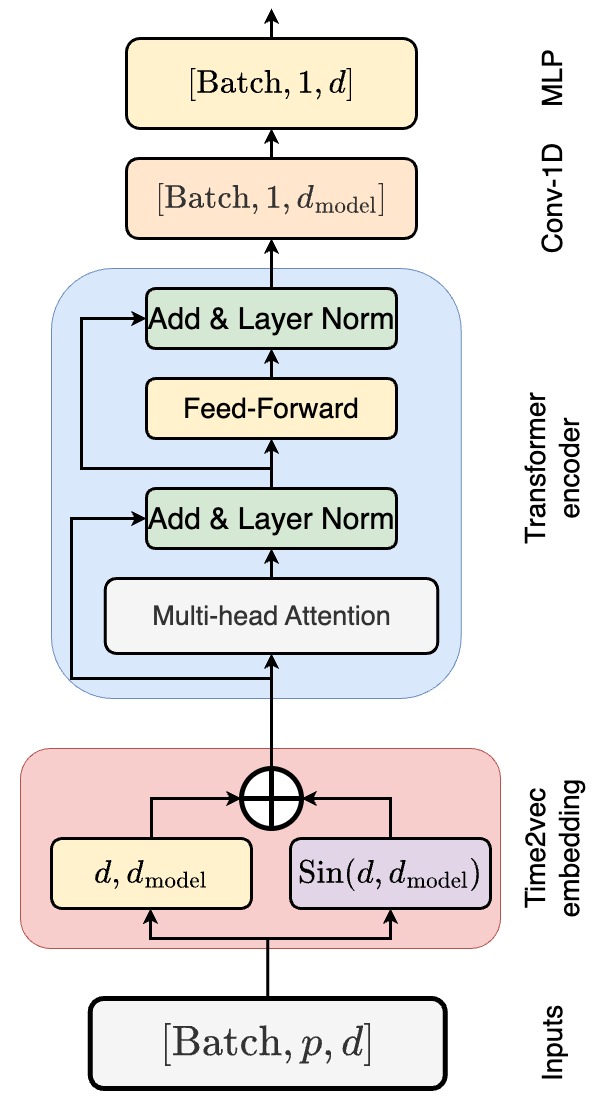}
    \caption{Schematic view of the transformer architecture for temporal-dynamics prediction. The dimension of the output for each layer has been indicated in each block, {where $p$ denotes} the size of time delay whereas $d$ and $d_{\rm model}$ denote the number of features and the embedding size, respectively.}
    \label{fig:arch_transformer}
\end{figure}

\begin{table}[h!]
    \centering
        \begin{tabular}{c|cccccc}
        \hline
        \textbf{Name}        & \textbf{$p$} & \textbf{$d_{\rm model}$} & {No.head} & \textbf{$d_V$} & {Feed-forward} & {No.Block} \\ \hline
        Easy-Attn   & 64          & 64                       & 4 & 16            & 64          & 1                 \\
        Sparse-Easy & 64           & 64                       & 4 & 16            & 64           & 1                 \\
        Self-Attn   & 64           & 64                      & 4 & 16            & 64          & 1                 \\ \hline
        \end{tabular}
    \caption{Summary of transformer architecture employed in the Lorenz-system prediction. {We denote the size of time delay as $p$}, the embedding size as $d_{\rm model}$ and the values projected dimension as \textbf{$d_V$}, respectively. Note that the dense easy attention is denoted as Easy-Attn while the sparse easy attention is denoted as Sparse-Easy.}
    \label{tab:attn_arch_lorentz}
\end{table}
\noindent In addition, to investigate the difference in performance between easy-attention-based transformer and recurrent neural networks (RNNs), we also use the one-layer long short-term memory (LSTM) architecture. The LSTM model uses the same time delay $p$ as the transformer architectures with the number of units in the hidden layer being 128. We adopt Adam as the optimiser with a learning rate of $1 \times 10^{-3}$, and the MSE as loss function. Finally, the models are trained for 100 epochs with batch size of 32. Note that we concatenate the predicted variables into the inputs for predicting the variables at the next time step, such that the transformer's predictions are fed back as inputs on the time delay. In Table~\ref{tab:metrics_lorentz} we show the specifications of each model after the training stage. The most relevant results are mentioned in the main article. 

\begin{table}[ht]
    \centering
    \resizebox{0.5\textwidth}{!}{%
        \begin{tabular}{c|cccc}
        \hline \\[-1em]
        \textbf{Name}    &  Parameters           & \textbf{$\varepsilon$} (\%) & $t_c$ (s) & $N_f$\\[0.2em] \hline \\[-1em]
        {Easy-attention}& 29.75 $\times$ $10^3$ & \textcolor{red}{1.99}  & 9.16 $\times$ $10^3$ &     49.22 $\times$ $10^3$           \\
        {Sparse-Easy} & 13.44 $\times$ $10^3$  & {2.79}                  & 9.10 $\times$ $10^3$ & 49.22 $\times$ $10^3$  \\
        {Self-Attn}&  25.48 $\times$ $10^3$  & 7.36 & 11.07 $\times$ $10^3$ &   65.61 $\times$ $10^3$        \\
        {LSTM}   &   37.51 $\times$ $10^3$  & {37.68} &  4.85 $\times$ $10^3$ &   41.00 $\times$ $10^3$       \\ \hline
        \end{tabular}%
        }
    
    \caption{Summary of results on the Lorenz system obtained from different models. The error is computed for the time series until step 512. In red we show the lowest prediction error.}
    \label{tab:metrics_lorentz}
\end{table}

\noindent Long-term predictions of the variables over 10,000 time steps are shown in Fig.~\ref{fig:attractor_Visualisation} for various models. The results clearly show that the LSTM and self-attention-based transformer model are not able to reproduce the correct dynamical behavior over long periods of time, while the transformer models using easy attention are able to achieve better predictions, with an error of $\epsilon = 1.99\%$; this can be clearly observed in the temporal evolution of variable $x$ from the Lorenz system, shown in Fig.~\ref{fig:Lorentz_temporal_evo} and Tab.~\ref{tab:metrics_lorentz}.

\begin{figure}[ht]
    \centering
    \begin{subfigure}{0.2\linewidth}
        \centering
        \caption{{Easy attention}}
        \includegraphics[width=\textwidth]{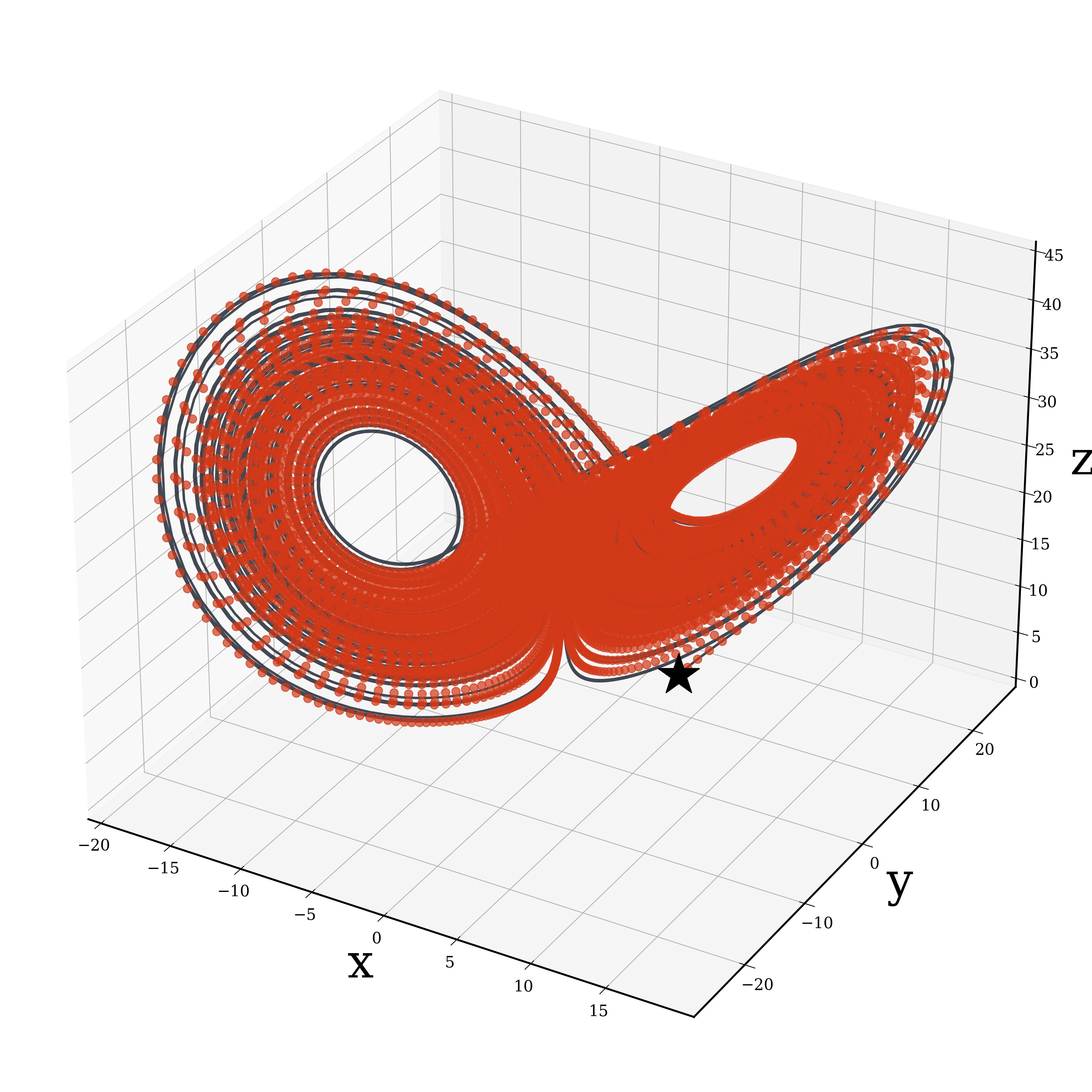}
    \end{subfigure}
    \quad
    \begin{subfigure}{0.2\linewidth}
        \centering
        \caption{{Sparse easy attention}}
        \includegraphics[width=\textwidth]{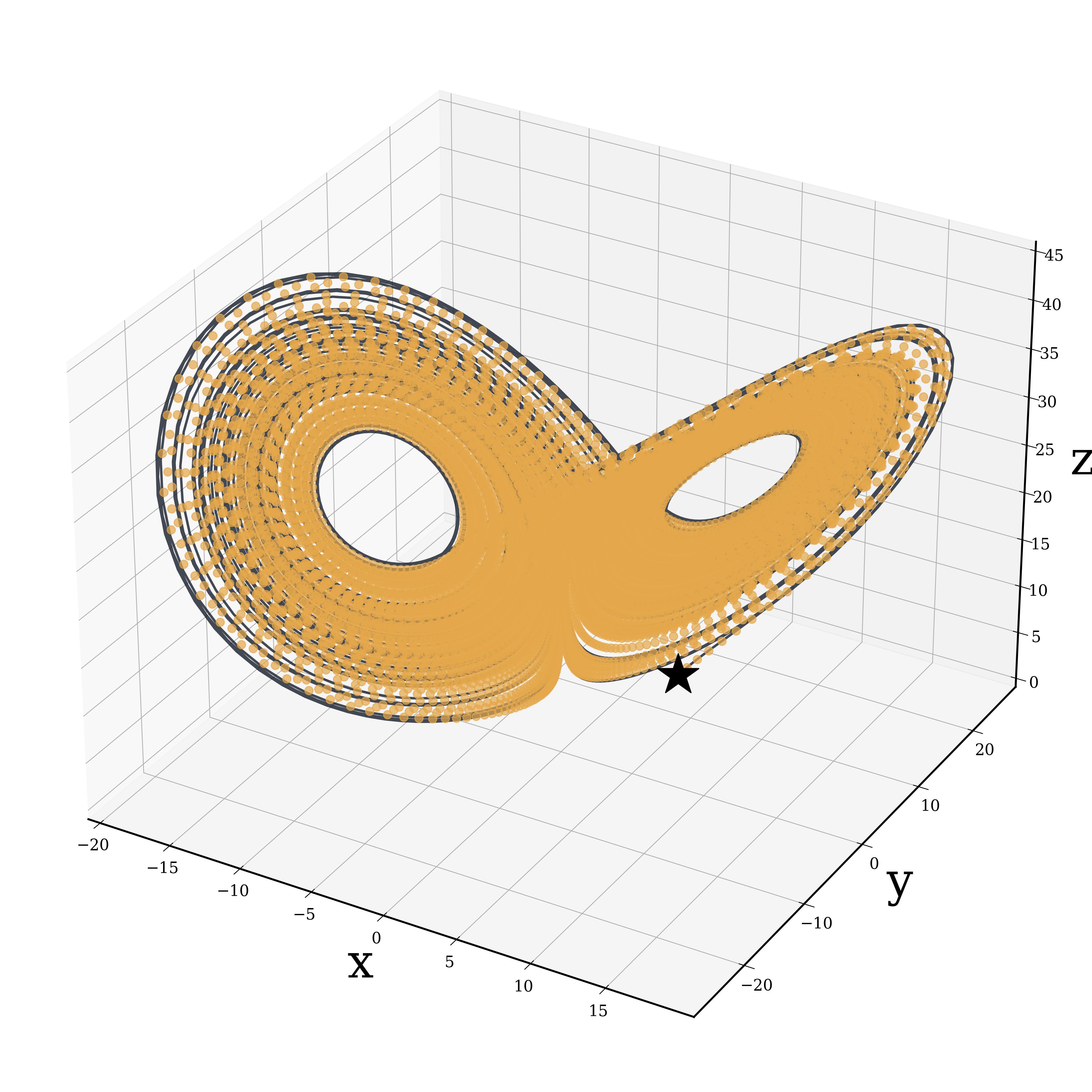}
    \end{subfigure}
    \quad
    \begin{subfigure}{0.2\linewidth}
        \centering
        \caption{{Self attention}}
        \includegraphics[width=\textwidth]{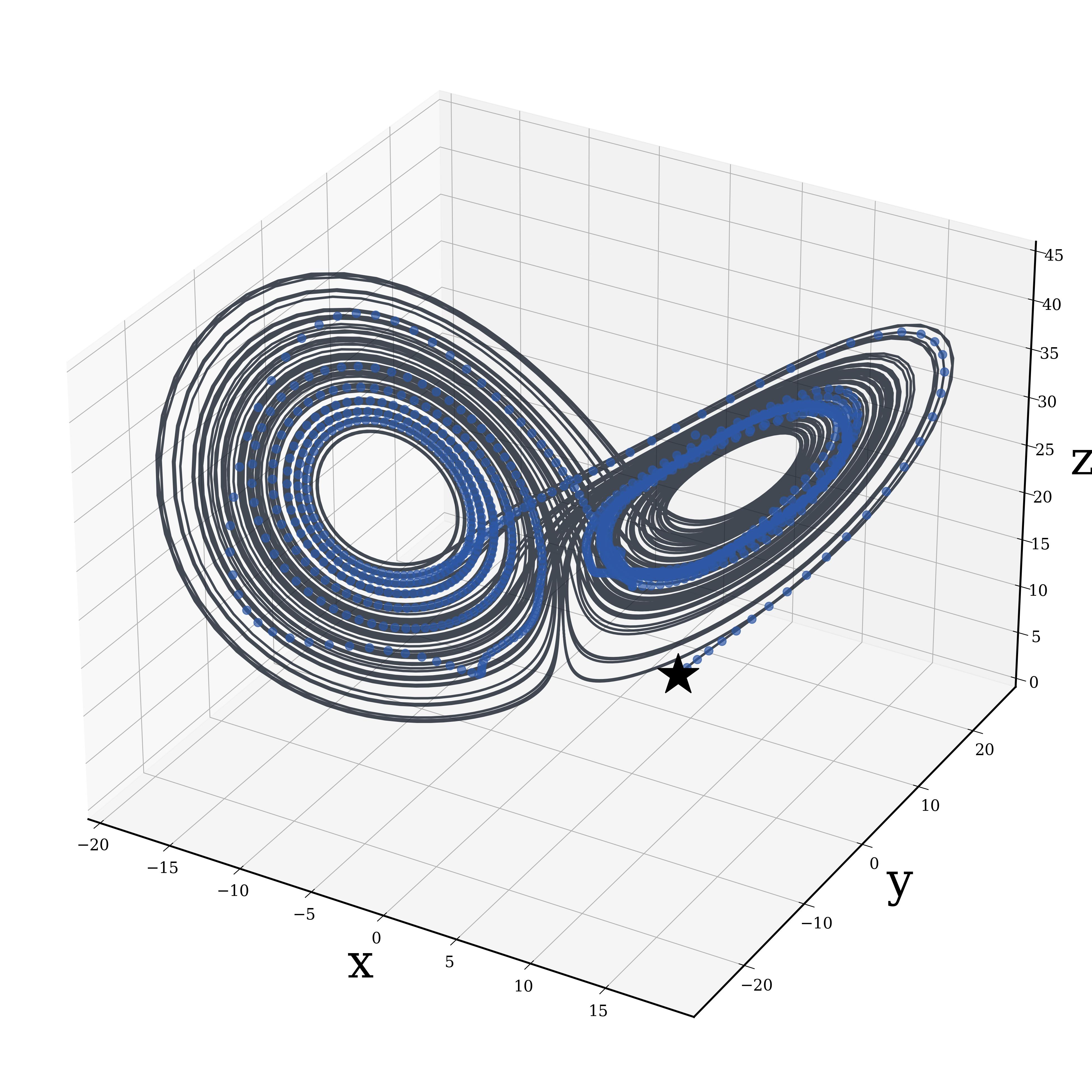}
    \end{subfigure}
    \quad
    \begin{subfigure}{0.2\linewidth}
        \centering
        \caption{{LSTM}}
        \includegraphics[width=\textwidth]{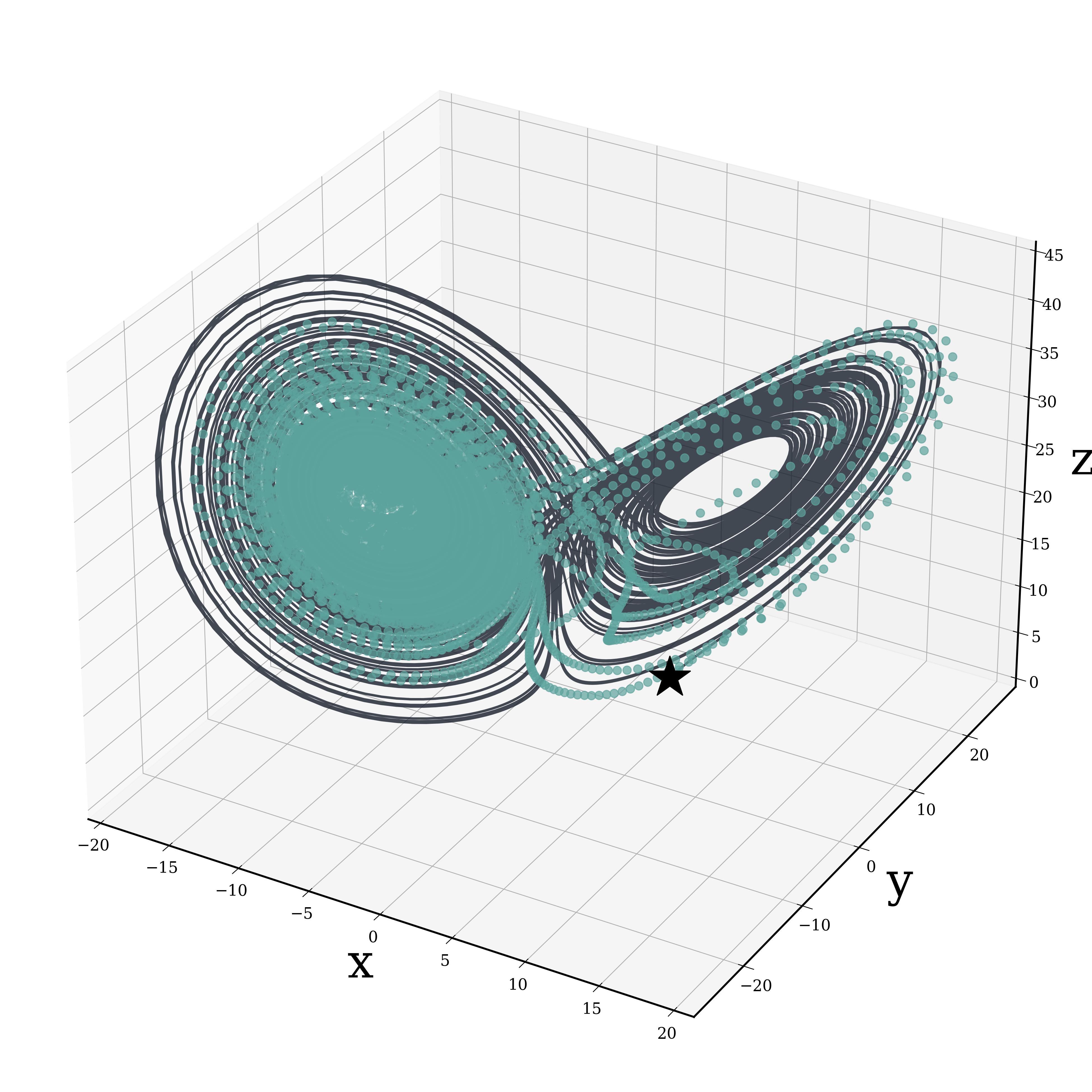}
    \end{subfigure}

    \caption{\textbf{Visualization of Lorenz system predicted by the different models.}
    The test data is indicated in black as reference and the location of initial condition ($x_0$, $y_0$, $z_0$) is marked as a black star.}
    \label{fig:attractor_Visualisation}
\end{figure}

\begin{figure}[ht]
    \centering
    \begin{subfigure}{0.48\linewidth}
        \centering
        \caption{{Easy attention}}
        \includegraphics[width=\textwidth]{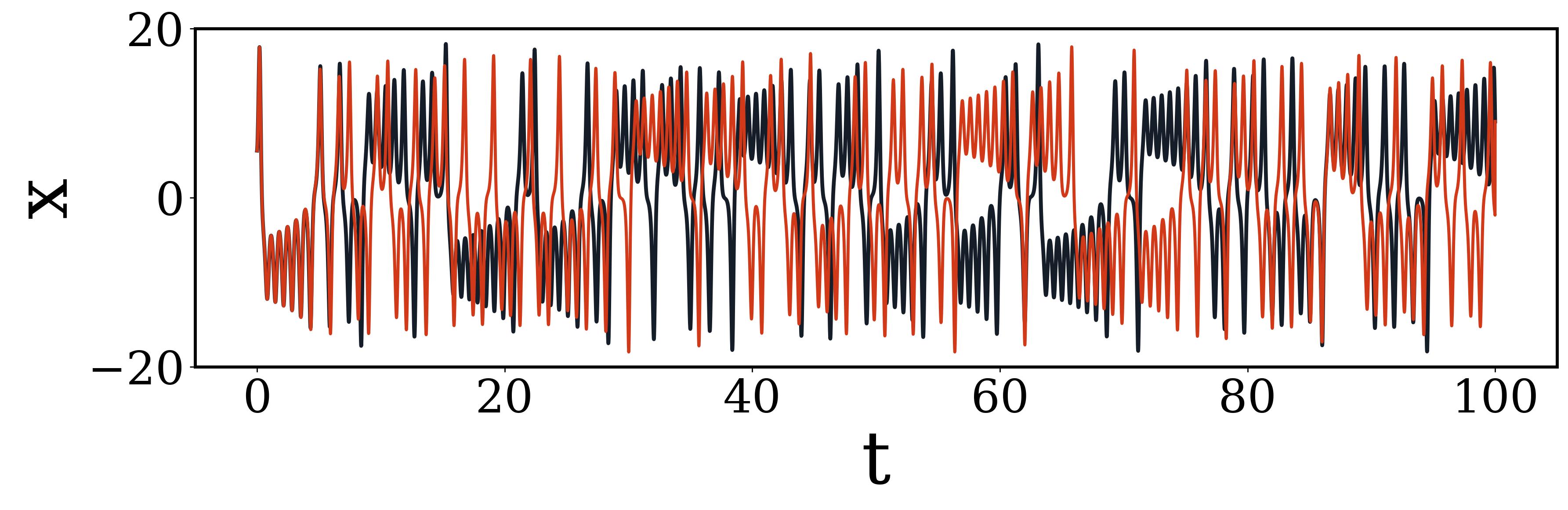}
    \end{subfigure}
    \quad
    \begin{subfigure}{0.48\linewidth}
        \centering
        \caption{{Sparse easy attention}}
        \includegraphics[width=\textwidth]{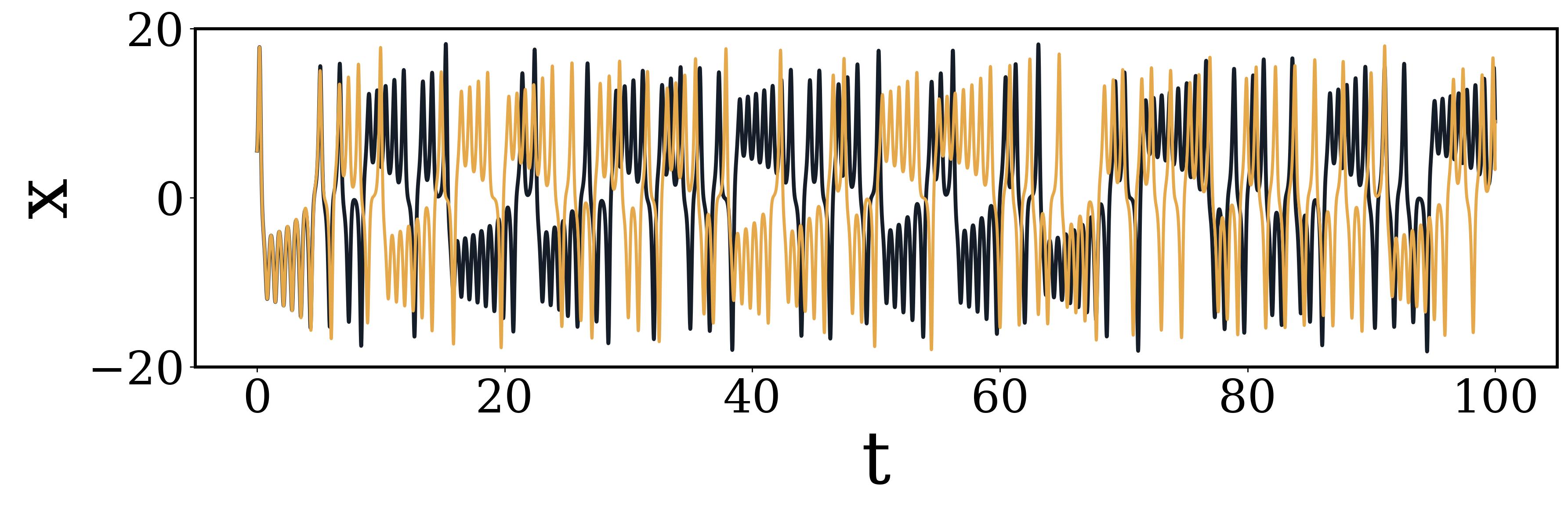}
    \end{subfigure}
    \quad
    \begin{subfigure}{0.48\linewidth}
        \centering
        \caption{{Self attention}}
        \includegraphics[width=\textwidth]{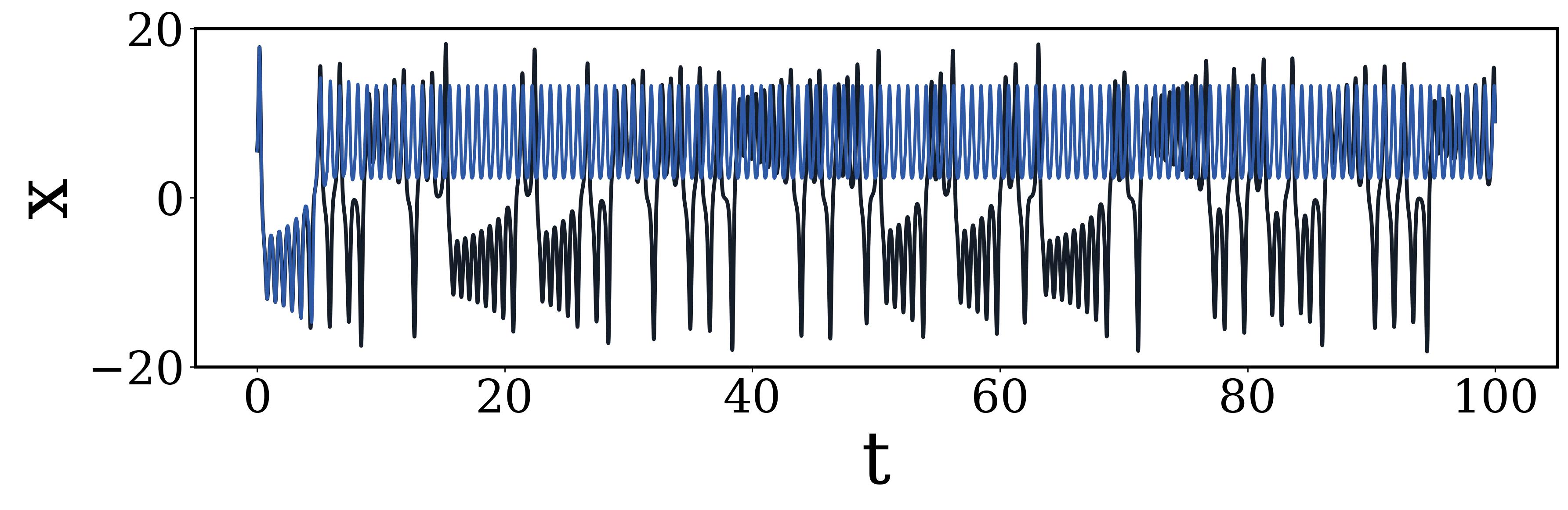}
    \end{subfigure}
    \quad
    \begin{subfigure}{0.48\linewidth}
        \centering
        \caption{{LSTM}}
        \includegraphics[width=\textwidth]{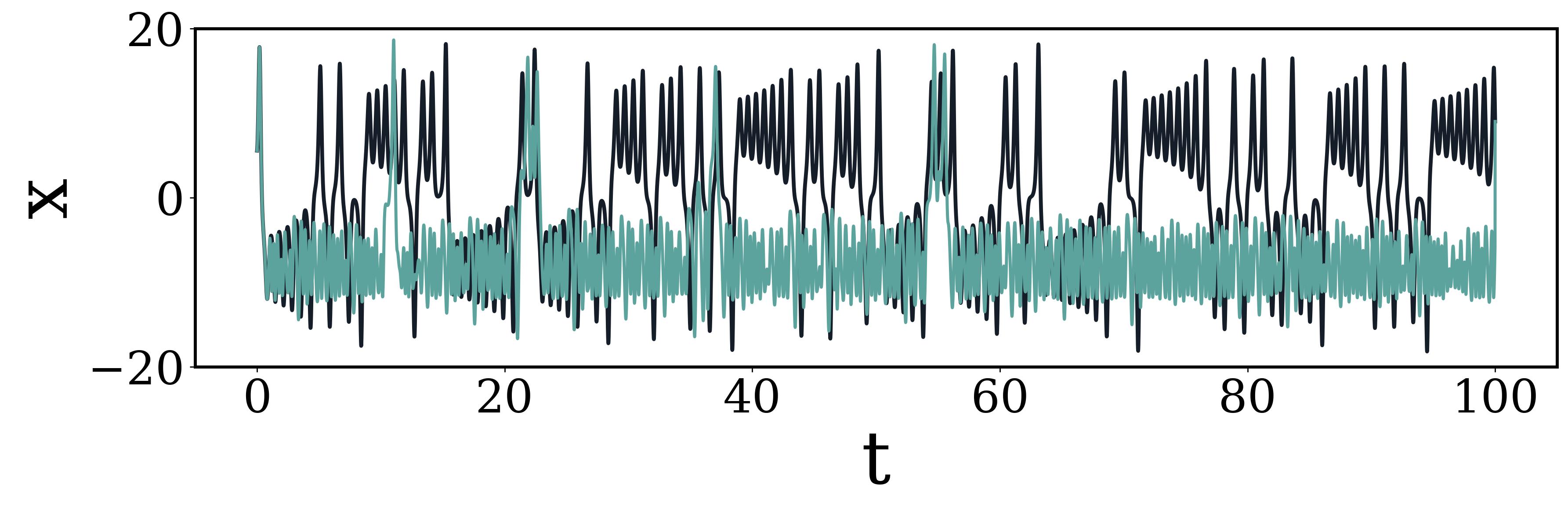}
    \end{subfigure}
    \caption{Temporal evolution of variable $x$ from the Lorenz system predicted by the different models.
    The reference data is indicated in black in all the panels.}
    \label{fig:Lorentz_temporal_evo}
\end{figure}

\subsubsection{Positionally encoded self attention}

In order to highlight that our main contribution in this work is a novel, simplified definition of the attention, we implement the self attention with positionally encoded query and key and compare it with the easy-attention. 
\noindent Attention mechanisms can be implemented in many ways, as shown in the literature in Refs.~\cite{bahdanau2016neural,luong2015effective}. Self attention, as defined in the transformer architecture, has been extensively used primarily because of its computational efficiency. In this work, we show how the use of the Q and K matrices does not provide any advantage and how the easy attention mechanism outperforms the self attention.

\noindent Our approach can be considered a special case of a conventional transformer, as we are proposing a new attention mechanism rather than a new architecture. The positional encoding is summed to the states, rather than concatenated. Thus, the proposed architecture cannot be implemented without modifying the transformer architecture in the first place. The concatenation approach would increase the number of learnable parameters and reduce the computational efficiency of the architecture. While this new implementation of the self attention should have a similar behavior than the easy-attention mechanism, there are two main differences. First, the easy attention gives a more efficient solution, since it requires fewer operations. In the multihead case, as the $W_Q$ and $\bf W_K$ are usually of size $(d_{\rm model}) \times (d_{\rm model}/n_{\rm heads})$, the resulting attention matrix should be $\alpha = encoding^T * W_Q^T * W_K * encoding$ (case 1), in contrast to our learnable $\alpha$ (case 2). Note that the difference lies in the fact that case 1 matrix is a low-rank representation of case 2, so the proposed easy attention allows more freedom on global relations. Nevertheless, we have implemented the proposed method, encoding the input of queries and keys, to validate our method and we obtain an error of 2.22\%, at t=$5.12$.

\begin{figure}[h]
    \centering
    \includegraphics[width=0.5\linewidth]{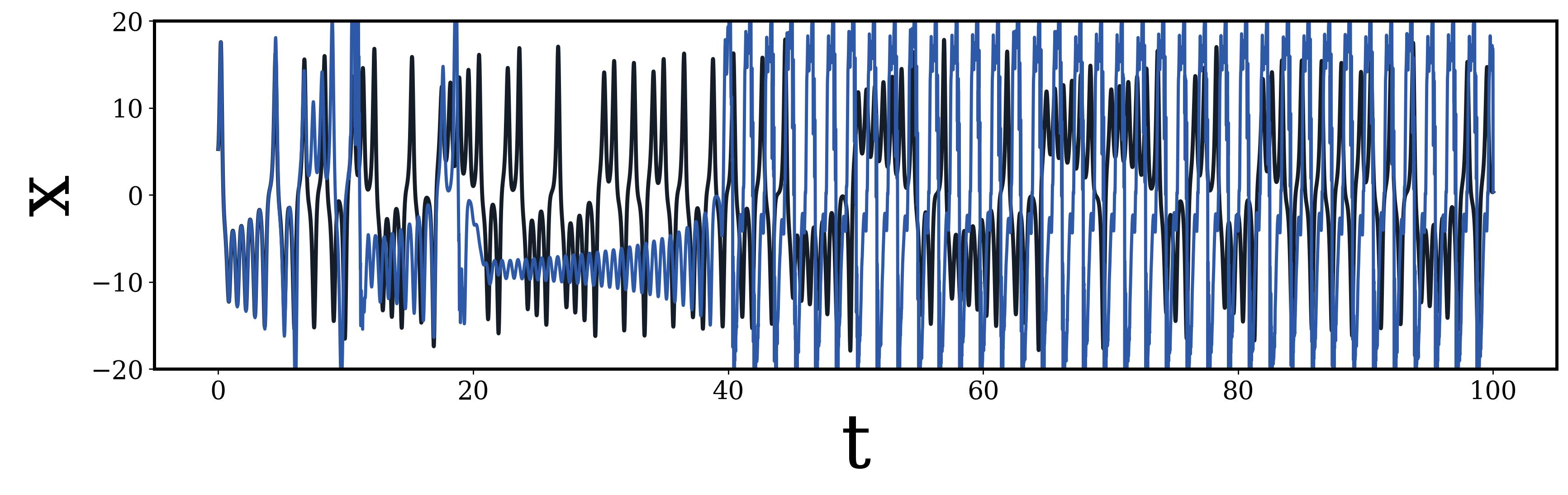} \quad
    \includegraphics[width=0.2\linewidth]{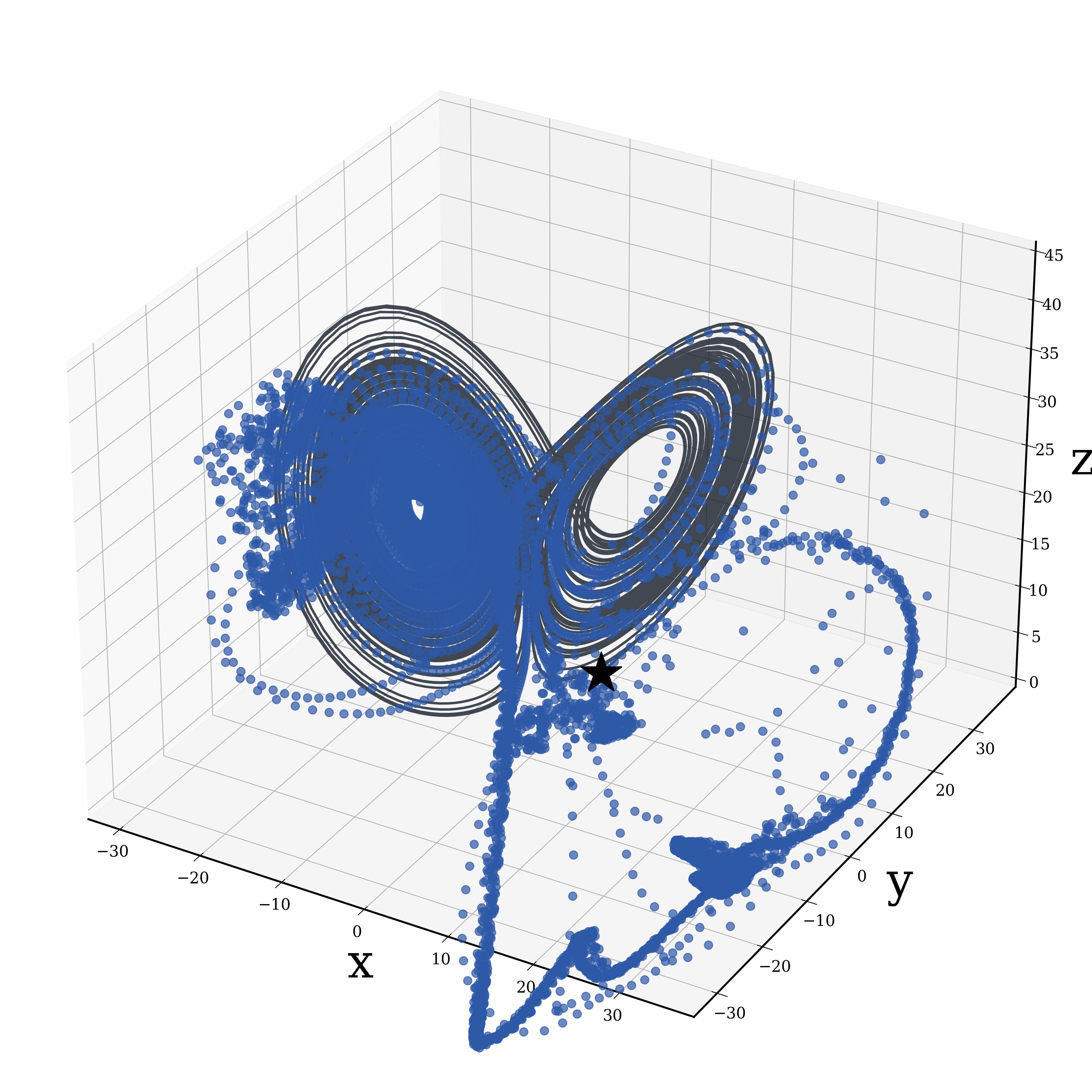}
    \caption{Temporal prediction of a trajectory of $x$ using the proposed self attention model with keys and queries only dependent of the position.}
    \label{fig:self_pos}
\end{figure}

\noindent The easy-attention method gives freedom to the network to find hidden temporal and spatial correlations by not restricting through inner products of the input and softmax, as both temporal and spatial evolutions may not be correlated. Furthermore, the underlying dynamics of the system will remain invariant for quasi and fully chaotic systems, due to the nature of the attractors. Consequently, we believe that it is more convenient to keep the attention frozen and independent of the input matrices

\subsubsection{Hankel DMD for predicting the temporal dynamics of the Lorenz system}
To further analyze the performance of the easy-attention we will implement the Hankel-DMD and HAVOK algorithms. We employ the Hankel-DMD (HDMD) approach as a finite-dimensional Koopman-based operator for predicting the temporal dynamics of the Lorenz system in the same chaotic regime as in the manuscript. Note that we follow the implementation in Ref.~\cite{eivazi_2021108816} for the HDMD algorithm. The HDMD is trained by using 100 time series generated by random initial conditions with $x_0$ $\sim$ $[-5, 5]$, $y_0 \sim [-5, 5]$, $z_0 \sim [0, 5]$ whereas each series contains 10,000 time steps with a time-step size of $\Delta t = 0.01$.  The delay-embedding dimension is considered equal to 64, which is identical to the proposed ML models. The test dataset again is generated by integrating the system with new initial states of $x_0 =  y_0 =  z_0 = 6$ with independent  perturbations $\epsilon_x$, $\epsilon_y$ and $\epsilon_z$ where $\epsilon_{i} \sim \mathcal{N}(0,1)$ {(note that $\mathcal{N}(0, 1)$ denotes a normal distribution with 0 average and a standard deviation of 1) for each variable. 

\noindent We evaluate the prediction obtained by HDMD over 512 time steps on the test dataset, which yields a relative $l_2$-norm error of $60.80\%$, indicating the poor performance on short-term prediction of temporal dynamics. Fig.~\ref{fig:hdmd_attractor} shows the long-term prediction of the variables over 10,000 time steps for HDMD on the test dataset. It can be observed that the HDMD is not able to reproduce the correct dynamical behavior over long times. Moreover, Fig.~\ref{fig:hdmd_signal} clearly depicts the temporal evolution obtained by HDMD, showcasing how the algorithm can only capture the mean for each signal. Additionally, these results align with the discussions in Ref.~\cite{khodkar2019koopman}, which highlight the insufficiency of linear combinations of a finite number of DMD modes in accurately representing the long-term nonlinear characteristics of chaotic dynamical systems. To conclude, one can also observe in Fig.~\ref{fig:hdmd_attractor} the implementation of the HAVOK algorithm, a Koopman based machine learning framework introduced in Ref.~\cite{brunton2017chaos}.

\begin{figure}[ht]
    \centering
    \begin{subfigure}{0.40\linewidth}
        \centering
        \includegraphics[width=\textwidth]{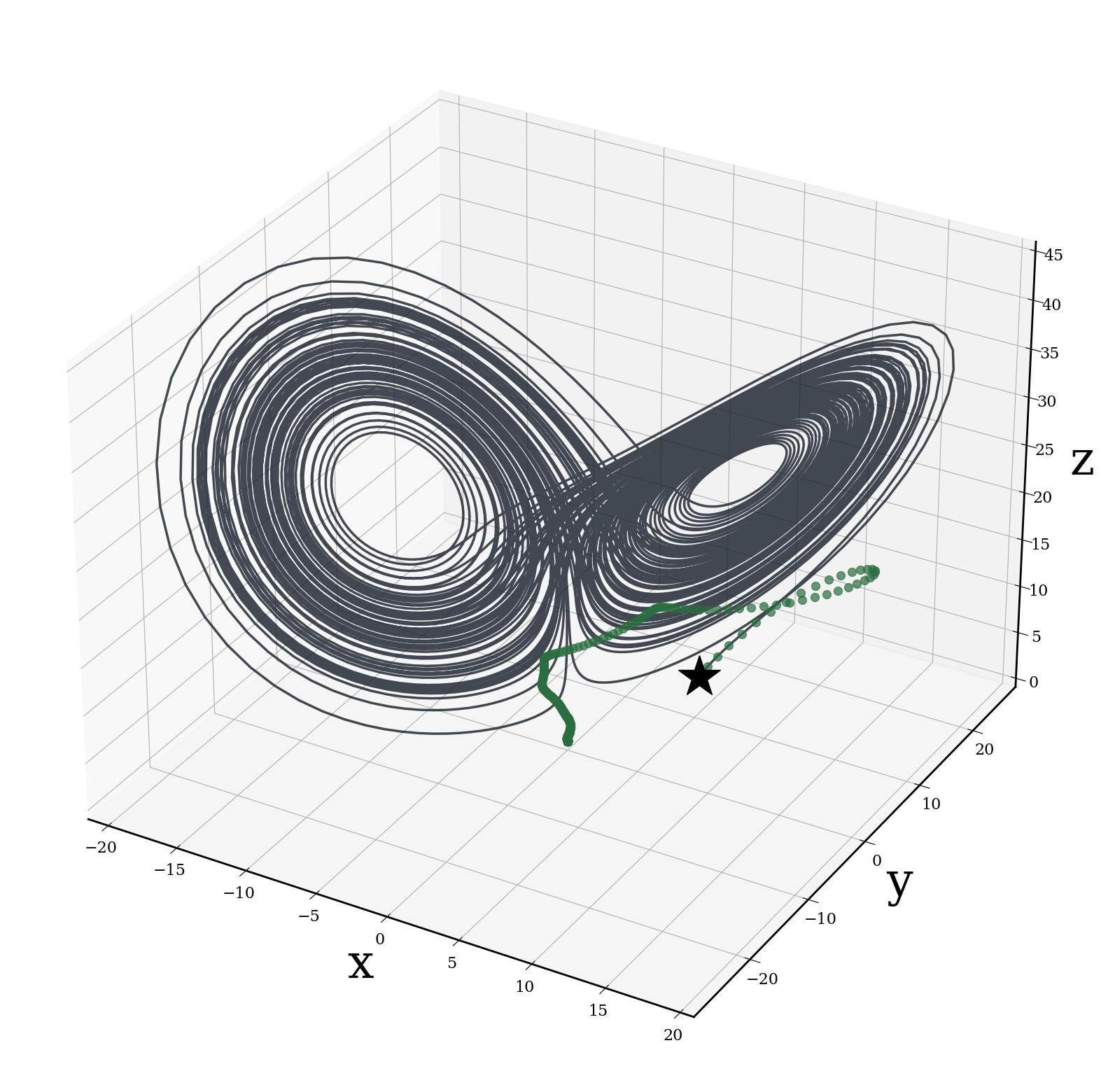}
    \end{subfigure}
    \quad
    \begin{subfigure}{0.40\linewidth}
        \centering
        \includegraphics[width=\textwidth]{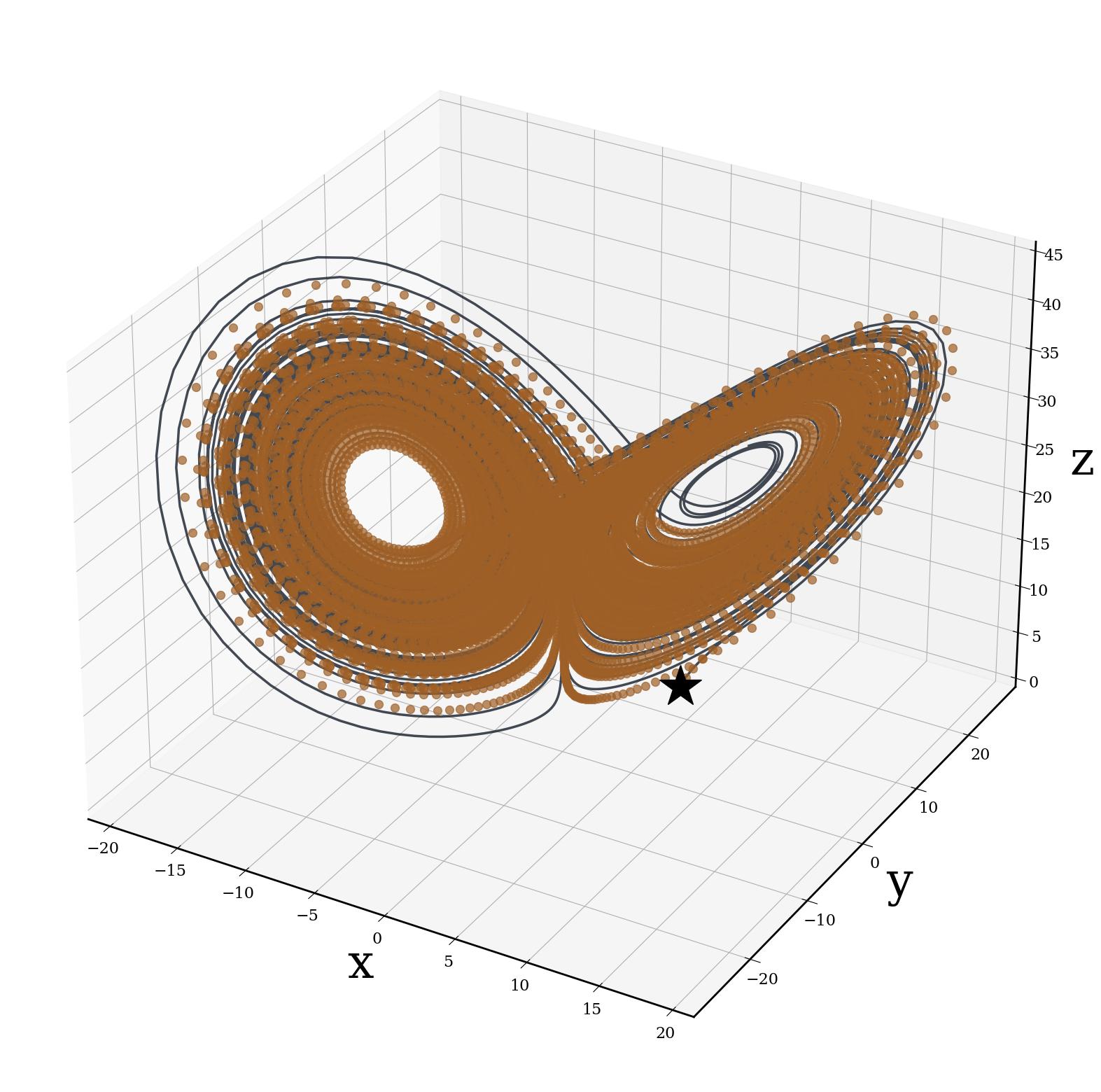}
    \end{subfigure}
    \caption{Visualization of Lorenz system predicted by HDMD (green) and KNF (orange). The test data is generated by integrating the system with new initial states of $x_0 =  y_0 =  z_0 = 6$ and adding perturbations $\epsilon_x$, $\epsilon_y$ and $\epsilon_z$ where $\epsilon_{i} \sim \mathcal{N}(0,1)$ for each variable. Note that the test data is indicated in black as reference and the location of initial condition ($x_0$, $y_0$, $z_0$) is marked as a black star.}
    \label{fig:hdmd_attractor}
\end{figure}

\begin{figure}[ht]
    \centering
    \begin{subfigure}{\linewidth}
        \centering
        \includegraphics[width=0.7\textwidth]{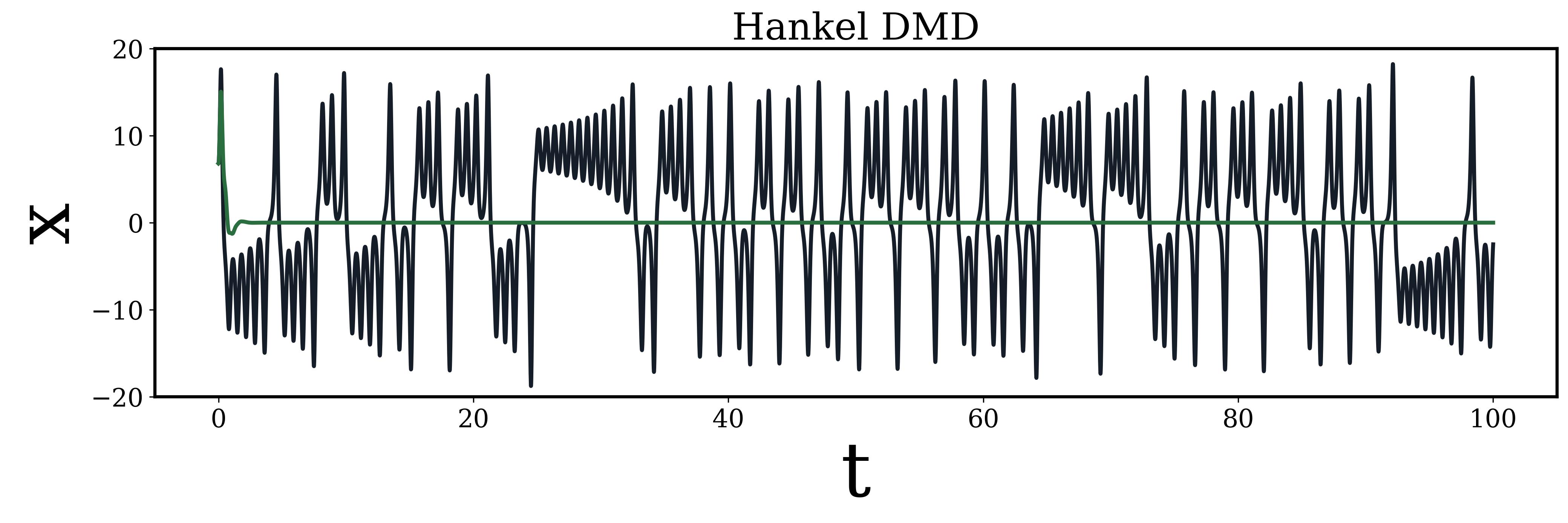}
    \end{subfigure}
    \begin{subfigure}{\linewidth}
        \centering
        \includegraphics[width=0.7\textwidth]{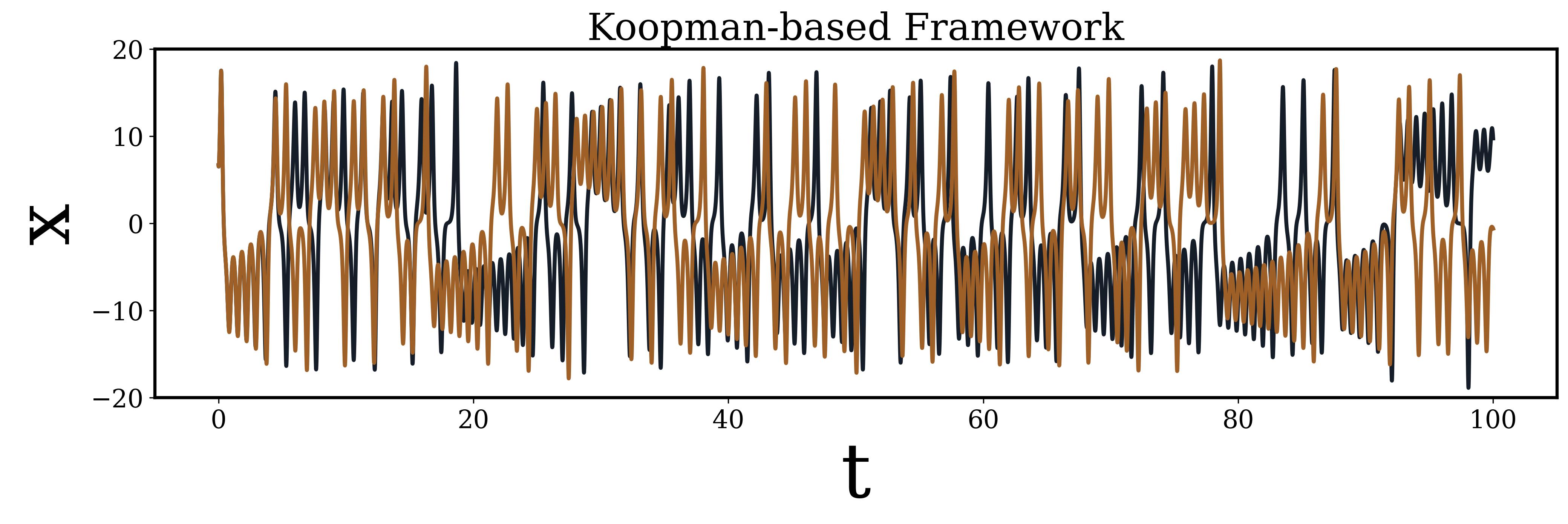}
    \end{subfigure}
    
    \caption{Temporal evolution of variable $x$ from the Lorenz system predicted by HDMD. The test data is generated by integrating the system with new initial states of $x_0 =  y_0 =  z_0 = 6$ and adding the perturbations $\epsilon_x$, $\epsilon_y$ and $\epsilon_z$ where $\epsilon_{i} \sim \mathcal{N}(0,1)$ for each variable. Note that the test data is indicated in black as reference.}
    \label{fig:hdmd_signal}
\end{figure}}

\subsection{Architecture and results for the turbulent-flow case}
In the following section, we introduce the architecture employed in the Low-order model of turbulence shear flows section of the main article. The corresponding transformer architectures are summarized in Table~\ref{tab:attn_arch_9eq}.
\begin{table}[H]
    \centering
    \resizebox{0.7\textwidth}{!}{
    \begin{tabular}{c|cccccc}
        \hline
        \textbf{Name}        & \textbf{$p$} & \textbf{$d_{\rm model}$} & {No.head}  & {Feed-forward} & {No.Block}  & {No.Param}\\ \hline
        Easy-Attn   & 64          & 128                       & 4  & 256          & 1  & 122,378               \\
        Sparse-Easy & 64          & 128                       & 4  & 256          & 1  &  106,250              \\
        Self-Attn   & 64         & 128                      & 4   & 256          & 1   & 155,658              \\ \hline
    \end{tabular}
    }
    \caption{Summary of transformer architectures employed in the nine-equation turbulence model prediction. {We denote the size of time delay as $p$} and the embedding size as $d_{\rm model}$, respectively. Note that the dense easy attention is denoted as Easy-Attn while the sparse easy attention is denoted as Sparse-Easy.}
    \label{tab:attn_arch_9eq}
\end{table}

\noindent It is important to note that transformer models and LSTM networks have different time-delay inputs, {\it i.e.} 64 and 10 respectively, as reported on Table~\ref{tab:attn_arch_9eq}, but all are trained on 1$\times 10^4$ datasets spanning 10,000 time units each. Furthermore, the performance of the various models when reproducing the mean velocity profile $\overline{u}$ and the streamwise velocity fluctuation $\overline{u'^2}$ are shown in Table~\ref{tab:9eq_error}. In Fig.~\ref{fig:arch_transformer_time2space} we depict the transformer architecture used for the nine-equation model by Moehlis {\it et al.}~\cite{moehlis2004low} representing a wall-bounded turbulent flow. The main difference between this architecture and and the one shown in Fig.~\ref{fig:arch_transformer} is the novel embedding used in Fig.~\ref{fig:arch_transformer_time2space}. This new embedding block has as main objective to represent time-averaged trends and dominant events (in magnitude) through average and max poolings.
\begin{table}[ht]
    \centering
    \begin{tabular}{c|cc}
                \hline
                Model & $E_{\overline{u}}$ [$\%$] & 21 [$\%$]\\ \hline
                LSTM &  \textbf{0.45}  & 2.49 \\
                Easy-Attn & 0.82 & \textbf{0.42} \\
                Sparse Easy-Attn & 1.53 & 1.30\\
                Self-Attn & 2.04 & 15.22 \\ \hline
    \end{tabular}
    \caption{Relative error when calculating the mean flow and the streamwise velocity fluctuations  by means of the various models.}
    \label{tab:9eq_error}
\end{table}

\begin{figure}[ht]
    \centering
    \includegraphics[angle=270,width=0.55\textwidth]{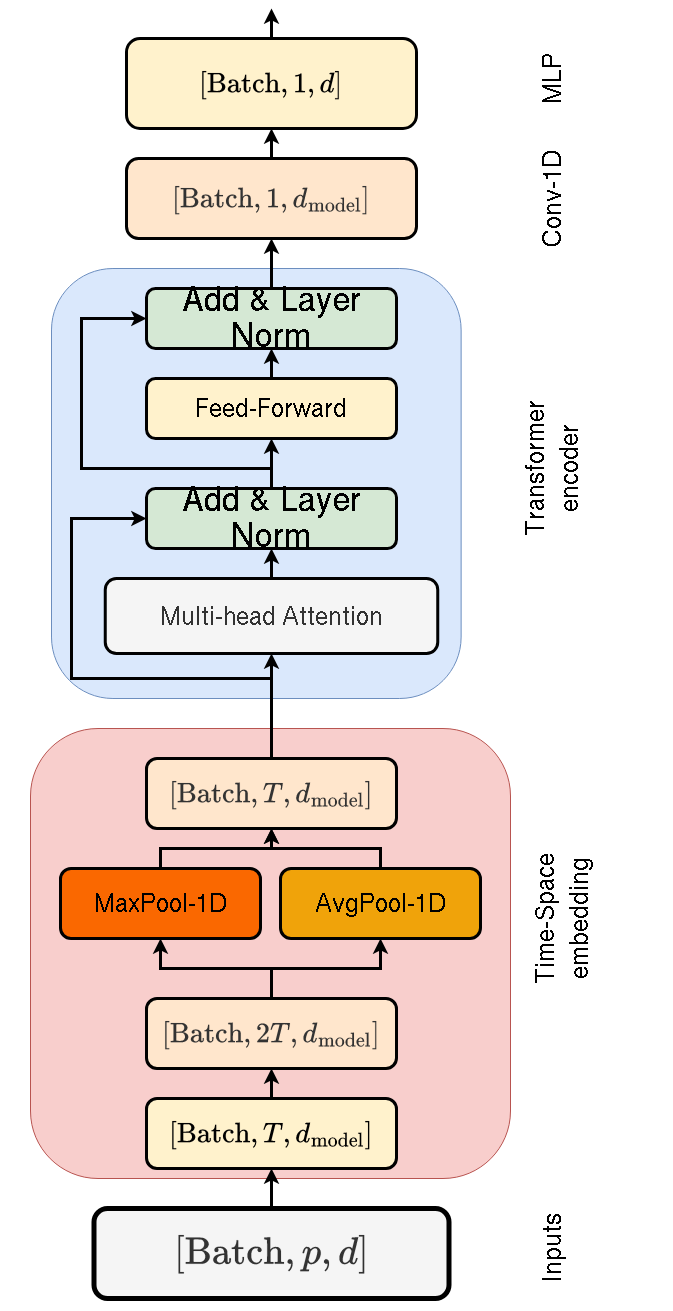}
    \caption{Schematic view of the transformer architecture for temporal-dynamics prediction employed in the turbulent-flow case. The dimension of the output for each layer has been indicated in each block, {where the $p$ denotes} the size of time delay whereas $d$ and $d_{\rm model}$ denote the number of the features and the embedding size, respectively. In the time-space embedding, the yellow and orange blocks denote the fully-connected layer (FC) and the one-dimensional convolution layer (Conv-1D), respectively.}
    \label{fig:arch_transformer_time2space}
\end{figure}

\subsection{Architecture and results for the nuclear-reactor model}

The System Analysis Module (SAM) generic fluoride-salt-cooled high-temperature-reactor (gFHR) model, shown in Fig.\ref{fig:gFHR_reactor_configuration}, is developed in collaboration by the University of Michigan~\cite{Li:22} and Argonne National Laboratories~\cite{Dave:23} using the publicly available information by Karios Power on their Karios Power gFHR (KP-gFHR)~\cite{osti_1868762}. The KP-gFHR focuses primarily on the reactor core, which enabled the creation of the core-power profiles, reactivity-feedback coefficients and the control rod function worth. The system is assumed to be configured in a primary and intermediate loop heat-transport system. The full model contains the thermal-hydraulics for a primary heat-transport system loop, an intermediate solar-salt loop, a point-kinetics reactor core model with Xenon effects. A low-level PID enables the system to reach the user-inputted target power profiles by manipulating the control rod-position in the reactor core. Additionally, the pump power of the primary and intermediate salt pumps are controlled to maintain the reactor inlet temperature and the reactor outlet temperature respectively. 

\begin{figure}[H]
\centering
\includegraphics[width=0.65\textwidth]{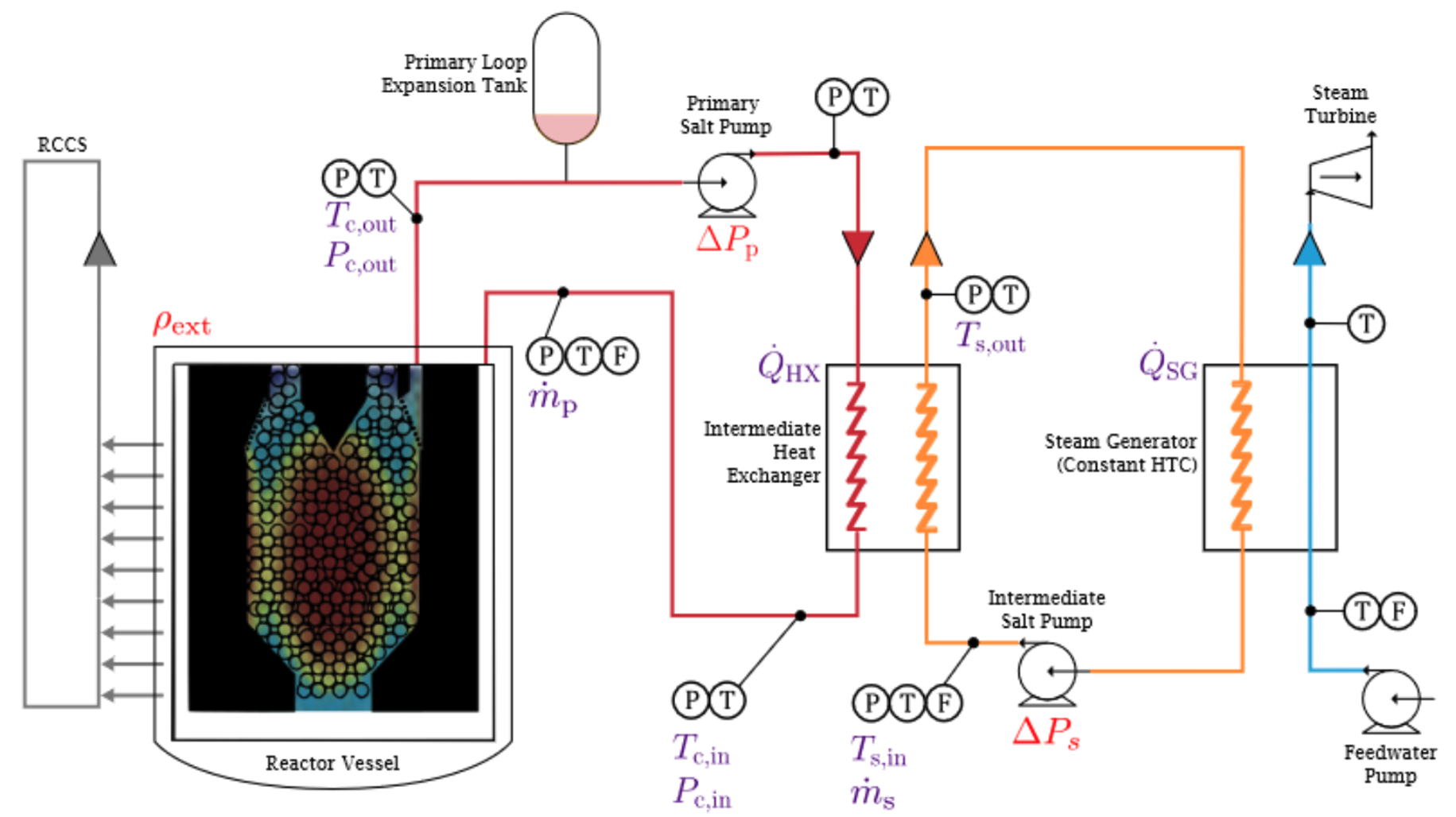}
\caption{Schematic diagram of a generic fluoride-salt-cooled high-temperature reactor (gFHR) by the University of Michigan\cite{Dave:23}. }
\label{fig:gFHR_reactor_configuration}
\end{figure}

\noindent In the following section we introduce the architecture employed in the Nuclear Reactor section of the main article. The employed transformer and LSTM architectures are summarized in Tables~\ref{tab:attn_arch_nuclear} and~\ref{tab:attn_arch_nuclear_lstm} respectively. The transformer structure is the same as in Fig.~\ref{fig:arch_transformer_time2space}. As already introduced the gFHR is a reactor which retains the main neutronics and thermal-hydraulics information of their proprietary Karios Power FHR (KP-FHR) model~\cite{osti_1868762}.


\begin{table}[ht]
    \centering
    \begin{tabular}{c|ccccccc}
                \hline
                \textbf{Name}        & \textbf{$p$} & \textbf{$d_{\rm model}$} & {No.head}  & {Feed-forward} & {No.Block}  & {No.Param} & $t_c$ (s)\\ \hline
                Easy-Attn   & 128          & 256                       & 8  & 512          & 1  & 569,870    & 108,944           \\
                Sparse-Easy & 128          & 256                       & 8  & 512          & 1  &  440,838    & 58,951          \\
                Self-Attn   & 256          & 512                      & 8   & 512          & 1   & 1,993,230   & 122,427           \\ \hline
    \end{tabular}
    \caption{Summary of transformer architecture employed for the prediction of the state variables in the reactor model, considering uniform time step. {We denote the size of time delay as $p$} and the embedding size as $d_{\rm model}$, respectively. Furthermore, the dense easy attention is denoted as Easy-Attn while the sparse easy attention is denoted as Sparse-Easy.}
    \label{tab:attn_arch_nuclear}
\end{table}

\begin{table}[ht]
    \centering
    \begin{tabular}{c|ccccccc}
                \hline
                \textbf{Name}        & \textbf{$p$} & {Hidden Layers}  & {Hidden Size} &  {No.Param} & $t_c$ (s)\\ \hline
                LSTM   & 32    & 2  & 256   & 807,181 & 9,293               \\  \hline
    \end{tabular}
    \caption{Summary of LTSM architecture employed for the prediction of the state variables in the reactor model, considering uniform time step.}
    \label{tab:attn_arch_nuclear_lstm}
\end{table}

\noindent In Fig.~\ref{fig  :reactor_evo_equi} it is shown the temporal signal for the Core Flow [\rm{kg/s}] of the reactor. The signal depicted for the easy attention matches accurately the dynamics of the reference data, both in amplitude and oscillations. This can be easily assessed on the main article when studying the PDF of the Core flow. The sparse and LSTM networks lack the oscillation reconstruction while the self-attention over predicts the amplitude of the signal. Finally, it is important to clarify that the transformer was trained on the 13 most relevant variables for monitoring our model.

\begin{figure}[ht]
    \centering
    \begin{subfigure}{0.48\linewidth}
        \centering
        \caption{{Easy attention}}
        \includegraphics[width=\textwidth]{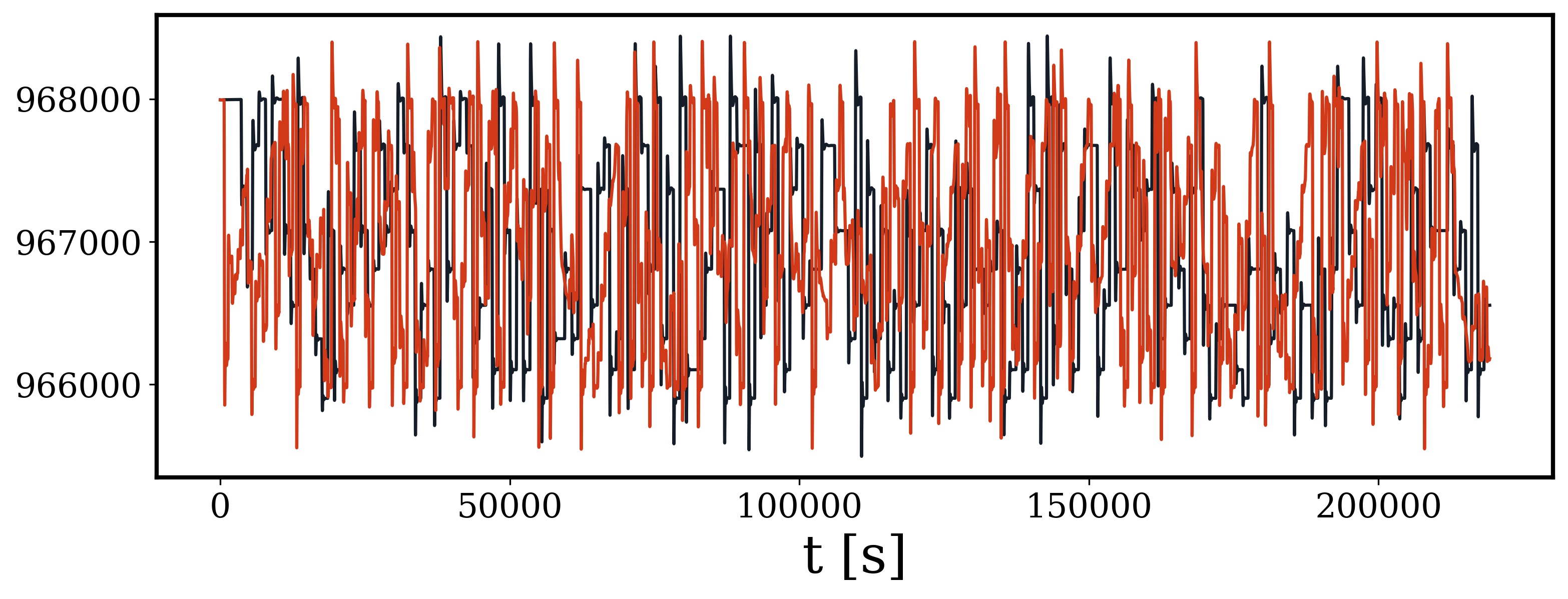}
    \end{subfigure}
    \quad
    \begin{subfigure}{0.48\linewidth}
        \centering
        \caption{{Sparse easy attention}}
        \includegraphics[width=\textwidth]{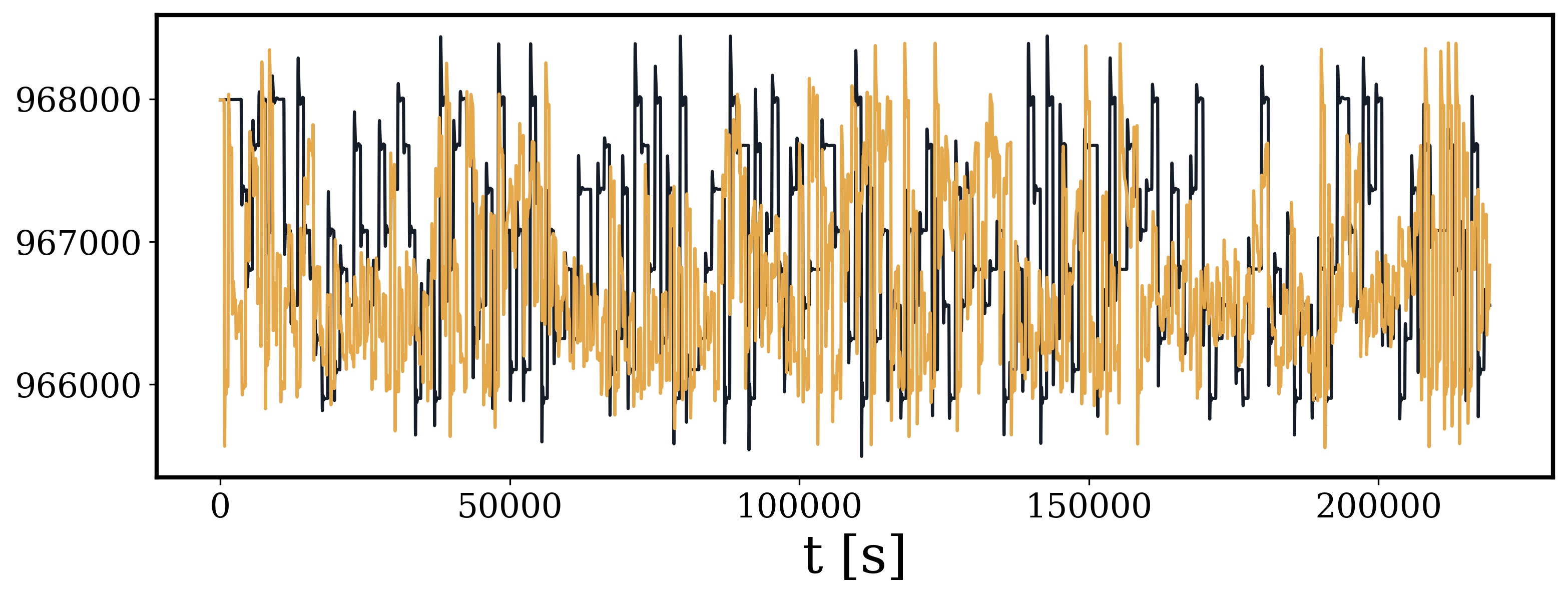}
    \end{subfigure}
    \quad
    \begin{subfigure}{0.48\linewidth}
        \centering
        \caption{{Self attention}}
        \includegraphics[width=\textwidth]{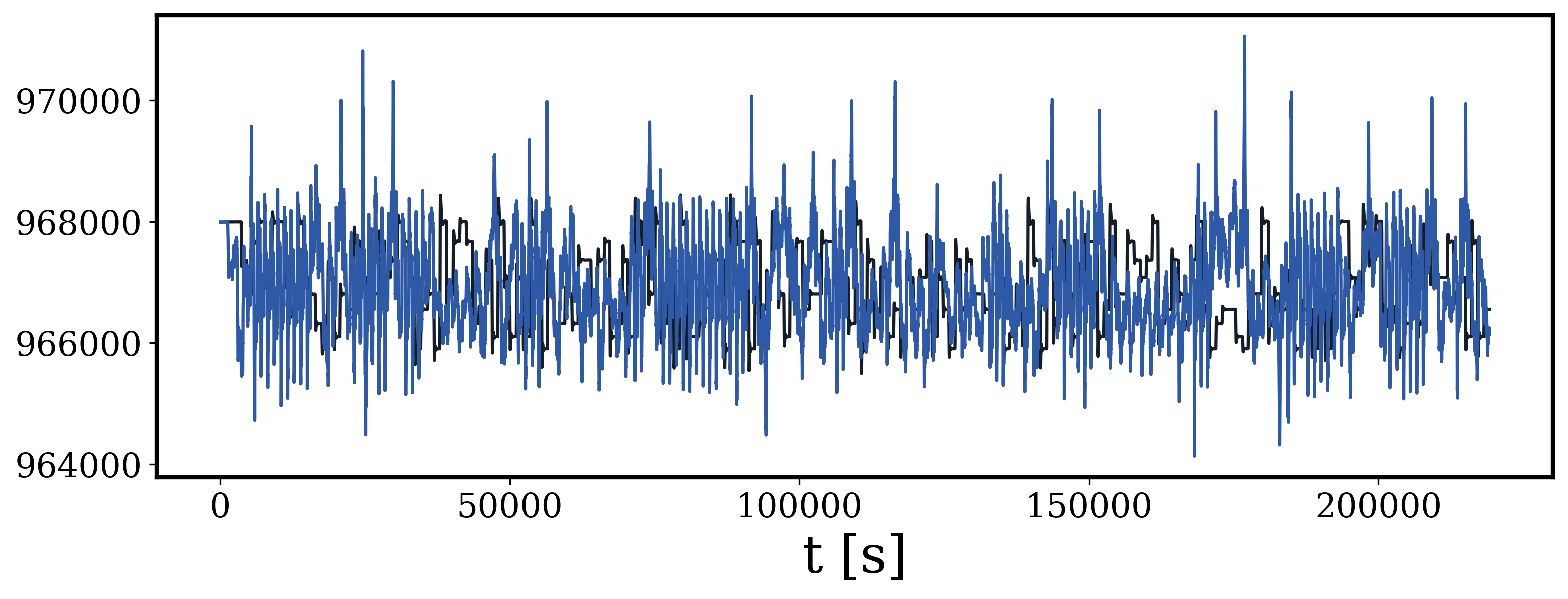}
    \end{subfigure}
    \quad
    \begin{subfigure}{0.48\linewidth}
        \centering
        \caption{{LSTM}}
        \includegraphics[width=\textwidth]{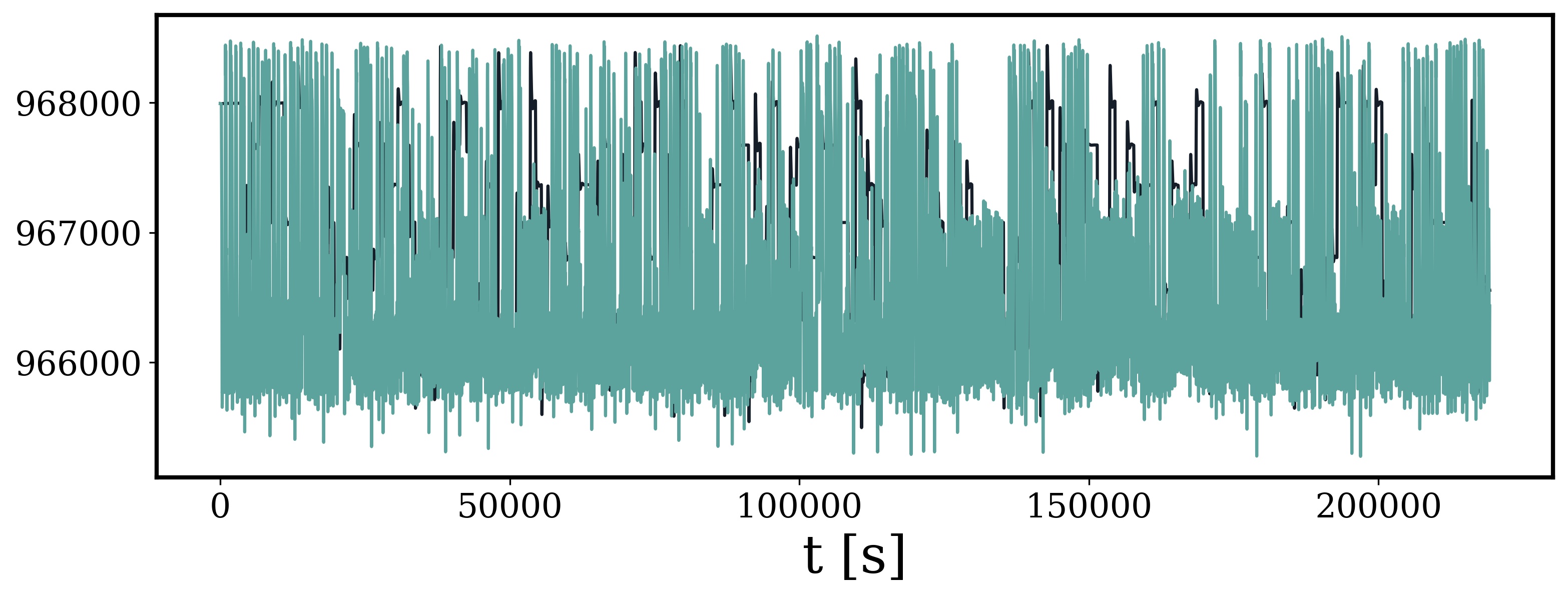}
    \end{subfigure}
    \caption{Temporal prediction of the Core Flow [kg/s] for the four methods under study. The black line denotes the reference in all the cases.}
    \label{fig  :reactor_evo_equi}
\end{figure}

\section*{Schematic representations of the easy-attention mechanism}
In Fig.~\ref{fig:Easy-attn-schematic} we show a diagram illustrating the multi-head easy-attention implementation, where it is possible to differentiate between temporal and feature dependencies.

\begin{figure}[H]
	\centering
	
	\begin{subfigure}{0.45\linewidth}
		\centering
		\includegraphics[width=0.85\textwidth]{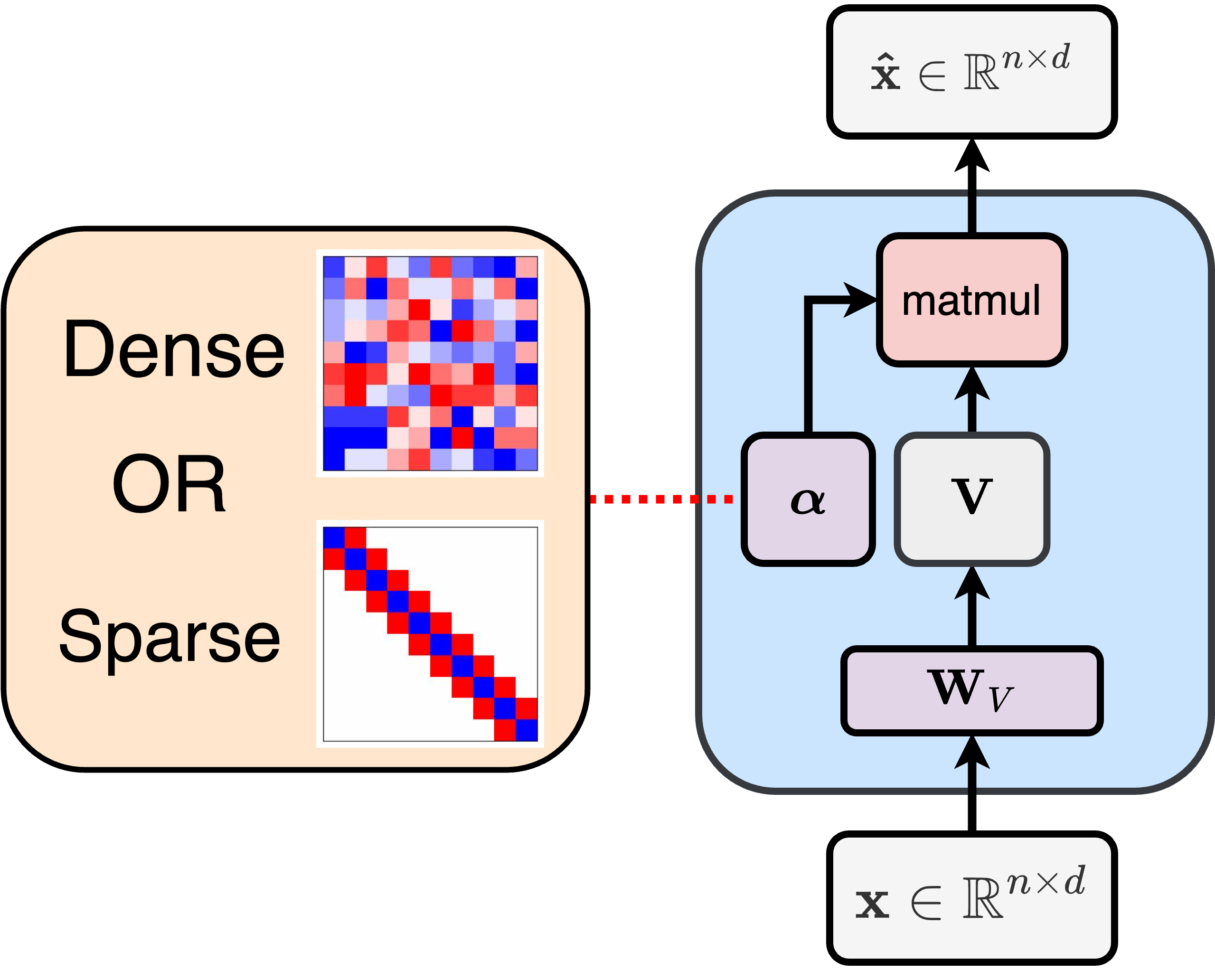}
		\label{fig:single_easy_attn}
	\end{subfigure}
	\quad
	\begin{subfigure}{0.5\linewidth}
		\centering
		\includegraphics[width=0.45\textwidth]{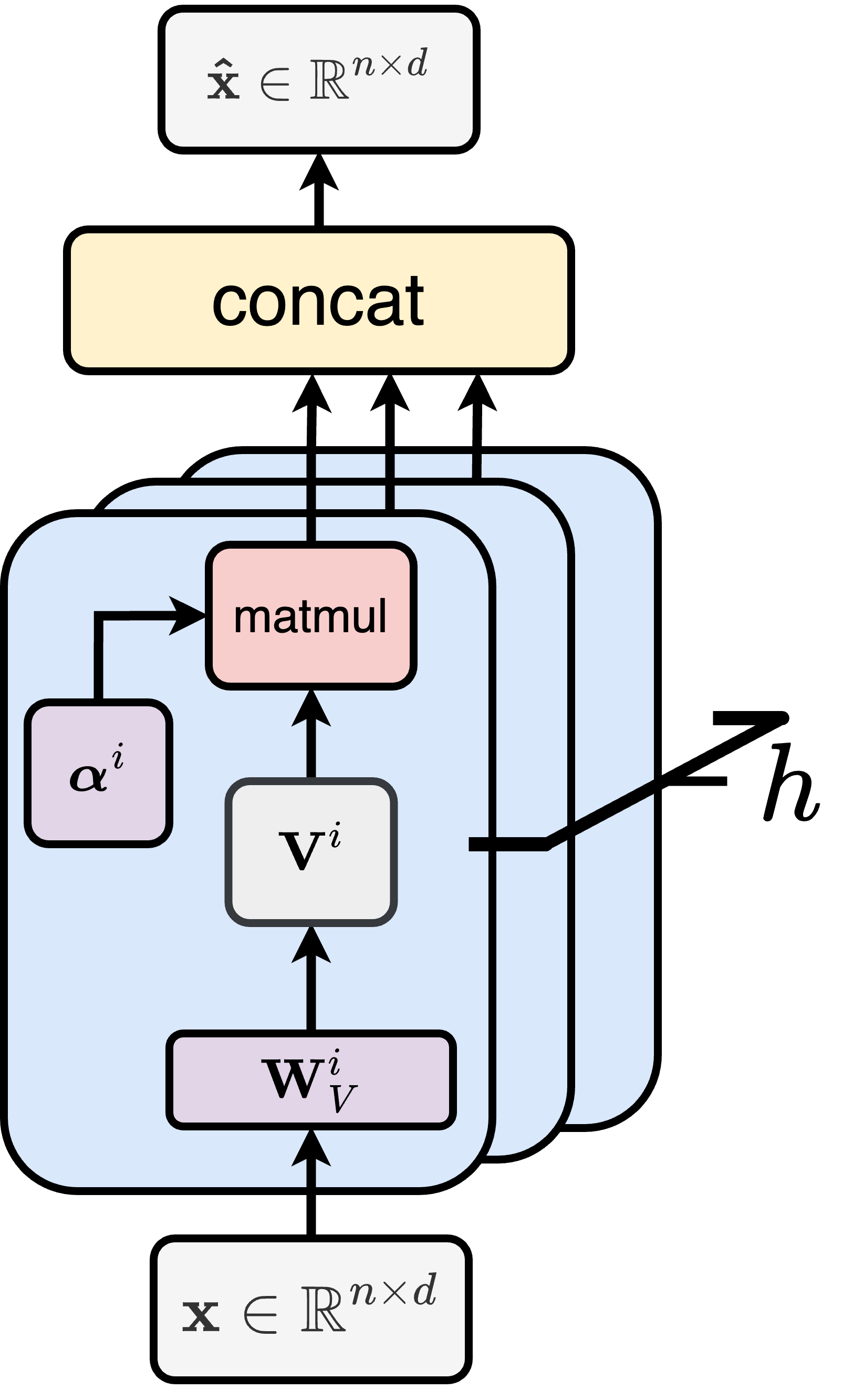}
		\label{fig:multi_easy_attn}
	\end{subfigure}
	\caption{
		Schematic view of single easy-attention mechanism (left) and multi-head easy attention (right). {Note that} the trainable weight {tensors} $\mathbf{W}_V$ and $\boldsymbol{{\alpha}}$ are denoted by purple blocks. The difference between the dense and sparse easy-attention methods is illustrated in the orange block. Note that {$\rm matmul$} in the red block and {$\rm concat$} in the yellow one denote the matrix-product operator and tensor concatenation, respectively.}
	\label{fig:Easy-attn-schematic}
\end{figure}

\end{document}